\documentclass{article}
\usepackage{iclr2027_conference,times}

\usepackage{amsmath,amsfonts,bm}

\def\eqref#1{equation~\ref{#1}}
\def\1{\bm{1}}

\DeclareMathAlphabet{\mathsfit}{\encodingdefault}{\sfdefault}{m}{sl}
\SetMathAlphabet{\mathsfit}{bold}{\encodingdefault}{\sfdefault}{bx}{n}

\usepackage{hyperref}
\usepackage{url}
\usepackage[hyphens]{xurl}
\usepackage{amsmath}
\usepackage{amssymb}
\usepackage{amsthm}
\usepackage{amsfonts}
\usepackage{booktabs}
\usepackage{multirow}
\usepackage{graphicx}
\usepackage{xcolor}
\usepackage{nicefrac}
\usepackage{microtype}
\usepackage{enumitem}
\usepackage{caption}
\usepackage{etoc}
\usepackage{wrapfig}
\usepackage{tabularx}
\usepackage{array}

\newcommand{\method}{\textsc{PTSDiff}}
\definecolor{phgray}{gray}{0.42}

\providecommand{\best}[1]{\begingroup\bfseries\boldmath #1\endgroup}
\providecommand{\secondbest}[1]{\underline{#1}}
\providecommand{\cmark}{\ensuremath{\checkmark}}

\newtheorem{proposition}{Proposition}

\newtheorem{definition}{Definition}

\theoremstyle{remark}

\title{Cyclostationary Phase Conditioning \\ for Medical Time Series Diffusion}

\author{Anonymous authors\\
Paper under double-blind review}

\author{\normalfont
\begin{tabular*}{\dimexpr\textwidth-2\tabcolsep\relax}[t]{@{\extracolsep{\fill}}lll@{}}
\textbf{Samuel Ruip\'erez-Campillo}\thanks{Equal contribution. Correspondence to \texttt{\{sruiperez,mcopetti,jogoncalves\}@inf.ethz.ch}.}
  & \textbf{Michele Copetti}\footnotemark[1] & \textbf{Jorge da Silva Gonçalves}\footnotemark[1] \\
Dept. of Computer Science & Dept. of Computer Science & Dept. of Computer Science \\
ETH Zurich & ETH Zurich & ETH Zurich \\[17pt]
\textbf{Sonia Laguna} & \textbf{Thomas Hofmann} & \textbf{Julia E. Vogt} \\
Dept. of Computer Science & Dept. of Computer Science & Dept. of Computer Science \\
ETH Zurich & ETH Zurich & ETH Zurich
\end{tabular*}}

\iclrfinalcopy % Uncomment for camera-ready, NOT for submission.

\begin{document}

\maketitle
\fancyhead{} % TO BE REMOVED

\begin{abstract}

Many physiological time series, such as cardiac and brain recordings, exhibit cyclostationarity: their statistics vary periodically with an underlying cycle phase. Corruption from motion, poor contact, and physiological interference obscures morphology needed for diagnosis, making signal restoration essential. Existing diffusion approaches condition on corrupted observations alone and must learn cyclic structure implicitly. We instead propose two inductive biases which encode cyclostationarity: a shift-covariant wavelet representation and dense per-sample phase conditioning inferred from the corrupted input. We further introduce a training-free cyclostationarity index that quantifies phase structure and predicts when phase conditioning will help. Finally, we propose antithetic coupling of reverse trajectories to reduce sampling variance while achieving comparable performance with fivefold fewer network evaluations. Across modalities, our results show that explicitly encoding measurable cyclic structure improves physiological time-series restoration.\looseness-1

\end{abstract}

\section{Introduction}
\label{sec:intro}

Recurrent structure is a common property of signals arising from natural processes. Physiological time series are a prominent example, reflecting the cyclic biological processes that generate them. A heartbeat appears in the electrocardiogram (ECG) and, moments later, as a pulse of blood volume at the wrist in photoplethysmography (PPG) \citep{xie_computational_2020,nitzan_9_2022}; a muscle contraction appears as a burst of electrical activity in electromyography (EMG) \citep{khorrami_chokami_identification_2021}; and the brain's electrical activity rises and falls at characteristic frequencies in the electroencephalogram (EEG) \citep{brenner_periodic_1990}. These signals support cardiac diagnosis, wearable monitoring, prosthetic control and brain--computer interfaces, yet are routinely corrupted by movement, poor electrode contact, and interference from other physiological sources \citep{chatterjee_review_2020,ismail_heart_2021,boyer_reducing_2023,jiang_removal_2019}. Restoration methods aim to remove these corruptions and recover the underlying clean signal.

These time series are not stationary: their mean and covariance change over time, and do so approximately periodically as a function of a latent cycle phase. This property is known as wide-sense \textit{cyclostationarity} \citep{gardner2018statistically, napolitano2025discovering}. Despite their recurring structure, existing restoration methods do not model this explicitly. This limitation is especially important for diffusion-based approaches, which have become the state of the art for recovering signals from corrupted observations. Diffusion models have been adapted across physiological modalities \citep{li_descod-ecg_2024,liu_sdemg_2024,huang_eegdfus_2025, ruiperez2025physics, ruiperez2026antithetic} and to spectral representations \citep{crabbe_time_2024,li_tfcdiff_2025, qin2026wavedist}, yet all condition the reverse process solely on the corrupted observation. Therefore, the denoiser must learn from data that samples separated by one cycle correspond to the same position within the underlying cycle.\looseness=-1

A natural question arises: can modeling cycle structure improve restoration? To answer this, we introduce~\method{}, to include direct cycle-phase information into the diffusion process, and investigate whether this benefit can be predicted before training. We encode this structure through two complementary inductive biases. First, instead of operating directly in the time domain, we perform diffusion in a shift-covariant wavelet representation, making the model insensitive to the arbitrary starting position of the input window, which we hypothesize should provide a consistent benefit across modalities. Second, we condition the denoiser on a phase field estimated from the corrupted input, a dense per-sample representation of position within the current cycle, which should be most useful when the signal is strongly structured by its cycle. Together, the two components remove sensitivity to absolute window position while providing explicit information about position within the underlying cycle.\looseness-1
 %This pairing yields a testable prediction: removing the shift-covariant representation should have a similar effect across modalities, whereas removing phase conditioning should degrade performance in proportion to the strength of the underlying cycle structure.

To understand the effect of phase conditioning we analyze it along two axes: how phase information is estimated and how strongly the underlying signal is structured by its cycle. We find that estimation quality is critical: a classical detector applied to the corrupted input performs no better than omitting phase conditioning, whereas a learned estimator substantially improves restoration. Moreover, we study whether the usefulness of phase conditioning can be anticipated prior to training. To this end, we introduce a cyclostationarity index that measures the fraction of signal variance explained by cycle phase. This provides a training-free estimate of which signals benefit from phase conditioning, avoiding unnecessary model complexity and training when cycle structure is weak.\looseness-1
% It also matters how the phase field is obtained, since a classical detector run on the corrupted input turns out to be worth nothing, while a learned estimator of the clean field recovers the entire gain. 
% The same reasoning bounds how much phase conditioning can help. 
% We define a cyclostationarity index and show that it equals the fraction of a signal's variance explained by cycle phase.
% Because the index is computed from clean data alone, it predicts the benefit of phase conditioning before any model is trained: signals should improve in the order indicated by the index, while signals without cycle structure should not benefit.

Beyond improving the conditional model with both inductive biases, diffusion restoration remains stochastic at inference: different reverse trajectories can yield different estimates, while averaging several trajectories improves the estimate but is computationally expensive. We therefore study antithetic coupling \citep{hammersley_morton_1956,ruiperez2026antithetic} as an inference-time variance-reduction strategy and show that one antithetic pair can match or outperform ten independent trajectories, with fivefold fewer network evaluations. Overall, our contributions are summarized as:\looseness-1
\vspace{-0.1cm}
\begin{itemize}[nosep, leftmargin=*]

\item We propose \method{}, a novel formulation of physiological signal restoration under a cyclostationary prior. We address a limitation of existing diffusion-based methods that overlook this structure, through two inductive biases: shift-covariant wavelet diffusion and cycle-phase conditioning.\looseness-1

%\item We introduce a cyclostationarity index that quantifies cycle structure from clean data and serves as a training-free predictor to identify when phase conditioning improves restoration quality.
\item We introduce a cyclostationarity index that quantifies cycle structure from clean data and gives a training-free indication of when phase conditioning is likely to help. %softened claim.

\item We achieve state-of-the-art restoration and improve downstream task performance across studied benchmarks in four modalities, with a sampling-efficient solution incorporating antithetic coupling.

% \item A formulation of medical time-series restoration under a cyclostationary signal model, in which cycle phase is an explicit conditioning variable rather than structure the denoiser must infer; 
% \item A cyclostationarity index, computable from clean data alone, that measures how much of a signal is explained by its cycle, together with an identity relating it to the variance that phase explains; 
% \item A phase-conditioned diffusion model that incorporates this structure on various physiological scenarios; and 
% \item Evidence that this improves restoration and downstream performance on clinical tasks across those modalities, at a fraction of the sampling cost when reverse trajectories are coupled antithetically.
\end{itemize}

\begin{figure}[t]
\centering
\includegraphics[width=\linewidth, trim={25 50pt 35 10pt}, clip]{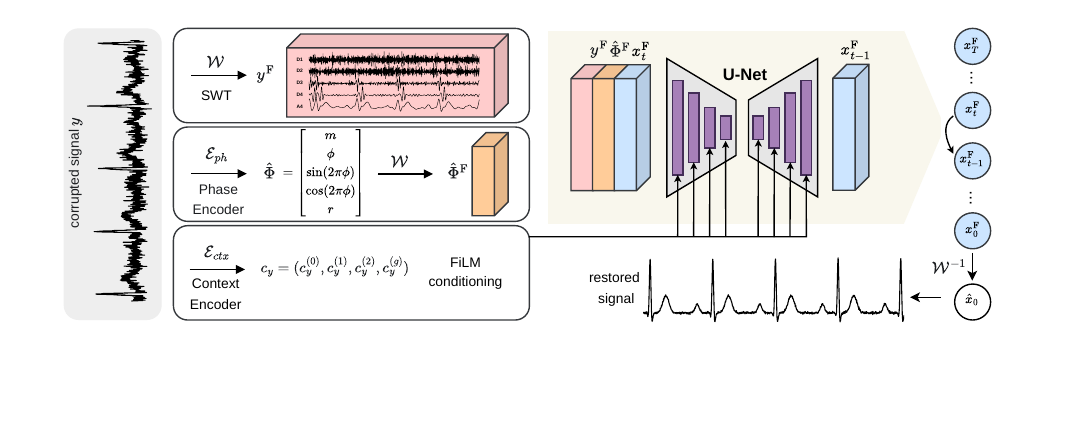}
\caption{Overview of \method. The corrupted observation $y$ feeds the context encoder and the phase encoder; it is frame-transformed (top) and concatenated with the frame-transformed phase channels (middle) and the latent $x_t^{\mathrm{F}}$ as U-Net input. The context encoder (bottom) modulates the backbone through FiLM embeddings, and the U-Net drives the update to $x_{t-1}^{\mathrm F}$. Block-level detail is in Figure~\ref{fig:architecture-detail}.\looseness-1 \vspace{-0.5cm}}
\label{fig:architecture}
\end{figure}

\section{Related work}
\label{sec:related}

\paragraph{Restoration for physiological signals.} 
Medical recordings are noisy. Artifacts obscure the details that clinical interpretation relies on, and corrupted segments are otherwise discarded, from intensive care to ambulatory and consumer recordings \citep{chatterjee_review_2020,ismail_heart_2021}. Classical restoration pipelines built on Kalman filtering, empirical mode decomposition, adaptive filtering and wavelet shrinkage remain standard~\citep{sayadi_ecg_2008,blanco-velasco_ecg_2008,gao_denoising_2010, lee_reduction_2003,ismail_heart_2021,jiang_removal_2019,boyer_reducing_2023}. However, these methods assume fixed signal and noise spectra~\citep{ruiperez2024can, ruiperez-campillo_reducing_2026}, and they degrade under non-stationary artifacts and suppress the low-amplitude morphology that carries diagnostic content \citep{chatterjee_review_2020,watanabe_beyond_2025}. Learned deterministic restorers, primarily convolutional, recurrent and attention-based autoencoders, outperform them~\citep{chiang_noise_2019,antczak_deep_2019,chen_elimination_2024, lai_enhanced_2025,cui_dual-branch_2024,wang_ecg_2023}, but learn a single deterministic map and model the conditional mean implicitly, with no control over estimator variance. Generative restoration \citep{ho_denoising_2020,song_score-based_2021, saharia2022paletteimagetoimagediffusionmodels} addresses this for different time-series modalities, i.e. DeScoD-ECG~\citep{li_descod-ecg_2024}, SDEMG~\citep{liu_sdemg_2024} and EEGDfus~\citep{huang_eegdfus_2025}. Moreover, general-purpose restoration samplers \citep{kawar2022denoising, chung2022diffusion}  focus on the inverse problem, while accelerated solvers \citep{song2020denoising, lu2022dpm} reduce the number of sampling steps; neither directly targets estimator variance at fixed Number of Function Evaluations (NFE). Further per-modality details are provided in Appendix~\ref{app:related-extended}\looseness-1. %

\vspace*{-0.15cm}
\paragraph{Transformed-domain diffusion.}% 
Diffusion-based time-series models typically operate in the native time domain, but recent work has explored whether alternative representations provide more suitable inductive biases.
\citet{crabbe_time_2024} argue that diffusing in a spectral basis suits signals with localized spectral structure, and FIDE \citep{galib_fide_2024} prevents high-frequency components from vanishing in the reverse process. Global bases sacrifice temporal localization, motivating time--frequency representations \citep{li_short_time_2022}: MECG-E uses the STFT \citep{hung_mecg-e_2024} and TFCDiff the DCT \citep{li_tfcdiff_2025}. Common to these, and made explicit here, is that the choice of frame, i.e., the signal representation in which diffusion operates, determines which symmetries are built into the denoiser. In this work, we adopt the Stationary Wavelet Transform (SWT) for its shift-covariance rather than its spectral properties \citep{nason_stationary_1995,coifman1995translation}.

\vspace*{-0.15cm}
\paragraph{Periodicity and variance reduction.} Periodic structure is routinely exploited in forecasting architectures \citep{wu2022timesnet, wu2021autoformer}. In physiological modeling, PulseImpute \citep{xu_pulseimpute_2023} and PulseDiff \citep{jenkins_improving_2023} exploit cardiac repetition for imputation, while latent-trajectory models represent cycles geometrically \citep{ryser_anomaly_2022}. Furthermore, oscillatory and burst structure is documented for EEG and EMG \citep{brenner_periodic_1990, khorrami_chokami_identification_2021}. These works establish that periodicity can be useful, but not how much a given signal should benefit from it, which is the gap addressed in Section~\ref{sec:cyclo-index}. Antithetic variates are a variance-reduction technique \citep{hammersley_morton_1956,owen2013montecarlo, choi2026enhanced} and were introduced into conditional diffusion sampling in physiological time series by \citet{ruiperez2026antithetic}. We adopt this coupling, and derive its effect at matched NFE in the restoration setting.\looseness-1 %, measure the correlation on which this effect depends, and test whether the gain is sampler-specific or model-specific.

\section{Method: \method{}}
\label{sec:method}

We propose \method{}%\footnote{The code is publicly available at: \url{https://to_be_released_upon_acceptance}.}
, a method to incorporate the cyclostationary structure (Section~\ref{sec:setting}) into diffusion through a shift-covariant representation and explicit cycle-phase conditioning. We then define the training objective, a training-free predictor of phase utility, and an antithetic inference estimator.\looseness-1

\subsection{Preliminaries}
\label{sec:setting}

\paragraph{Cyclostationary restoration.}
Let $x_0, y \in \mathbb{R}^L$ denote a clean signal and its corrupted observation, respectively, each of length $L$ and with sample indices $n,n'\in \{1,\ldots,L\}$; restoration seeks to recover $x_0$ given $y$. We assume $x_0$ is drawn from a process that is cyclostationary in the \textit{wide sense}: there exists a latent phase $\varphi_n \in [0,1)$, advancing at a locally varying rate, with
\begin{equation}
\label{eq:cyclostationary}
\mathbb{E}\!\left[x_0[n]\,\middle|\,\varphi_n = \varphi\right] = \mu(\varphi),
\qquad
\operatorname{Cov}\!\left(x_0[n], x_0[n'] \,\middle|\, \varphi_n, \varphi_{n'}\right)
= C(\varphi_n, \varphi_{n'}),
\end{equation}
so first- and second-order statistics depend on phase, not on absolute time. This is weaker than periodicity, since successive cycles differ in length and morphology, and stronger than stationarity. Different physiological systems may satisfy this to different degrees.

\paragraph{Conditional diffusion.}
We use an $\epsilon$-parameterized DDPM \citep{ho_denoising_2020}.
For a generic representation $u_0$ with variance schedule $\{\beta_t\}_{t=1}^{T}$, $\alpha_t=1-\beta_t$, and $\bar{\alpha}_t=\prod_{i=1}^{t}\alpha_i$. The forward process is\looseness-1
\begin{equation}
\label{eq:forward}
u_t
=
\sqrt{\bar\alpha_t}\,u_0
+
\sqrt{1-\bar\alpha_t}\,\epsilon,
\qquad
\epsilon\sim\mathcal N(0,I).
\end{equation}
A denoising network $\epsilon_\theta$ is trained to predict the injected noise $\epsilon$ at each diffusion step. At inference, the reverse process generates samples starting from noise $u_T\sim\mathcal{N}(0, I)$. Sections~\ref{sec:method-frame}--\ref{sec:method-phase} specify the representation $u_0$ and the conditioning supplied to the denoiser.

% Our method follows from the cyclostationary structure introduced in Section~\ref{sec:setting}. We incorporate this structure into a diffusion model through two complementary inductive biases: a shift-covariant representation that removes dependence on absolute window position, and a dense phase field that supplies position relative to the underlying cycle. We first describe these two mechanisms and derive a training-free measure of when phase information should be useful. We then specify the training objective and, finally, an antithetic estimator that reduces the variance of the stochastic reverse process at inference.

\subsection{Diffusion in a shift-covariant frame}
\label{sec:method-frame}

The first inductive bias removes sensitivity to the arbitrary position at which a signal window begins. Under the cyclostationary model, absolute position is not informative; what matters is position relative to the underlying cycle. We therefore perform diffusion in a shift-covariant representation that preserves temporal alignment.\looseness-1

\paragraph{Shift-covariant frame.} 
We instantiate the shift-covariant representation with the stationary wavelet transform (SWT)~\citep{nason_stationary_1995}. Let $\mathcal{W}$ denote its analysis operator and $\mathcal{W}^{-1}$ its inverse, with
$x_0^{\mathrm{F}}=\mathcal{W}(x_0)$ and
$y^{\mathrm{F}}=\mathcal{W}(y)$ denoting the corresponding frame coefficients of the clean signal $x_0$ and its corrupted observation $y$, respectively.
For an input $a_0$, the level-$j$ decomposition is
\begin{equation}
\label{eq:swt}
a_j[n] = \sum_k h_j[k]\,a_{j-1}[n-k],
\qquad
d_j[n] = \sum_k g_j[k]\,a_{j-1}[n-k],
\end{equation}
with $h_j$ and $g_j$ the upsampled low- and high-pass filters at level $j$.
Unlike the standard discrete wavelet transform (DWT), the SWT omits \emph{decimation}, i.e.\ the downsampling applied after each level.
As a result, $\mathcal{W}:\mathbb{R}^{L}\to\mathbb{R}^{(J+1)\times L}$ preserves temporal resolution and is exactly shift-covariant, or equivalently translation-equivariant. Writing $(S_\tau x)[n]=x[n-\tau]$ for the cyclic shift by $\tau$ samples,
\begin{equation}
\label{eq:shift-cov}
\mathcal{W}(S_\tau x)
=
S_\tau\,\mathcal{W}(x)
\qquad
\forall \tau\in\mathbb{Z},
\end{equation}
where $S_\tau$ acts identically on every channel.
The decimated DWT does not satisfy this property for arbitrary shifts: downsampling by $2^j$ at level $j$ restricts equivariance at that level to shifts that are multiples of $2^j$, and therefore to multiples of $2^J$ for the full $J$-level transform.
This loss of shift covariance motivates the undecimated construction used in translation-invariant wavelet denoising
\citep{nason_stationary_1995,coifman1995translation}.
Because $\mathcal{W}$ maps $L$ samples to $(J{+}1)L$ coefficients, it is redundant rather than a basis.
More precisely, it forms an \emph{overcomplete frame}: a redundant representation that nevertheless admits stable reconstruction.
By definition \citep{daubechies1992ten}, there exist constants $0<A\le B<\infty$ such that
\begin{equation}
\label{eq:frame-bounds}
A\|x\|_2^2
\;\le\;
\|\mathcal{W}x\|_2^2
\;\le\;
B\|x\|_2^2.
\end{equation}
For the orthogonal wavelet families used here, the undecimated transform is a tight frame with
$A=B=J+1$, so
$\mathcal{W}^{-1}=A^{-1}\mathcal{W}^{\!\top}$ and reconstruction is non-expansive up to the factor $A^{-1/2}$.
Two consequences are useful for our model. First, the subbands remain aligned on the original sample grid, allowing them to be stacked as channels of a length-preserving backbone. Second, the time-domain phase field introduced in Section~\ref{sec:method-phase} can be transformed by \(\mathcal{W}\) and concatenated channel-wise without resampling. 
We use $J=4$ levels, yielding detail channels $(d_1,\dots,d_4)$ and one approximation channel $a_4$.
For sampling rate $f_s$, the channel $d_j$ corresponds approximately to
$[f_s2^{-(j+1)},f_s2^{-j}]$, while $a_J$ contains the residual frequencies below $f_s2^{-(J+1)}$.
The resulting per-modality band mappings and the choice of wavelet family are reported in Appendix~\ref{app:band-mapping}; the choice of $J$ is evaluated in Appendix~\ref{app:swt-loss-ablation}.

\paragraph{Diffusion in the frame domain.}
We apply the process of Section~\ref{sec:setting} to $x_0^{\mathrm F}=\mathcal W(x_0)$ rather than $x_0$, so the latent is $x_t^{\mathrm F}\in\mathbb R^{(J+1)\times L}$ and the time-domain output is $\hat x_0=\mathcal W^{-1}(\hat x_0^{\mathrm F})$. We use $T=50$ and the quadratic noise schedule of \citet{li_tfcdiff_2025}.

% We apply the process of Section~\ref{sec:setting} to $x_0^{\mathrm F}=\mathcal W(x_0)$ rather than $x_0$, so the diffusion latent is $x_t^{\mathrm F}\in\mathbb R^{(J+1)\times L}$ and the final frame-domain estimate $\hat x_0^{\mathrm F}$ is mapped back to the time domain as $\hat x_0=\mathcal W^{-1}(\hat x_0^{\mathrm F})$. We use $T=50$ and the quadratic noise schedule of \citet{li_tfcdiff_2025}.

%
\subsection{Phase conditioning}
\label{sec:method-phase}

The second inductive bias makes position relative to the underlying physiological cycle explicit to the denoiser. Let $\mathcal{T}(x)=\{e_1<\dots<e_M\}$ denote the detected cycle-event indices of a signal (e.g., R-peaks\footnote{The R peak is the prominent spike in an ECG and serves as a reference point for heartbeat timing.} in an ECG). For a sample $n$ satisfying $e_k \le n < e_{k+1}$, we define its cycle phase, local cycle rate, and soft event mask as
\begin{equation}
\label{eq:phase-defs}
\phi_n = \frac{n-e_k}{e_{k+1}-e_k} \in [0,1),
\qquad
r_n = \frac{f_s}{e_{k+1}-e_k},
\qquad
m_n = \max_{j}\exp\!\left(-\frac{(n-e_j)^2}{2s^2}\right),
\end{equation}
where $f_s$ denotes the sampling rate and $s$ the event width.
These quantities form the dense phase field
\begin{equation}
\label{eq:phase-field}
\Phi = \big(m,\; \phi,\; \sin 2\pi\phi,\; \cos 2\pi\phi,\; r\big) \in \mathbb{R}^{5\times L}.
\end{equation}
Here, $m$ localizes cycle events, $\phi$ gives the relative position within each cycle, and $r$ captures local variation in cycle rate.
The sine and cosine channels embed phase on the circle, avoiding the discontinuity between $\phi=1^{-}$ and $\phi=0$, which would otherwise force the network to learn that $\phi = 0.99$ and $\phi = 0.01$ are adjacent. 

\paragraph{Phase encoder.}
At inference, the clean signal $x_0$ is unavailable, so its phase field must be estimated from the corrupted observation $y$. We use a learned, shift-equivariant convolutional phase encoder $\mathcal{E}_{ph}$ and set $\hat{\Phi}=\mathcal{E}_{ph}(y)$. As illustrated in Figure~\ref{fig:architecture}, the predicted field is transformed into the frame domain, applying $\mathcal W$ independently to each channel, as $\hat{\Phi}^{\mathrm F}=\mathcal{W}(\hat{\Phi})$ and concatenated channel-wise with $x_t^{\mathrm{F}}$ and $y^{\mathrm{F}}$, providing the denoiser with dense per-sample phase information. Since both $\mathcal{E}_{ph}$ and $\mathcal{W}$ are shift-equivariant, it holds that
$\mathcal{W}\mathcal{E}_{ph}(S_\tau y)=S_\tau\mathcal{W}\mathcal{E}_{ph}(y)$. Thus, the dense phase conditioning shifts consistently with the input.  The phase encoder $\mathcal{E}_{ph}$ can be trained through the restoration objective alone. For ECG, where a reliable cycle-event detector is available, we additionally warm-start it using a target phase field $\Phi^\star=\mathcal{D}(x_0)$ computed from the clean signal. Here, $\mathcal{D}$ denotes the analytical construction of Equations~\ref{eq:phase-defs}--\ref{eq:phase-field}. The analytical detector is used only for initialization and not at inference. For the remaining modalities, $\mathcal{E}_{ph}$ is trained without warm-starting. Full warm-start details are given in Appendix~\ref{app:hyperparameters}.\looseness-1

\paragraph{Context encoder.}
In parallel, a context encoder $\mathcal{E}_{\mathrm{ctx}}$ maps $y$ to pooled, shift-invariant context embeddings $c_y=(c_y^{(0)},c_y^{(1)},c_y^{(2)},c_y^{(g)})$, with one embedding for each U-Net resolution and one global embedding. The resolution-specific embeddings modulate the corresponding U-Net features through FiLM~\citep{perez2018film}, while $c_y^{(g)}$ is combined with the diffusion-step embedding. Unlike the phase field, this pathway provides global information about the corrupted observation rather than sample-wise conditioning. Because the phase conditioning is shift-equivariant and the pooled context conditioning is shift-invariant, neither introduces dependence on absolute window position. Architectural details are given in Appendix~\ref{app:hyperparameters}.\looseness-1

\subsection{Measuring cycle structure}
\label{sec:cyclo-index}

The shift-covariant representation and phase conditioning both exploit the cyclostationary structure of Equation~\ref{eq:cyclostationary}. Their usefulness should therefore depend on how strongly cycle phase structures the signal. We want to quantify this dependence from clean data before committing to model training.

\begin{definition}[Cyclostationarity index]
\label{def:pi}
For a process with latent phase $\varphi$,
\begin{equation}
\label{eq:pi}
\Pi \;=\; \frac{\operatorname{Var}_{\varphi}\!\big(\mathbb{E}[x \mid \varphi]\big)}
                {\operatorname{Var}(x)} \;\in\; [0,1]
\end{equation}
is the fraction of signal variance explained by cycle phase.
\end{definition}
$\Pi$ is estimated from clean data by phase-aligned averaging: detect cycle events, resample each cycle to a common phase grid, average to obtain $\hat{\mu}(\varphi)$, and take the ratio of its variance to the total. The estimator, its finite-sample bias and a cheaper autocorrelation proxy are given in Appendix~\ref{app:cyclo-estimator}. Where no reliable cycle event detector exists, we use the autocorrelation proxy instead.

\begin{proposition}[Phase-explained variance]
\label{prop:mmse}
Let $x$ be cyclostationary in the sense of Equation~\ref{eq:cyclostationary}. Then
\begin{equation}
\label{eq:mmse-bound}
\operatorname{Var}(x) - \mathbb{E}_\varphi\!\big[\operatorname{Var}(x \mid \varphi)\big]
\;=\; \operatorname{Var}_\varphi\!\big(\mathbb{E}[x \mid \varphi]\big)
\;=\; \Pi \cdot \operatorname{Var}(x),
\end{equation}
so in the absence of an observation, conditioning on phase reduces the minimum mean-squared error exactly by $\Pi\operatorname{Var}(x)$.\looseness-1
\end{proposition}
Proof in Appendix~\ref{app:proof-mmse}. The identity establishes that the variance phase explains is computable from clean data before any model exists. It does not extend to the case where $y$ is observed (Appendix~\ref{app:proof-mmse}), so we treat $\Pi$ as a measure of how strongly phase structures a signal rather than as a certificate.

\subsection{Learning objective}
\label{sec:method-objective}

The denoiser predicts the injected noise as
$\hat{\epsilon}_t=\epsilon_\theta(x_t^{\mathrm F},y^{\mathrm F},\hat{\Phi}^{\mathrm F},c_y,t)$
and is trained with the standard noise-prediction loss~\citep{ho_denoising_2020} on frame coefficients:
\begin{equation}
\label{eq:loss-diff}
\mathcal{L}_{\mathrm{diff}}
=
\mathbb{E}_{(x_0,y),\,t,\,\epsilon}
\left[
\left\|\epsilon-\hat{\epsilon}_t\right\|_2^2
\right],
\qquad
t\sim\mathcal U\{1,\dots,T\}.
\end{equation}
However, low noise-prediction error does not by itself control time-domain reconstruction error, and
the relation between the two depends on $t$. Substituting Equation~\ref{eq:forward} into the
clean estimate $\hat{x}_0^{\mathrm{F}} = (x_t^{\mathrm{F}} - \sqrt{1-\bar{\alpha}_t}\,\hat{\epsilon}_t)/\sqrt{\bar{\alpha}_t}$ yields
\begin{equation}
\label{eq:error-transfer}
\hat{x}_0^{\mathrm{F}} - x_0^{\mathrm{F}}
= \sqrt{\tfrac{1-\bar{\alpha}_t}{\bar{\alpha}_t}}\;(\epsilon - \hat{\epsilon}_t),
\qquad
\big\|\hat{x}_0 - x_0\big\|_2
\;\le\; A^{-1/2}\sqrt{\tfrac{1-\bar{\alpha}_t}{\bar{\alpha}_t}}\;
\big\|\epsilon - \hat{\epsilon}_t\big\|_2 ,
\end{equation}
where the bound uses $\hat{x}_0 - x_0 = \mathcal{W}^{-1}(\hat{x}_0^{\mathrm{F}} - x_0^{\mathrm{F}})$
together with $\|\mathcal{W}^{-1}\|_2 = A^{-1/2}$, which holds because $\mathcal{W}$ is tight
with $A = B = J+1$ (Equation~\ref{eq:frame-bounds}). The first identity in Equation~\ref{eq:error-transfer} states that $\mathcal{L}_{\mathrm{diff}}$ is a signal-space error weighted by the signal-to-noise ratio $\bar{\alpha}_t/(1-\bar{\alpha}_t)$: uniform weight in noise space is sharply decaying weight in signal space. Under the fixed quadratic schedule of
\citet{li_tfcdiff_2025}, the amplification factor $\sqrt{(1-\bar{\alpha}_t)/\bar{\alpha}_t}$ spans four orders of magnitude.
%, from $\ldots$ at $t=1$ to $161$ at $t=T$, crossing $1$ at $t=22$ and $10$ at $t=40$.
Thus, sampling $t$ uniformly does not weight reconstruction error uniformly:
$\mathcal{L}_{\mathrm{diff}}$ is least sensitive to coefficient error at exactly the steps with large $t$, where that error is amplified most. We counteract this with two terms applied to $\hat{x}_0 = \mathcal{W}^{-1}(\hat{x}_0^{\mathrm{F}})$
at the same sampled $t$:
\begin{equation}
\label{eq:loss-aux}
\mathcal{L}_{\mathrm{rec}} = \mathbb{E}\|\hat{x}_0 - x_0\|_1,\;
\mathcal{L}_{\mathrm{morph}} = \mathbb{E}\,\bar{\alpha}_t\|\nabla \hat{x}_0 - \nabla x_0\|_1,\;
\mathcal{L} = \mathcal{L}_{\mathrm{diff}}
+ \lambda_{\mathrm{rec}}\mathcal{L}_{\mathrm{rec}}
+ \lambda_{\mathrm{morph}}\mathcal{L}_{\mathrm{morph}},
\end{equation}
with $\nabla$ the first-order temporal difference, which penalizes slope error and encourages
sharp transients; $\bar{\alpha}_t$ restricts it to low noise, where slopes are informative. We use $\ell_1$ rather than $\ell_2$ for robustness to the large residuals
that $\hat{x}_0$ exhibits at high $t$. We set $\lambda_{\mathrm{rec}}=0.3$ and $\lambda_{\mathrm{morph}}=0.1$; Appendix~\ref{app:swt-loss-ablation}
reports a sweep over both weights.
% large $t$, where that error is amplified most. We counteract this with two terms applied to $\hat{x}_0 = \mathcal{W}^{-1}(\hat{x}_0^{\mathrm{F}})$
% at the same sampled $t$, which contribute to signal-space error with $t$-independent weight:
% \begin{equation}
% \label{eq:loss-aux}
% \mathcal{L}_{\mathrm{rec}} = \|\hat{x}_0 - x_0\|_1,
% \quad
% \mathcal{L}_{\mathrm{morph}} = \|\nabla \hat{x}_0 - \nabla x_0\|_1,
% \quad
% \mathcal{L} = \mathcal{L}_{\mathrm{diff}}
% + \lambda_{\mathrm{rec}}\mathcal{L}_{\mathrm{rec}}
% + \lambda_{\mathrm{morph}}\mathcal{L}_{\mathrm{morph}},
% \end{equation}
% with $\nabla$ the first-order temporal difference, which penalizes slope error and encourages
% sharp transients. We use $\ell_1$ rather than $\ell_2$ for robustness to the large residuals
% that $\hat{x}_0$ exhibits at high $t$. We set $\lambda_{\mathrm{rec}}=0.3$ and $\lambda_{\mathrm{morph}}=0.1$; Appendix~\ref{app:swt-loss-ablation}
% reports a sweep over both weights.

\subsection{Antithetic reverse estimation}
\label{sec:method-av}

The preceding components shape $p_\theta(x_0\mid y)$; we now fix it and change only how it is sampled. We target the model's conditional mean $\bar{\mu}_\theta(y)=\mathbb{E}_{p_\theta}[x_0\mid y]$, the squared-error-optimal estimate under $p_\theta$, and compare two estimators with the same budget of $KT$ network-function evaluations (NFEs). MC-$K$ averages $K$ independent trajectories, whereas AV-$K$ averages $K/2$ antithetic pairs ($K$ even):
\begin{equation}
\label{eq:mc-av}
\hat{x}_{0,\mathrm{MC}\text{-}K}
=\frac{1}{K}\sum_{k=1}^{K}\hat{x}_0^{(k)},
\qquad\qquad
\hat{x}_{0,\mathrm{AV}\text{-}K}
=\frac{1}{K}\sum_{j=1}^{K/2}\big(\hat{x}_{0,+}^{(j)}+\hat{x}_{0,-}^{(j)}\big).
\end{equation}
Both estimators are unbiased for $\bar{\mu}_\theta(y)$ and differ only in variance. Each antithetic pair starts from $x_{T,+}^{\mathrm{F}}\sim\mathcal{N}(0,I)$ and $x_{T,-}^{\mathrm{F}}=-x_{T,+}^{\mathrm{F}}$. Since $\mathcal{N}(0,I)$ is symmetric, its two members can receive opposite noise at every reverse step without changing their marginal $p_\theta(x_0\mid y)$:
\begin{equation}
\label{eq:av-pair}
x_{t-1,\pm}^{\mathrm{F}}
=
\mu_\theta\!\left(x_{t,\pm}^{\mathrm{F}},y^{\mathrm{F}},\hat{\Phi}^{\mathrm F},c_y,t\right)
\pm \sigma_t z_t,
\qquad
z_t\sim\mathcal{N}(0,I),
\quad
\sigma_t=\sqrt{\tilde{\beta}_t}.
\end{equation}
\begin{proposition}[Error decomposition at matched NFE]
\label{prop:av}
Fix $y$ and write $\bar{\mu}_\theta(y) = \mathbb{E}_{p_\theta}[x_0 \mid y]$. Let $\sigma^2(\tau)$ be the per-trajectory variance of the reconstruction at sample index $\tau$ and $\rho(\tau)$ the correlation between the two members of an antithetic pair. Then the $K$-trajectory estimator satisfies
\begin{equation}
\label{eq:av-variance}
\operatorname{Var}\!\left[\hat{x}_{0,\mathrm{AV}\text{-}K}(\tau)\right]
= \frac{\sigma^{2}(\tau)}{K}\big(1+\rho(\tau)\big),
\end{equation}
and its expected squared error decomposes as
\begin{equation}
\label{eq:mse-decomp}
\mathbb{E}\big\|\hat{x}_{0,K} - x_0\big\|_2^2
= \underbrace{\big\|\bar{\mu}_\theta(y) - x_0\big\|_2^2}_{\text{model bias}}
\;+\; \frac{1+\bar{\rho}}{K}\,
\underbrace{\textstyle\sum_{\tau}\sigma^{2}(\tau)}_{\text{trajectory variance}},
\qquad
\bar{\rho} = \frac{\sum_\tau \rho(\tau)\sigma^2(\tau)}{\sum_\tau \sigma^2(\tau)},
\end{equation}
with $\bar{\rho}=0$ for independent trajectories. Coupling reduces the compute-controllable term by the factor $1+\bar{\rho}$ at equal NFE whenever $\bar{\rho}<0$, and is no worse than independent sampling to $O(\bar{\rho})$ when $\bar{\rho}\approx 0$.\looseness-1
\end{proposition}
Proof in Appendix~\ref{app:proof-av}. At matched NFE, antithetic coupling reduces sampling variance when $\bar{\rho}<0$; Appendix~\ref{app:av-empirical} evaluates this condition and the quality--compute trade-off.

\begin{figure*}[t]
\centering

% ==================== TABLE ====================
\begin{minipage}[t]{0.65\textwidth}
\vspace{0pt}
\vspace{1em}
\centering

\resizebox{\linewidth}{!}{%
\begin{tabular}{llccc}
\toprule
 & Model & $\Delta$SNR $\uparrow$ & PRD [\%] $\downarrow$ & CC $\uparrow$ \\
\midrule
\multirow{3}{*}{PPG} & DDAE & $8.97 \pm 0.14$ & $36.25 \pm 0.38$ & $0.91 \pm 0.00$ \\
 & \textbf{\method-MC-10} & \secondbest{$12.99 \pm 0.11$} & \secondbest{$24.64 \pm 0.30$} & \secondbest{$0.96 \pm 0.00$} \\
 & \textbf{\method-AV-10} & \best{$13.23 \pm 0.10$} & \best{$24.03 \pm 0.27$} & \best{$0.96 \pm 0.00$} \\
\midrule
\multirow{5}{*}{ECG} & TCDAE & $12.49 \pm 0.65$ & $33.24 \pm 2.08$ & $0.95 \pm 0.00$ \\
 & TFCDiff-10 & $13.35 \pm 0.13$ & $30.27 \pm 0.45$ & $0.94 \pm 0.00$ \\
 & DeScoD-10 & $13.85 \pm 0.04$ & $28.76 \pm 0.15$ & $0.95 \pm 0.00$ \\
 & \textbf{\method-MC-10} & \secondbest{$15.58 \pm 0.12$} & \secondbest{$23.81 \pm 0.27$} & \secondbest{$0.96 \pm 0.00$} \\
 & \textbf{\method-AV-10} & \best{$15.87 \pm 0.12$} & \best{$23.08 \pm 0.25$} & \best{$0.97 \pm 0.00$} \\
\midrule
\multirow{4}{*}{EMG} & FCN & $19.28 \pm 0.20$ & $26.51 \pm 0.53$ & $0.96 \pm 0.00$ \\
  & SDEMG & $21.04 \pm 0.30$ & $21.91 \pm 0.75$ & $0.97 \pm 0.00$ \\
 & \textbf{\method-MC-10} & \secondbest{$22.41 \pm 0.20$} & \secondbest{$18.80 \pm 0.57$} & \secondbest{$0.98 \pm 0.00$} \\
 & \textbf{\method-AV-10} & \best{$22.70 \pm 0.19$} & \best{$18.23 \pm 0.55$} & \best{$0.98 \pm 0.00$} \\
\midrule
\multirow{4}{*}{EEG} & EEGIFNet & $15.05 \pm 0.04$ & \best{$19.96 \pm 0.10$} & \best{$0.97 \pm 0.00$} \\
 & EEGDfus & $15.03 \pm 0.21$ & $22.18 \pm 0.79$ & $0.96 \pm 0.00$ \\
 & \textbf{\method-MC-10} & \secondbest{$15.31 \pm 0.07$} & $20.84 \pm 0.19$ & $0.97 \pm 0.00$ \\
 & \textbf{\method-AV-10} & \best{$15.39 \pm 0.07$} & \secondbest{$20.72 \pm 0.19$} & \secondbest{$0.97 \pm 0.00$} \\
\bottomrule
\end{tabular}%
}
\vspace{0.75em}
\captionof{table}{
Restoration across four modalities against each modality's own baselines (3 training seeds).
Values are mean $\pm$ standard deviation. \textbf{Bold} marks best, \underline{underline} second best, within each modality. Modalities are ordered by phase-shuffle margin.
% Input conditions: PPG $9.0$\,dB (PRD $154.51$, CC $0.68$ uncorrected); ECG $-2.0$\,dB ($152.86$, $0.61$); EMG $-7.0$\,dB ($256.33$, $0.43$); EEG $0.0$\,dB ($106.76$, $0.69$).
}
\label{tab:main-modalities}

\end{minipage}
\hfill
% ==================== FIGURE ====================
\begin{minipage}[t]{0.33\textwidth}
\vspace{0pt}
\centering

\includegraphics[width=\linewidth]{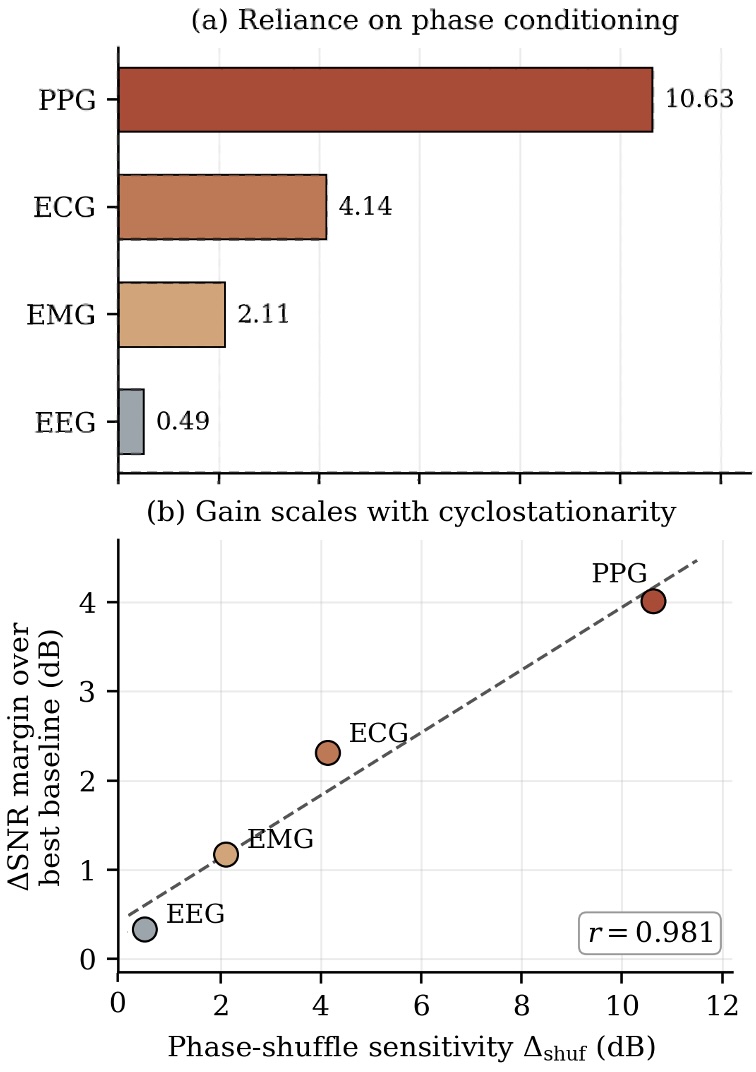}
\vspace{-0.48cm}
\captionof{figure}{
\textbf{(a)} $\Delta$SNR drop under phase shuffling.
\textbf{(b)} Phase-shuffle sensitivity versus the margin over the best baseline.
}
\label{fig:scaling}

\end{minipage}
\vspace{-0.5cm}
\end{figure*}

\section{Experimental Setup}
\label{sec:setup}
\paragraph{Physiological signals and datasets.} We evaluate datasets of four physiological modalities with different structures. First, for PPG we use PPG-DaLiA \citep{attila_reiss_ppg-dalia_2019}, wrist blood-volume pulse (BVP) at 64\,Hz corrupted with Noise Stress Test Database (NSTDB)\citep{moody_mit-bih_1992,goldberger_physiobank_2000} motion and baseline artifacts. Second, ECG uses PTB-XL \citep{PhysioNet-ptb-xl-1.0.3,wagner_ptb-xl_2020} for training and in-domain test, with MIT-BIH Arrhythmia \citep{moody_impact_2001} and QTDB \citep{laguna_qt_1997} held out to evaluate distribution shift, corrupted from the MIT-BIH NSTDB. Next, EMG uses NinaPro DB2 \citep{atzori_electromyography_2014} contaminated with ECG interference from Normal Sinus Rhythm Database (NSRDB) following the SDEMG protocol \citep{liu_sdemg_2024}. Finally, EEG uses EEGdenoiseNet \citep{zhang_eegdenoisenet_2021} in the EEG--EOG setting following the EEGDfus protocol \citep{huang_eegdfus_2025}. To test distribution shift, the PPG, EMG and EEG models are also evaluated on one held-out dataset each: pulse-oximeter PPG from critically ill patients in BIDMC \citep{pimentel_toward_2017}, sEMG from trans-radial amputees in NinaPro DB3 \citep{atzori_electromyography_2014}, and EEG from the PhysioNet Motor Movement/Imagery Database (EEGMMIDB) \citep{schalk_bci2000_2004}, each corrupted with the test protocol of its training dataset. Preprocessing follows each benchmark's established protocol; details in Appendix~\ref{app:preprocessing}.\looseness-1

\paragraph{Baselines.} Each modality is compared against its strongest published restoration methods, retrained on identical splits and preprocessing: DDAE \citep{lai_enhanced_2025} for PPG; TCDAE \citep{chen_elimination_2024}, DeScoD-ECG \citep{li_descod-ecg_2024} and TFCDiff \citep{li_tfcdiff_2025} for ECG; SDEMG \citep{liu_sdemg_2024} and FCN \citep{wang_ecg_2023} for EMG; EEGDfus \citep{huang_eegdfus_2025} and EEGIFNet \citep{cui_dual-branch_2024} for EEG. Other common baselines (FIR, SWT shrinkage) are reported in Appendix~\ref{app:full-ecg}. Stochastic baselines use the sampling procedure specified in their own publications. Every baseline figure was produced under our protocol. Appendix~\ref{app:baseline-details} provides implementation and adaptation details.\looseness-1

\paragraph{Metrics and protocol.} 
We report $\Delta\mathrm{SNR}$ (dB), percentage root-mean-square difference (PRD, \%) and Pearson correlation ($\mathrm{CC}$) to evaluate restoration performance, computed per signal and averaged, with 95\%
bootstrap confidence intervals (CI) over 1{,}000 resamples. All results use a single training seed, except Table~\ref{tab:main-modalities}, which is trained and evaluated across three seeds and reported as mean $\pm$ standard deviation (exact seeds in Appendix~\ref{app:hyperparameters}). Benchmark-specific metrics are given in Appendix~\ref{app:metrics}. Paired comparisons against each baseline use one-sided Wilcoxon signed-rank tests with Holm--Bonferroni correction (Appendix~\ref{app:statistics}). Finally, sampling variants are labeled MC-$K$ for Monte Carlo averaging over $K$ independent trajectories and AV-$K$ for $K/2$ antithetic pairs; both use $K$ reverse trajectories and hence $KT$ NFEs. Additional experimental details are included in Appendix~\ref{app:extended-setup}.\looseness-1

\section{Results}
\label{sec:results}
We evaluate \method{} across four physiological modalities and answer:
\textit{(i)~Restoration,} does it improve the quality and preserve downstream utility?
\textit{(ii) Phase conditioning,} when does phase help, and can we predict its benefit?
\textit{(iii) Components,} which parts drive the gains? \textit{(iv) Efficiency,} can antithetic sampling improve the inference quality--compute trade-off? Further results in Appendix~\ref{app:supp-results}.\looseness-1

\paragraph{Restoration performance.}
Table~\ref{tab:main-modalities} reports successful restoration performance for all four modalities against modality-specific baselines, and Figures~\ref{fig:ecg-qualitative-high}--\ref{fig:all-modalities-qualitative} (Appendix~\ref{app:qualitative}) show representative reconstructions. \method{}-AV-10 achieves the best $\Delta\mathrm{SNR}$ on all four modalities and the best PRD and CC on three. On EEG, EEGIFNet retains the best PRD and CC: these metrics are averaged on a linear rather than a decibel scale, so poorly restored segments weigh more heavily, and EEGIFNet's quality is more uniform across segments. On ECG, \method{} matches or exceeds every baseline, classical filters included, on the metrics of prior work (Table~\ref{tab:denoise-ecg-combined}); all paired comparisons are significant after Holm--Bonferroni correction (Table~\ref{tab:paired-statistical-comparison}), and the improvement shifts the whole per-sample distribution (Figure~\ref{fig:persample}). Its advantage there widens at higher noise levels (Tables~\ref{tab:denoise-ptbxl-bins-combined}--\ref{tab:denoise-mitbih-bins-combined}) and holds on held-out datasets without retraining (Figure~\ref{fig:cross-dataset}), highlighting strong generalization capabilities. On MIT-BIH, which is rich in arrhythmias, the other diffusion models lose their edge over the deterministic TCDAE while \method{} keeps a clear lead, consistent with a phase field defined beat by beat rather than by a fixed rhythm. Figures~\ref{fig:cross-modality-bins} and~\ref{fig:cross-modality-original} stratify the other modalities by noise level, with the main metrics and with each benchmark's own (Table~\ref{tab:metrics}), respectively. Under dataset shift without retraining, \method{} keeps its lead over the strongest baseline on PPG, ECG and EMG but falls behind EEGIFNet on EEG, the modality where its in-domain margin is also smallest (Table~\ref{tab:crossdataset-modalities}, Appendix~\ref{app:crossdataset}).\looseness-1
 
\begin{wraptable}{l}{0.4\textwidth}
\vspace{-14pt}
\centering
\caption{Downstream utility across modalities. Extended results in Table~\ref{tab:placeholder_downstream}.}
\label{tab:downstream}
\scriptsize
\setlength{\tabcolsep}{3pt}
\renewcommand{\arraystretch}{1.0}

\begin{tabular}{@{}l l r r@{}}
\toprule
& Task & Baseline & \method{} \\
\midrule
PPG & HR MAE $\downarrow$                & 5.947 & \textbf{5.155} \\
ECG & PTB-XL AUROC $\uparrow$            & 0.840  & \textbf{0.848} \\
ECG & MIT-BIH F1 $\uparrow$              & 0.543  & \textbf{0.630} \\
EMG & Activation bal.\ acc.\ $\uparrow$  & 0.749  & \textbf{0.750} \\
EEG & Band-power error $\downarrow$      & 4.56  & \textbf{3.31} \\
\bottomrule
\end{tabular}

\vspace{-10pt}
\end{wraptable}

The improvements also carry over downstream: \method{} matches or exceeds the strongest restoration baseline on every task across the four modalities (Table~\ref{tab:downstream}), and on ECG it recovers most of the macro-AUROC and all of the abnormal-beat sensitivity lost to noise, with the best F1 among restoration methods (Tables~\ref{tab:ptbxl-macro-classifier}--\ref{tab:ptbxl-macro-classifier-appendix}). Across modalities, the restoration margin over the strongest baseline is largest on PPG and smallest on EEG, an ordering we relate to cycle structure next.\looseness=-1

\vspace*{-0.2cm}
\paragraph{Phase conditioning and cyclostationarity are connected.}

\begin{wrapfigure}{r}{0.49\textwidth}
\vspace{-11pt}
\centering
\includegraphics[width=\linewidth]{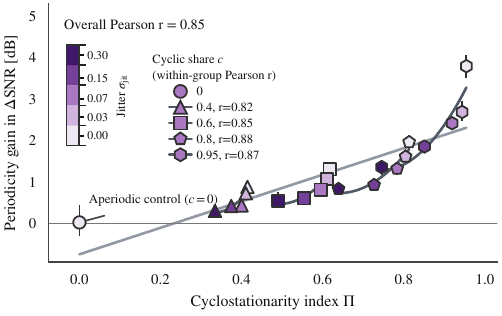}
\vspace{-15pt}
\caption{Synthetic sweep: $\Delta$SNR gain over the no-periodicity
ablation versus the cyclostationarity index $\Pi$.
Further details in Appendix~\ref{app:jitter}.}
\label{fig:jitter_sweep}
\vspace{-20pt}
\end{wrapfigure}

Phase conditioning improves restoration more when cycle structure is stronger. To measure how much the denoiser relies on phase, we permute the phase fields across the samples of each test batch while keeping the inputs unchanged. This lowers $\Delta\mathrm{SNR}$ in every modality, most on PPG and least on EEG (Figures~\ref{fig:scaling}a and~\ref{fig:shuffle}), and the drop tracks the margin of \method{} over the strongest baseline, with $r=0.981$ across the four modalities (Figure~\ref{fig:scaling}b). 
A controlled synthetic study, which varies the cyclic share of variance and the cycle-length jitter with all else fixed, supports this: the $\Delta\mathrm{SNR}$ improvement from phase conditioning increases with the cyclostationarity index $\Pi$ of Definition~\ref{def:pi} ($r=0.85$; Figure~\ref{fig:jitter_sweep}, Appendix~\ref{app:jitter}). Table~\ref{tab:cyclicity-index} reports the proxy $\Pi_{AC}$ for all nine datasets (Appendix~\ref{app:cyclicity}), ordering them as the ablations (Table~\ref{tab:placeholder_ablation_modalities}) suggest, in which removing phase conditioning altogether costs most on the two cardiac modalities, PPG and ECG, and little or nothing on EMG and EEG. Unlike the shuffle (Appendix~\ref{app:shuffle}), which probes how much a trained model relies on phase, this ablation retrains the model without it (Appendix~\ref{app:ablation-modalities}). Learning the phase is key. On ECG, the learned encoder improves every metric over the phase-free model, whereas the analytical QRS detector of Table~\ref{tab:analytical-encoder} is not enough: applied to the noisy input, it performs on par with omitting phase (Table~\ref{tab:ablation}). The detector's errors propagate into the phase field, while the encoder learns to infer the clean signal's phase from the same input.\looseness=-1

\begin{table}[h!]
\vspace{-6pt}
\centering
\setlength{\tabcolsep}{4pt}
\renewcommand{\arraystretch}{0.78}
\setlength{\aboverulesep}{0.2ex}
\setlength{\belowrulesep}{0.2ex}
\resizebox{0.75\linewidth}{!}{%
\begin{tabular}{ccccc|ccc}
\toprule
Frame & & & Analytical & Auxiliary & & & \\
(SWT) & Phase & Context & phase & loss & $\Delta$SNR $\uparrow$ & PRD [\%] $\downarrow$ & CC $\uparrow$ \\
\midrule
\cmark & -- & -- & -- & \cmark & $15.05 \pm 0.21$ & $25.51 \pm 0.60$ & $0.96 \pm 0.00$ \\
\cmark & -- & \cmark & -- & \cmark & $15.36 \pm 0.21$ & $24.54 \pm 0.55$ & $0.96 \pm 0.00$ \\
\cmark & \cmark & -- & -- & \cmark & $15.28 \pm 0.20$ & $25.35 \pm 0.65$ & $0.96 \pm 0.00$ \\
\cmark & \cmark & \cmark & \cmark & \cmark & $15.34 \pm 0.21$ & $24.50 \pm 0.54$ & $0.96 \pm 0.00$ \\
\cmark & \cmark & \cmark & -- & -- & \secondbest{$15.52 \pm 0.21$} & $24.01 \pm 0.53$ & $0.96 \pm 0.00$ \\[2pt]
-- & \cmark & \cmark & -- & \cmark & $15.48 \pm 0.21$ & \secondbest{$23.97 \pm 0.53$} & \secondbest{$0.96 \pm 0.00$} \\[2pt]
\midrule
\cmark & \cmark & \cmark & -- & \cmark & \best{$16.00 \pm 0.21$} & \best{$22.68 \pm 0.50$} & \best{$0.97 \pm 0.00$} \\
\bottomrule
\end{tabular}%
}
\vspace{-2pt}
\caption{Component ablation on ECG (PTB-XL, SNR $-2.0$\,dB,
one antithetic pair, single seed; mean $\pm$ 95\% CI). Phase from a classical
detector performs similarly to no phase; the learned encoder improves on both.}
\label{tab:ablation}
\vspace{-6pt}
\end{table}

\paragraph{Further Analyses.}
In Table~\ref{tab:placeholder_ablation_modalities}, we remove one component at a time in each modality; Table~\ref{tab:ablation} details the ECG case, where the full model is best on every metric. Each component improves $\Delta\mathrm{SNR}$ in three of the four modalities. Phase and context conditioning are complementary: on ECG, each improves the model more when the other is present. The time-domain auxiliary losses (Equation~\ref{eq:loss-aux}) help, as the weighting argument of Section~\ref{sec:method-objective} predicts, and most of all on EEG, which uses its own wavelet and loss weight (Table~\ref{tab:hyperparameters}). The shift-covariant frame contributes comparably on ECG, EMG and EEG, whereas on PPG, whose restoration relies most on phase, removing the frame or the context slightly improves $\Delta\mathrm{SNR}$. On ECG, the choice of wavelet and decomposition level matters more than the loss weights, and four levels (Figure~\ref{fig:swt-example}) are best for every wavelet tested (Tables~\ref{tab:swt-ablation} and~\ref{tab:loss-ablation}).\looseness=-1

\paragraph{Sampling efficiency.}

\begin{wrapfigure}{r}{0.44\textwidth}
\vspace{-14pt}
\centering
\includegraphics[width=\linewidth]{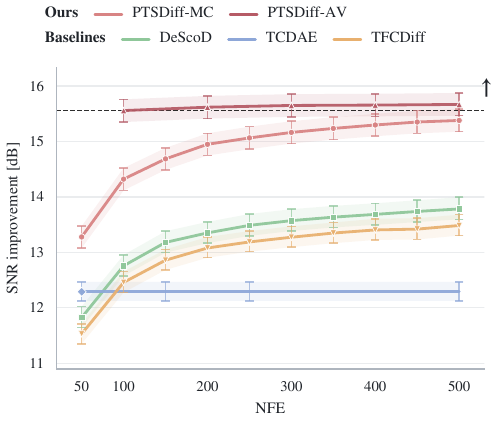}
\vspace{-15pt}
\caption{
Sampling efficiency on ECG. A single antithetic pair (AV-2)
outperforms ten independent trajectories (MC-10) with $5\times$
fewer NFEs. Wall-clock time decreases similarly. Extended results in Figure~\ref{fig:shots}.
}
\label{fig:shots_small}
\vspace{-8pt}
\end{wrapfigure}

Antithetic coupling helps at no extra cost: at equal NFE, \method{}-AV-10 improves $\Delta\mathrm{SNR}$ and PRD over \method{}-MC-10 in every modality (Table~\ref{tab:main-modalities}), consistent with Proposition~\ref{prop:av} for negatively correlated pairs (measured directly in Figure~\ref{fig:rho_measurement}). On ECG, a single antithetic pair already outperforms ten independent trajectories with five times fewer NFEs (Figure~\ref{fig:shots_small}), and further pairs add little because the remaining error is then dominated by the model-bias term of Equation~\ref{eq:mse-decomp}, which does not shrink with $K$ (Figure~\ref{fig:shots}). The benefit transfers to other diffusion restorers: applied to DeScoD-ECG and TFCDiff without retraining, one antithetic pair again matches or exceeds ten independent trajectories at a fifth of the wall-clock time, and \method{} is the fastest of the three at equal NFE (Table~\ref{tab:placeholder_av_transfer}). Because averaging approximates the conditional mean, these improvements concern distortion rather than the realism of individual samples (Appendix~\ref{app:proof-av}).\looseness=-1

\section{Conclusion}
\label{sec:discussion}

We introduced \method{}, a diffusion-based framework for restoring cyclostationary physiological signals, leveraging the additional structure present in the data that prior work disregards. The framework combines a shift-covariant frame representation with a learned dense phase field, a training-free cyclostationarity index for estimating when phase conditioning is likely to be useful, and antithetic reverse sampling for improved efficient point estimation. 
%The same approach applies across ECG, PPG, EMG, and EEG without changing its core components. 
Our experiments show that the benefit of phase conditioning increases with the cyclostationary structure of the signal: the cyclostationarity index separates the cardiac modalities from others of less cyclic nature, and its autocorrelation proxy, averaged over datasets, orders the modalities as the ablation gains do. Ablations further demonstrate that this benefit depends on learning a noise-robust phase representation as analytical phase estimated from the corrupted input provides no measurable improvement, whereas a learned encoder captures the required information to act as a directed bias. Finally, antithetic sampling improves distortion-based restoration metrics by better approximating the posterior mean, without modifying the architecture or diffusion schedule in an efficient manner. We show how the results of this cyclostationary-informed approach outperforms prior restoration baselines across four real world modalities and datasets.\looseness-1

\paragraph{Limitations and future work.}
%The relationship between cyclostationarity and performance is evaluated across only four modalities, which supports the observed ordering but is insufficient to establish its functional form. Evaluating additional datasets, corruption types, and controlled levels of cycle variability would provide a stronger test of this relationship. 
% Moreover, the experiments use synthetic corruption added to recordings treated as clean; evaluation on naturally corrupted signals is an important next step. The phase encoder is initialized using modality-specific detectors that may differ in quality, motivating systematic initialization ablations across modalities (\ph{\texttt{placeholder\_ablation\_modalities}}). Extending the current single-channel framework to multi-lead ECG and multi-channel EEG could exploit correlations between channels. Finally, diffusion inference remains more expensive than single-pass restoration, requiring $2T$ NFEs for one antithetic pair; reducing this cost through accelerated sampling or distillation is a further direction for future work.

The relationship between cyclostationarity and restoration benefit should be further tested across broader datasets and naturally occurring artifacts. Similarly, the insights focus on single-channel settings and extending the current framework to multi-lead ECG and multi-channel EEG could exploit correlations between channels.
% The current framework focuses on insights on single-channel studies, 
% As seen in the results, phase estimation may depend on the quality of the available cycle structure, hence our solution is restricted to .
Like prior diffusion-based methods, our approach remains more expensive than single-pass restoration despite antithetic sampling.
More broadly, this work opens a direction for designing time-series models around measurable cyclostationary structure. The proposed index provides a way to identify when explicit phase information is likely to be useful, while learned phase representations offer a mechanism for exploiting this structure when it is not directly observable from corrupted data. Extending these ideas to multichannel, irregular, and more weakly cyclic signals could enable adaptive inductive biases beyond physiological restoration, including further time-series reconstruction and generative modeling settings.\looseness-1

\newpage

\subsubsection*{Reproducibility statement}

The method is specified in Section~\ref{sec:method}, with architectural details, training hyperparameters, random seeds and the compute environment documented in Appendix~\ref{app:hyperparameters} and Table~\ref{tab:hyperparameters}. All clean-signal datasets \citep{PhysioNet-ptb-xl-1.0.3,moody_impact_2001,laguna_qt_1997,attila_reiss_ppg-dalia_2019,atzori_electromyography_2014,zhang_eegdenoisenet_2021,pimentel_toward_2017,schalk_bci2000_2004} and noise sources \citep{moody_mit-bih_1992,goldberger_physiobank_2000} are public; the corruption protocols, splits and preprocessing needed to regenerate every evaluation set are given in Appendix~\ref{app:preprocessing}, including the subject- and record-level disjointness constraints. Baselines were retrained under identical data protocols rather than quoted from their publications (Section~\ref{sec:setup}); porting, reimplementation and adaptation details are in Appendix~\ref{app:baseline-details}. Metric definitions are in Appendix~\ref{app:metrics}, the statistical procedure in Appendix~\ref{app:statistics}, and the downstream task setups in Appendix~\ref{app:downstream}. Proofs of Propositions~\ref{prop:mmse} and~\ref{prop:av} are in Appendices~\ref{app:proof-mmse} and~\ref{app:proof-av}; the estimator for $\Pi$ is in Appendix~\ref{app:cyclo-estimator} and the synthetic sweep protocol in Appendix~\ref{app:jitter}. %Anonymized source code is provided at \url{https://to_be_released_upon_acceptance}.

\subsubsection*{Ethics statement}

This work uses only publicly available, de-identified physiological recordings released for research under their respective licenses. No new human-subject data was collected and no re-identification was attempted; as a secondary analysis of public, de-identified data, the work did not require additional ethical review. The principal risk is clinical: a restoration model can produce a plausible waveform that is diagnostically wrong, and a clean-looking output may command more confidence than a visibly noisy one. Two aspects of our results bear on this. First, distortion metrics do not establish diagnostic fidelity, which is why we evaluate downstream tasks (Table~\ref{tab:downstream}, Appendix~\ref{app:downstream}). Second, averaging reverse trajectories approximates the conditional mean, which improves distortion but not the realism of individual samples (Appendix~\ref{app:proof-av}), and can therefore suppress rare morphology that may carry diagnostic weight. Restoration of this kind should not be deployed in a diagnostic pathway without task-level validation on the population of interest, and outputs should not be presented to clinicians as recordings. Benchmark corruption is also synthetic and drawn from a small set of artifact sources, so robustness to real ambulatory artifacts across diverse populations and devices remains unverified.

\subsubsection*{AI use statement}

In this work, we used generative AI tools to help implement our methods and experiments in code.
We did not use them to generate synthetic datasets, clean or reformat datasets, help develop theoretical models or conceptual frameworks, formulate mathematical claims, provide critical ingredients for proving mathematical claims, assist in the writing of proofs, propose or refine hypotheses, design or provide feedback on research methodology or experiments, or interpret results.
Translation and qualitative or thematic data analysis do not apply to this work.
Additionally, we used generative AI tools to draft and edit parts of the paper text, including condensing captions and laying out tables and figures in \LaTeX{}, but not to search for, identify, or summarize related literature.
We reviewed all AI-assisted output: code was checked and tested before any experiment used it, and text was verified against our methods and results.
We take full responsibility for everything in this paper, including any text, claims, and artifacts created with generative AI assistance.

 \subsubsection*{Acknowledgments}

 SRC is supported by the Department of Computer Science at ETH Zurich and reports equity, intellectual property and consulting with Physcade Inc.---no disclosures are conflicting with or related to this work. JSG is supported by the StimuLoop grant \#1-007811-002 and the Vontobel Foundation. SL is supported by the Swiss State Secretariat for Education, Research, and Innovation (SERI) under contract number MB22.00047. Computational data analysis was performed at Leonhard Med,\footnote{\footnotesize\url{https://sis.id.ethz.ch/services/sensitiveresearchdata/}} a secure trusted research environment at ETH Zurich.

\newpage

\bibliography{iclr2027_conference}
\bibliographystyle{iclr2027_conference}

\appendix
\clearpage

\appendix
\clearpage

\section*{Appendix Contents}

% Treat the appendix as one local TOC region.
\etocsetlocaltop.toc{part}
\etocsetnexttocdepth{subsection}

\begingroup
\small
\setlength{\parindent}{0pt}

% Suppress etoc's default "Contents" heading.
\etocsettocstyle{}{}

% Top-level appendix sections: A, B, C, ...
\etocsetstyle{section}
  {\par}
  {}
  {%
    \noindent
    \textbf{\etocnumber\hspace{0.7em}\etocname}
    \nobreak\leaders\hbox{\normalfont.\kern0.35em}\hfill
    \nobreak\hbox{\etocpage}\par
  }
  {}

% Appendix subsections: A.1, A.2, ...
\etocsetstyle{subsection}
  {}
  {}
  {%
    \noindent\hspace{1.4em}
    \etocnumber\hspace{0.6em}\etocname
    \nobreak\leaders\hbox{\normalfont.\kern0.35em}\hfill
    \nobreak\hbox{\etocpage}\par
  }
  {}

\localtableofcontents

\endgroup

\vspace{1.2em}

% -------------------------------------------------------------------------
% Appendix starts here
% -------------------------------------------------------------------------

\clearpage
\section{Extended Related Work}
\label{app:related-extended}

Section~\ref{sec:related} organizes prior work by idea. This appendix gives the per-modality detail compressed out of the main text.

\paragraph{ECG.} Classical restoration spans Kalman filtering \citep{sayadi_ecg_2008}, empirical mode decomposition \citep{blanco-velasco_ecg_2008}, adaptive filtering and wavelet shrinkage \citep{gao_denoising_2010}, with dedicated treatments of baseline wander \citep{romero_baseline_2019} and powerline interference \citep{chatterjee_review_2020}. Learned approaches progress from convolutional autoencoders \citep{chiang_noise_2019} through recurrent models \citep{antczak_deep_2019,sherstinsky_fundamentals_2020} to transformer autoencoders \citep{chen_elimination_2024,vaswani_attention_2017} and adversarial formulations \citep{singh_new_2021}. Generative restoration begins with DeScoD-ECG \citep{li_descod-ecg_2024}; time--frequency variants include MECG-E \citep{hung_mecg-e_2024} and TFCDiff \citep{li_tfcdiff_2025}. Wavelet choice for ECG has its own literature \citep{addison_wavelet_2005,singh_optimal_2006,kumar_stationary_2021, ercelebi_electrocardiogram_2004,lahmiri_comparative_2014}, and the frequency content justifying our band assignment is documented by \citet{xie_computational_2020}.

\paragraph{PPG.} Motion artifacts and baseline drift dominate \citep{ismail_heart_2021}, addressed classically by wavelet thresholding \citep{lee_reduction_2003}, periodic moving-average filtering \citep{lee_periodic_2007} and EMD, and more recently by dilated denoising autoencoders \citep{lai_enhanced_2025}. The cardiac origin of PPG pulse structure \citep{nitzan_9_2022} places it at the high-$\Pi$ end of the spectrum, and distortion of the pulse contour is the principal failure mode of aggressive filtering \citep{watanabe_beyond_2025}.

\paragraph{EEG.} Preprocessing conventionally removes ocular, muscular, cardiac and instrumental artifacts by regression, ICA, PCA or EMD \citep{jiang_removal_2019}. Learned approaches include dual-branch convolutional--recurrent fusion \citep{cui_dual-branch_2024} and conditional diffusion \citep{huang_eegdfus_2025}. EEG carries oscillatory rather than event-locked structure \citep{brenner_periodic_1990}, giving it low $\Pi$.

\paragraph{EMG.} Classical work targets motion artifacts, cross-talk and high-frequency noise via filtering, regression and wavelet-ICA/EMD \citep{boyer_reducing_2023}. Fully convolutional models remove ECG interference effectively \citep{wang_ecg_2023}, and SDEMG \citep{liu_sdemg_2024} applies score-based diffusion with explicit burst modeling. Motor-unit firing gives EMG burst-locked rather than waveform-locked structure \citep{khorrami_chokami_identification_2021}, placing it between ECG and EEG.

\paragraph{Related restoration and representation work.} Broader context includes DCT-domain diffusion \citep{ning_dctdiff_2025}, learning-based reconstruction under structured corruption \citep{zhu_learning_2021}, imputation for physiological streams \citep{alcaraz_diffusion-based_2023,xu_pulseimpute_2023,jenkins_improving_2023}, and comparative evaluations of restoration pipelines \citep{adam_comparative_2025,basso_reduction_2025,bedin_leveraging_2024}.

\newpage
\section{Extended Methods}
\label{app:proofs}

\subsection{Proof of Proposition~\ref{prop:mmse} (phase-explained variance)}
\label{app:proof-mmse}

By the law of total variance applied to the conditioning on $\varphi$,
\begin{equation}
\operatorname{Var}(x)
= \underbrace{\operatorname{Var}_{\varphi}\!\big(\mathbb{E}[x\mid\varphi]\big)}_{=\,\Pi\,\operatorname{Var}(x)}
+ \; \mathbb{E}_{\varphi}\!\big[\operatorname{Var}(x\mid\varphi)\big],
\end{equation}
which rearranges to Equation~\ref{eq:mmse-bound}. Since $\mathrm{MMSE}(x) = \operatorname{Var}(x)$ and $\mathrm{MMSE}(x\mid\varphi) = \mathbb{E}_\varphi[\operatorname{Var}(x\mid\varphi)]$, the identity states that knowing $\varphi$ reduces the minimum mean-squared error by exactly $\Pi\operatorname{Var}(x)$: $\Pi$ is the share of total variance that phase explains, and $(1-\Pi)\operatorname{Var}(x)$ is the share it cannot. \hfill$\square$

\paragraph{Scope.} Equation~\ref{eq:mmse-bound} concerns the unconditional problem and does not extend to the case where $y$ is observed. Applying the law of total variance conditionally on $y$ gives
\begin{equation}
\label{eq:cond-reduction}
\mathrm{MMSE}(x\mid y) - \mathrm{MMSE}(x\mid y,\varphi)
= \mathbb{E}_y\!\big[\operatorname{Var}_{\varphi\mid y}\!\big(\mathbb{E}[x\mid y,\varphi]\big)\big],
\end{equation}
which $\Pi\operatorname{Var}(x)$ does not bound. Equation~\ref{eq:cyclostationary} permits $C(\varphi_n,\varphi_m)$ to depend on phase, whereas $\Pi$ captures only the first-moment dependence, so a process whose phase modulates conditional variance but not conditional mean has $\Pi=0$ while Equation~\ref{eq:cond-reduction} is strictly positive: knowing $\varphi$ tells the estimator which conditional variance to assume. A phase prior can therefore help more than $\Pi$ suggests. We use $\Pi$ as a measure of how strongly phase structures a signal, and the phase-conditioning analysis in Section~\ref{sec:results} tests empirically whether realized gains track it.

\subsection{Proof of Proposition~\ref{prop:av} (matched-NFE variance)}
\label{app:proof-av}

Fix the observation $y$ and a sample index $\tau$; all expectations are over the reverse-process noise with $y$ held fixed. Let $\hat{x}_{0,+}$ and $\hat{x}_{0,-}$ be the two reconstructions of an antithetic pair generated by Equation~\ref{eq:av-pair}. Because $z_t$ and $-z_t$ are identically distributed, and the two chains apply the same deterministic map $\mu_\theta$ with the same conditioning, the two members are exchangeable:
\begin{equation}
\mathbb{E}[\hat{x}_{0,+}(\tau)] = \mathbb{E}[\hat{x}_{0,-}(\tau)] = \mu(\tau),
\qquad
\operatorname{Var}[\hat{x}_{0,+}(\tau)] = \operatorname{Var}[\hat{x}_{0,-}(\tau)] = \sigma^2(\tau).
\end{equation}
The pair estimator $\hat{x}_{0,\mathrm{AV}} = \frac{1}{2}(\hat{x}_{0,+}+\hat{x}_{0,-})$ is therefore unbiased for $\mu(\tau)$, the same quantity estimated by independent averaging. Writing $\rho(\tau)$ for the correlation between the pair members,
\begin{equation}
\operatorname{Var}[\hat{x}_{0,\mathrm{AV}}(\tau)]
= \tfrac{1}{4}\big(\sigma^2(\tau) + \sigma^2(\tau) + 2\rho(\tau)\sigma^2(\tau)\big)
= \tfrac{1}{2}\sigma^2(\tau)\big(1+\rho(\tau)\big).
\end{equation}
The $K$-trajectory estimator averages $K/2$ pairs drawn independently, so
\begin{equation}
\operatorname{Var}\big[\hat{x}_{0,\mathrm{AV}\text{-}K}(\tau)\big]
= \frac{2}{K}\cdot\frac{\sigma^2(\tau)\big(1+\rho(\tau)\big)}{2}
= \frac{\sigma^2(\tau)}{K}\big(1+\rho(\tau)\big),
\end{equation}
establishing Equation~\ref{eq:av-variance}. Each pair costs two reverse chains, so $K$ antithetic trajectories and $K$ independent trajectories both require $KT$ network evaluations; the independent estimator has variance $\sigma^2(\tau)/K$. The ratio is exactly $1+\rho(\tau)$, which is $<1$ iff $\rho(\tau)<0$.

For Equation~\ref{eq:mse-decomp}, the estimator is unbiased for $\bar{\mu}_\theta(y)$ at every index, so the standard bias--variance split applies coordinatewise,
\begin{equation}
\mathbb{E}\big\|\hat{x}_{0,K}-x_0\big\|_2^2
= \sum_\tau \Big(\big(\bar{\mu}_\theta(y)(\tau)-x_0(\tau)\big)^2
+ \operatorname{Var}\big[\hat{x}_{0,K}(\tau)\big]\Big).
\end{equation}
Substituting Equation~\ref{eq:av-variance} into the second term and factoring out $K^{-1}$ gives $K^{-1}\sum_\tau \sigma^2(\tau)(1+\rho(\tau)) = K^{-1}(1+\bar{\rho})\sum_\tau \sigma^2(\tau)$ by the definition of the variance-weighted mean correlation $\bar{\rho}$, which is Equation~\ref{eq:mse-decomp}. Setting $\rho \equiv 0$ recovers the independent case. \hfill$\square$

Equation~\ref{eq:mse-decomp} separates the error of the learned posterior mean from the sampling variance that additional trajectories can reduce. At matched NFE, antithetic coupling improves over independent averaging whenever $\bar{\rho}<0$, with gains decaying as $1/K$ and eventually saturating at the model-bias term. These gains improve the estimate of the conditional mean, and therefore distortion, but do not imply a better learned distribution or better individual samples.

% Because the learned reverse process is a nonlinear function of its driving noise, negative correlation is not guaranteed; Appendix~\ref{app:persample} therefore measures $\rho$ and the resulting distortion--compute frontier empirically across timesteps and modalities. These gains improve estimation of the conditional mean and therefore distortion, but do not imply improved learned distribution or individual-sample quality.

\paragraph{When $\rho<0$ holds.} For a function $f$ monotone in a single uniform variable, antithetic negative correlation is classical \citep[Ch.~8.2]{owen2013montecarlo}. No such argument applies here, since the map from $\{z_t\}$ to $\hat{x}_0$ passes through a learned nonlinear network and $\rho$ could in principle be positive, in which case Equation~\ref{eq:av-variance} predicts a variance increase. In the conditional-diffusion setting \citet{ruiperez2026antithetic} report pair correlations between $-0.98$ and $-0.88$ through most of the reverse trajectory, relaxing near $t=0$. Appendix~\ref{app:av-empirical} measures the variance-weighted correlation of \method{} directly and finds it negative in every modality: it never rises above $-0.89$ along the reverse trajectory and relaxes only near $t=0$, in close agreement with \citet{ruiperez2026antithetic}.

\subsection{Estimating the Cyclostationarity Index}
\label{app:cyclo-estimator}

Definition~\ref{def:pi} is estimated from clean data by phase-aligned averaging. Given clean signals $\{x^{(i)}\}$:

\begin{enumerate}[leftmargin=*,itemsep=1pt,topsep=2pt]
\item \textbf{Detect cycle events.} Apply the analytical ECG detector $\mathcal{D}$ (Table~\ref{tab:analytical-encoder}) to the \emph{clean} signal, giving event times $\{t_k\}$. This is the same detector used for phase-encoder warm-start, so no additional machinery is introduced.
\item \textbf{Resample to a phase grid.} For each cycle $[t_k, t_{k+1})$, linearly resample the segment onto a fixed grid of $G$ phase bins ($G=100$ throughout), giving $s_k \in \mathbb{R}^{G}$. Cycles whose length falls outside the physiological range $[1/f_{\max}, 1/f_{\min}]$ are discarded.
\item \textbf{Average and normalize.} Set $\hat{\mu}(\varphi) = N^{-1}\sum_k s_k[\varphi]$ over the $N$ retained cycles and
\begin{equation}
\label{eq:pi-hat}
\hat{\Pi}
= \frac{\operatorname{Var}_{\varphi}\big(\hat{\mu}(\varphi)\big)}
       {G^{-1}\sum_{\varphi}\big(N^{-1}\sum_k \operatorname{Var}(s_k[\varphi])\big) + \operatorname{Var}_{\varphi}\big(\hat{\mu}(\varphi)\big)},
\end{equation}
amplitude-normalizing each cycle beforehand so that $\hat{\Pi}$ measures shape consistency rather than amplitude stability.
\end{enumerate}

\paragraph{Finite-sample bias.} $\operatorname{Var}_\varphi(\hat{\mu})$ is upward-biased at small $N$, since averaging few cycles leaves residual noise in $\hat{\mu}$ that inflates its variance. The bias is $O(1/N)$ and we correct it by subtracting the within-bin variance divided by $N$; with $N \geq 200$ cycles per dataset the correction changes $\hat{\Pi}$ by less than $0.01$ in all cases we checked.

\paragraph{A cheaper proxy.} Where no reliable detector exists, the peak normalized autocorrelation within a plausible cycle-lag window,
\begin{equation}
\Pi_{\mathrm{AC}} = \frac{1}{K}\sum_{k=1}^{K}\max_{\ell \in [L_{\min}, L_{\max}]} \frac{\hat{R}_k(\ell)}{\hat{R}_k(0)},
\qquad
\hat{R}_k(\ell) = \frac{1}{W-\ell}\sum_{t=1}^{W-\ell} x_{k,t}\,x_{k,t+\ell},
\end{equation}
requires no event detection and is computable in one pass. It is computed on $K$ mean-removed windows $x_k$ of $W=4f_s$ samples (or the whole segment if shorter) with 50\% overlap, with $L_{\min} = f_s/f_{\max}$ and $L_{\max} = \min(f_s/f_{\min}, W/2)$; the factor $1/(W-\ell)$ removes the downward bias at long lags, and the short windows track slow drift in cycle length.
This is a weaker quantity, measuring self-similarity at a single lag instead of the variance decomposition of Definition~\ref{def:pi}, and is not the quantity Proposition~\ref{prop:mmse} bounds, but it works as a screen. We report both for ECG and the proxy for the other modalities in Table~\ref{tab:cyclicity-index} (Appendix~\ref{app:cyclicity}).

\newpage
\section{Extended Experimental Setup}
\label{app:extended-setup}

\subsection{Per-Modality Preprocessing}
\label{app:preprocessing}

This section details the datasets, splits, held-out test sets and corruption protocols summarized in Section~\ref{sec:setup}. All datasets follow their established benchmark protocols. For ECG, EMG and EEG, corruption noise is always drawn from held-out records or record halves, disjoint from training noise in both channel and segment, whereas for PPG training and testing noise are sampled from the same segments.

\paragraph{ECG (PTB-XL, MIT-BIH, QTDB).} PTB-XL \citep{PhysioNet-ptb-xl-1.0.3, wagner_ptb-xl_2020} contains 21{,}799 ten-second 12-lead records from 18{,}869 patients; we use the 500\,Hz records, lead II, and the official stratified folds (1--8 train, 9 validation, 10 test). Following \citet{li_tfcdiff_2025}, signals are converted to fixed 10\,s single-lead windows at 360\,Hz, band-passed with a zero-phase fifth-order Butterworth filter (0.5--40\,Hz), smoothed with a short moving average, and polyphase-resampled. QTDB \citep{laguna_qt_1997, goldberger_physiobank_2000} uses the same pipeline, excluding records sourced from MIT-BIH to avoid overlap. MIT-BIH \citep{moody_impact_2001} is natively 360\,Hz and is filtered, baseline-corrected and split into non-overlapping 10\,s windows without resampling; lead MLII, split DS1/DS2 following \citet{chazal_automatic_2004}. Corruption mixes baseline-wander, electrode-motion and muscle-artifact records from NSTDB \citep{moody_mit-bih_1992} as $y = x_0 + \lambda\,\frac{\operatorname{ptp}(x_0)}{\operatorname{ptp}(n)}\,n$ with $\lambda \in \{0.20,\dots,2.00\}$; training noise comes from the first half of channel 1, evaluation noise from the second half of channel 2.\looseness-1

\paragraph{PPG (PPG-DaLiA, BIDMC).} Wrist BVP recorded with an Empatica E4 at 64\,Hz \citep{attila_reiss_ppg-dalia_2019}. No reference implementation of the DDAE baseline or its preprocessing was available, so both were reimplemented following \citet{lai_enhanced_2025}; the resulting dataset is frozen and shared identically between \method{} and the baseline, so differences are attributable to the model, not to data handling. Signals are split into non-overlapping 512-sample ($\approx$8\,s) windows and screened by signal-quality index: segments are discarded if nearly flat, if they lack two pulse peaks separated by at least 20 samples, or if skewness falls below $0.3$; the smoothness test requires $\frac{1}{L-1}\sum_n |s[n{+}1]-s[n]| \geq 0.002\operatorname{ptp}(s)$. Accepted segments are min--max normalized. NSTDB noise is resampled to 64\,Hz and mixed with weights $(\beta_{\mathrm{MA}},\beta_{\mathrm{BW}},\beta_{\mathrm{EM}}) \in \{(0.33,0.33,0.34),(0.60,0.20,0.20),(0.20,0.60,0.20),(0.20,0.20,0.60)\}$, covering balanced, muscle-dominant, baseline-dominant and motion-dominant mixtures. Subject-level 80/10/10 split; all $4\times6$ noise-configuration/SNR combinations are materialized for test. The held-out BIDMC set \citep{pimentel_toward_2017} contains 53 eight-minute pulse-oximeter recordings at 125\,Hz; the PLETH channel is polyphase-resampled to 64\,Hz and then windowed, screened, normalized and corrupted exactly as the PPG-DaLiA test split, which keeps 2{,}317 of 3{,}180 windows from 52 records (55{,}608 test segments).

\paragraph{EMG (NinaPro DB2, DB3).} Following the SDEMG protocol \citep{liu_sdemg_2024}, clean sEMG comes from NinaPro DB2 \citep{atzori_electromyography_2014} at 2000\,Hz and ECG interference from NSRDB \citep{goldberger_physiobank_2000}. EMG channels are band-passed (fourth-order zero-phase Butterworth, 20--500\,Hz), downsampled to 1000\,Hz, normalized per recording by maximum absolute amplitude, and split into non-overlapping 10\,s windows ($L=10{,}000$). Training uses channel 2, exercise 1, subjects S11--S40; validation channel 2, exercise 3, subjects S1--S10; test channel 9, exercise 2, subjects S1--S10. ECG interference is resampled from 128\,Hz to 1000\,Hz and band-limited (third-order zero-phase, 10--200\,Hz), with records split disjointly across partitions (test records include 16420, 16539, 16786). Mixing uses $y = x_0 + \alpha w$ with $\alpha = \sqrt{\sum_n x_0[n]^2 / (10^{\lambda/10}\sum_n w[n]^2)}$; training and validation SNR grid $\{-5,-7,-9,-11,-13,-15\}$\,dB, test grid $\{0,-2,\dots,-14\}$\,dB, with 10 contaminated copies per training segment and 3 for validation and test. The held-out NinaPro DB3 \citep{atzori_electromyography_2014} records 11 trans-radial amputees with the same 12-electrode, 2\,kHz setup as DB2. We apply the DB2 test configuration unchanged (exercise 2, channel 9, same filtering, normalization, windowing, SNR grid and NSRDB test records, which are disjoint from the training and validation interference), excluding subjects 6 and 7, whose channel 9 is flat. Each of the 1{,}065 windows is mixed once with each test record at every SNR (25{,}560 test segments).

\paragraph{EEG (EEGdenoiseNet, EEGMMIDB).} EEG--EOG setting, following the EEGDfus protocol \citep{huang_eegdfus_2025} on EEGdenoiseNet \citep{zhang_eegdenoisenet_2021}. Epochs have $L=512$; clean EEG and EOG artifact epochs are randomly paired after truncating the EEG array to the number of available EOG epochs, and each signal is standardized before mixing. An 80/10/10 split is used, with 11 random recombinations of clean and artifact sources to enlarge the training set. Training SNR is drawn uniformly from $[-5,5]$\,dB; all integer levels in $\{-5,\dots,5\}$\,dB are materialized for test. Mixing follows the same $\alpha$ formula as EMG. The held-out EEGMMIDB \citep{schalk_bci2000_2004,goldberger_physiobank_2000} contains 64-channel recordings of 109 subjects. Channel Oz is band-passed (1\,Hz to 80HZ), notch-filtered at 60\,Hz, resampled to 256\,Hz and split into non-overlapping 512-sample windows. As these recordings are not curated to be artifact-free, windows with blinks, excess high-frequency (30--80\,Hz) power, or outlying kurtosis or amplitude are rejected; the kurtosis and high-frequency thresholds are the 99th percentiles of the EEGdenoiseNet clean epochs. Up to ten windows per subject (1{,}066 from 108 subjects) are standardized and mixed with held-out EEGdenoiseNet EOG epochs at every SNR in $\{-5,\dots,5\}$\,dB (11{,}726 test segments).

\subsection{Architecture and Training Details}
\label{app:hyperparameters}

\begin{figure}[t]
\centering
\includegraphics[width=\linewidth]{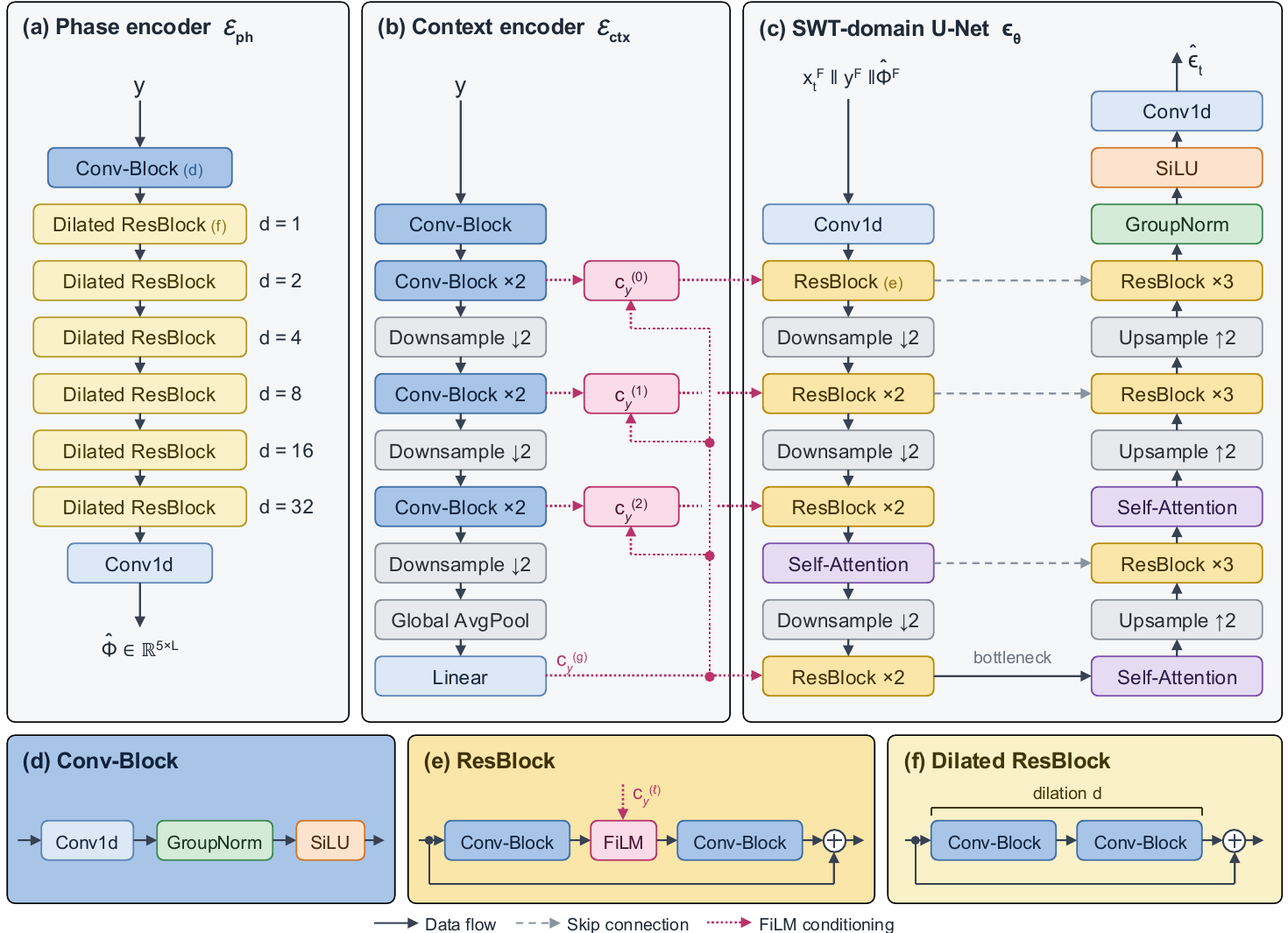}
\caption{Block-level architecture of \method, complementing the overview in
Figure~\ref{fig:architecture}.
\textbf{(a)} Phase encoder $\mathcal{E}_{ph}$: a convolutional stem, six dilated
residual blocks ($d=1,2,4,8,16,32$) whose receptive field spans several cycles,
and a projection to the five channels of the predicted phase field $\hat{\Phi}$,
which is then frame-transformed as $\hat{\Phi}^{\mathrm{F}}=\mathcal{W}(\hat{\Phi})$.
\textbf{(b)} Context encoder $\mathcal{E}_{ctx}$: three strided stages followed by
global average pooling, emitting one pooled embedding $c_y^{(\ell)}$ per U-Net
resolution and a global embedding $c_y^{(g)}$.
\textbf{(c)} SWT-domain denoising U-Net $\epsilon_\theta$, taking the channel-wise
concatenation $x_t^{\mathrm{F}}\,\|\,y^{\mathrm{F}}\,\|\,\hat{\Phi}^{\mathrm{F}}$
($7(J{+}1)=35$ channels at $J=4$) and predicting $\hat{\epsilon}_t$; dashed arrows
are skip connections.
\textbf{(d--f)} Internal structure of the Conv-Block, the FiLM-modulated ResBlock
and the dilated residual block. Each detail panel carries the colour of the block
it expands, and the letter shown at a block type's first occurrence points to its
panel. $\times k$ denotes $k$ stacked blocks. Decoder FiLM connections and the
diffusion-step embedding are omitted for clarity.}
\label{fig:architecture-detail}
\end{figure}

\paragraph{Architecture.}
The denoising network $\epsilon_\theta$ is a one-dimensional U-Net~\citep{ronneberger2015unet,ho_denoising_2020} with three resolution levels, two residual blocks per level, stride-2 downsampling, nearest-neighbor upsampling with skip connections, and self-attention at the bottleneck and deepest encoder and decoder blocks. Its dense input is the channel-wise concatenation of $x_t^{\mathrm F}$, $y^{\mathrm F}$, and $\hat{\Phi}^{\mathrm F}$. Since $x_t^{\mathrm F}$ and $y^{\mathrm F}$ each contain $J+1$ channels and $\hat{\Phi}$ contains five channels, the SWT-transformed phase field contributes $5(J+1)$ channels, giving $7(J+1)=35$ input channels for $J=4$. The phase encoder $\mathcal{E}_{ph}$ consists of a convolutional stem followed by six dilated residual blocks with dilations $1,2,4,8,16,32$~\citep{bai2018empiricalevaluationgenericconvolutional}. Its receptive field spans multiple local cycles, allowing it to estimate both cycle position and the local cycle rate $r_n$ of Equation~\ref{eq:phase-defs}. The context encoder $\mathcal{E}_{ctx}$ produces the pooled conditioning $c_y=\{c_y^{(\ell)}\}$ used in the main text: one embedding for each U-Net resolution and one global embedding. The resolution-specific embeddings modulate the corresponding U-Net features through FiLM~\citep{perez2018film}, while the global embedding is combined with the diffusion-step embedding. Figure~\ref{fig:architecture-detail} shows the block-level architecture.

\begin{table}[t]
\centering
\caption{Analytical ECG cycle-event detector used to construct the warm-start target $\Phi^\star=\mathcal{D}(x_0)$. Its parameters are fixed and the detector is not used at inference.}
\label{tab:analytical-encoder}
\resizebox{\textwidth}{!}{%
\begin{tabular}{lccccccc}
\toprule
Modality & Detector mode & Refractory [ms] & Smoothing [ms] & Event width [ms] & Threshold scale & $f_{\min}$ [Hz] & $f_{\max}$ [Hz] \\
\midrule
ECG & QRS peaks & $220$ & $80$ & $35$ & $0.5$ & $0.5$ & $3.0$ \\
% PPG & Pulse-like peaks & $250$ & $150$ & $60$ & $0.3$ & $0.5$ & $3.0$ \\
% EEG & Positive envelope & $60$ & $80$ & $35$ & $0.4$ & $0.5$ & $15.0$ \\
% EMG & Derivative energy & $3$ & $3$ & $3$ & $0.0$ & $10.0$ & $300.0$ \\
\bottomrule
\end{tabular}%
}
\end{table}

\paragraph{Band mapping across modalities.}
\label{app:band-mapping}
For a signal sampled at $f_s$ with $J$ decomposition levels, detail channel $d_j$ covers approximately $[f_s 2^{-(j+1)}, f_s 2^{-j}]$ and the approximation channel $a_J$ the residual below $f_s 2^{-(J+1)}$. With $J=4$: at 360\,Hz (ECG) the channels are 90--180, 45--90, 22.5--45, 11.25--22.5 and 0--11.25\,Hz, so the physiological band \citep{xie_computational_2020} concentrates in $a_4, d_4, d_3$ while $d_1, d_2$ carry predominantly artifact energy. At 64\,Hz (PPG) the same structure lands at 16--32, 8--16, 4--8, 2--4 and 0--2\,Hz; at 1000\,Hz (EMG) at 250--500, 125--250, 62.5--125, 31.25--62.5 and 0--31.25\,Hz. Finally, at 256\,Hz, the bands are 64--128, 32--64, 16--32, 8--16 and 0--8\,Hz. The assignment is derived from $f_s$, not tuned per modality.

\paragraph{Training hyperparameters and compute environment.}
Table~\ref{tab:hyperparameters} summarizes the modality-specific training settings; all unlisted architectural and optimization settings are shared across modalities. Unless otherwise noted, all results use a single training seed ($67$); the main restoration table (Table~\ref{tab:main-modalities}) is the exception, trained and evaluated with three seeds ($42$, $67$, $123$) and reported as mean $\pm$ standard deviation. Experiments were run as single-GPU jobs on a Slurm cluster equipped primarily with NVIDIA RTX 4090 GPUs, with some runs performed on RTX 3090 GPUs. For \method{}, training used four CPU data-loading workers, while validation used synchronous loading.
\begin{table}[htbp]
\centering
\caption{Training hyperparameters across modalities. Frame, backbone, diffusion schedule, objective form and sampler are shared; only the specified entries differ. EEG uses a different wavelet and morphology weight inherited from an earlier configuration that continues to perform better for that modality; a full ablation matching Tables~\ref{tab:swt-ablation} and \ref{tab:loss-ablation} has not been run for EEG.}

\label{tab:hyperparameters}
\resizebox{\textwidth}{!}{%
\begin{tabular}{lcccc}
\toprule
Hyperparameter & ECG & PPG & EEG & EMG \\
\midrule
Learning rate & $1\times10^{-4}$ & $1\times10^{-3}$ & $5\times10^{-4}$ & $1\times10^{-4}$ \\
Scheduler & MultiStepLR & ReduceLROnPlateau & MultiStepLR & ReduceLROnPlateau \\
Scheduler settings &
milestones $(0.4,0.7,0.9)$, $\gamma=0.1$ &
factor $0.5$, patience $10$ &
milestones $(0.95)$, $\gamma=0.1$ &
factor $0.5$, patience $5$ \\
Phase warm-start & Yes & No & No & No \\
Wavelet & Sym4 & Sym4 & Db4 & Sym4 \\
Decomposition levels $J$ & 4 & 4 & 4 & 4 \\
Objective &
$\mathcal{L}_{\mathrm{diff}} + 0.1\mathcal{L}_{\mathrm{morph}} + 0.3\mathcal{L}_{\mathrm{rec}}$ &
same as ECG &
$\mathcal{L}_{\mathrm{diff}} + 0.05\mathcal{L}_{\mathrm{morph}} + 0.3\mathcal{L}_{\mathrm{rec}}$ &
same as ECG \\
Batch size & $32$ & $32$ & $32$ & $16$ \\
Gradient accumulation & $1$ & $1$ & $1$ & $2$ \\
Effective batch size & $32$ & $32$ & $32$ & $32$ \\
Epochs & $500$ & $300$ & $420$ & $60$ \\
Base channels & $64$ & $64$ & $128$ & $64$ \\
EMA & disabled & disabled & disabled & enabled, decay $0.995$ \\
\midrule
\multicolumn{5}{p{1.5\textwidth}}{\textit{Shared:} AdamW (weight decay $10^{-4}$), gradient clipping at norm $1.0$; $T=50$ diffusion steps with a quadratic schedule, $\beta_t$ from $10^{-4}$ to $0.5$; conditioning dropout $0.3$; per-sample input normalization by the standard deviation of the corrupted input; checkpoint selected by validation $\Delta$SNR.} \\
\bottomrule
\end{tabular}%
}
\end{table}

\paragraph{Phase-encoder warm-start.}
For ECG, we warm-start $\mathcal{E}_{ph}$ against the target phase field $\Phi^\star=\mathcal{D}(x_0)$ computed from the clean signal, where $\mathcal{D}$ applies a Pan--Tompkins-style QRS detector~\citep{pan_real-time_1985} with the fixed parameters of Table~\ref{tab:analytical-encoder}, followed by the phase-field construction of Equations~\ref{eq:phase-defs}--\ref{eq:phase-field}. The warm-start objective combines a dense field loss
$\mathcal{L}_{\mathrm{feat}}=\|\hat{\Phi}-\Phi^\star\|_2^2$,
a Huber loss on the local rate estimate $\mathcal{L}_{f} = H(\hat{f}-f^\star)$, and a circular phase-alignment loss $\mathcal{L}_{\phi} = (\sin 2\pi\hat{\phi}_0 - \sin 2\pi\phi^\star_0)^2 + (\cos 2\pi\hat{\phi}_0 - \cos 2\pi\phi^\star_0)^2$, summed with unit weights. We pretrain the phase encoder for 60 epochs, freeze it for the first 200 epochs of joint restoration training, and subsequently optimize it through the restoration objective. The analytical construction $\mathcal{D}$ is used only for this ECG initialization and is never used at inference; the phase encoders for PPG, EEG, and EMG are trained directly through the restoration objective.

\subsection{Baseline Implementation Details}
\label{app:baseline-details}
This section details the baselines listed in Section~\ref{sec:setup}, whose results appear in Table~\ref{tab:main-modalities} and Appendix~\ref{app:supp-results}. All learned baselines are retrained on the same prepared tensors, splits, and normalization as \method{}, using seed 67 and checkpoint selection by validation loss. We preserve the published architectures and objectives unless noted below. Deterministic methods produce one reconstruction; diffusion baselines use the original publications' sampling procedure.

\paragraph{ECG.}
The classical ECG baselines are a zero-phase, 721-tap Hamming-window FIR filter with a 0.5--40\,Hz passband and an eight-level \texttt{sym6} SWT denoiser using band pruning and soft BayesShrink thresholding. TCDAE~\citep{chen_elimination_2024} is a deterministic convolutional--Transformer autoencoder that combines local feature extraction with global temporal attention. We port its architecture and frequency-weighted reconstruction loss from Keras to PyTorch and cap training at 400 epochs with early stopping. DeScoD-ECG~\citep{li_descod-ecg_2024} is a conditional time-domain DDPM with a hierarchical multi-kernel denoiser, while TFCDiff~\citep{li_tfcdiff_2025} performs conditional diffusion on the first 1{,}000 DCT coefficients. Both use their published 50-step quadratic schedules and are trained for at most 400 epochs. The PTB-XL checkpoints are evaluated on MIT-BIH and QTDB without retraining.

\paragraph{PPG.}
DDAE~\citep{lai_enhanced_2025} is a deterministic dilated residual autoencoder designed for PPG motion-artifact removal. Because no reference implementation was available, we reimplemented the published architecture and its equally weighted MSE--$\ell_1$ objective. We use the reported 250 epochs, AdamW optimizer, and batch size 32, but disable additional noise augmentation because the shared training tensors are already corrupted.

\paragraph{EEG.}
EEGIFNet~\citep{cui_dual-branch_2024} separately estimates clean EEG and artifact components before combining them through a learned mask; we use its ocular-artifact variant. EEGDfus~\citep{huang_eegdfus_2025} is a dual-branch conditional diffusion model using a 500-step linear schedule. For both methods, we retain the published model and training settings but replace the runtime data split with the fixed split shared by all methods.

\paragraph{EMG.}
FCN~\citep{wang_ecg_2023} is a deterministic fully convolutional encoder--decoder. We transfer it to EMG by adapting only the input length and retraining it on NinaPro with its $\ell_1$ reconstruction objective. SDEMG~\citep{liu_sdemg_2024} is a conditional diffusion model specialized for removing ECG contamination from surface EMG; we retain its 50-step cosine schedule and $\ell_2$ noise-prediction objective, but we adapt its batch size from 64 to 16 and increase gradient accumulation from 1 to 4 batches due to memory constraints. We also adapt its data interface to the shared tensors.

\subsection{Evaluation Metrics}
\label{app:metrics}

Table~\ref{tab:metrics} defines the three metrics reported in the main text (Section~\ref{sec:setup}) and the benchmark-specific metrics used in Table~\ref{tab:denoise-ecg-combined} and Figure~\ref{fig:cross-modality-original}. All metrics are computed per test sample and averaged. Let $x_0$ be the clean signal, $y$ the corrupted input, $\hat{x}_0$ the estimate, $e = \hat{x}_0-x_0$, and $\mathrm{SNR}(x_0,u) = 10\log_{10}(\sum_n x_0[n]^2/\sum_n (u[n]-x_0[n])^2)$. For the EEG spectral metric, $\mathrm{PSD}(x)=|\mathrm{FFT}_{400}(x)|^2/400$. For EMG, $a_k(x) = |I_k|^{-1}\sum_{n\in I_k}|x[n]|$ is the mean rectified value on the $k$-th 1000-sample window and $\mathrm{MF}_m(x) = \sum_i f_i S_x[i,m]/\sum_i S_x[i,m]$ the mean frequency of frame $m$, with $S_x$ the STFT magnitude over 10--500\,Hz and $A$ the active-stimulus frames \citep{wang_ecg_2023}.

\begin{table}[htbp]
\centering
\caption{Evaluation metrics. The first block is reported for every modality; the remainder are the benchmark-specific metrics conventionally used in each literature.}
\label{tab:metrics}
\small
\begin{tabular}{@{}llcl@{}}
\toprule
Setting & Metric & Direction & Definition \\
\midrule
All modalities & $\Delta$SNR & $\uparrow$ & $\mathrm{SNR}(x_0,\hat{x}_0)-\mathrm{SNR}(x_0,y)$ \\
All modalities & PRD & $\downarrow$ & $100\sqrt{\sum_n e[n]^2 / \sum_n (x_0[n]-\bar{x}_0)^2}$ \\
All modalities & CC & $\uparrow$ & $\frac{\sum_n (x_0[n]-\bar{x}_0)(\hat{x}_0[n]-\bar{\hat{x}}_0)}{\sqrt{\sum_n (x_0[n]-\bar{x}_0)^2 \sum_n (\hat{x}_0[n]-\bar{\hat{x}}_0)^2}}$ \\
\midrule
ECG & SSD & $\downarrow$ & $\sum_n e[n]^2$ \\
ECG & MAD & $\downarrow$ & $\max_n |e[n]|$ \\
ECG & CosSim & $\uparrow$ & $\langle x_0,\hat{x}_0\rangle / (\|x_0\|_2\|\hat{x}_0\|_2)$ \\
\midrule
PPG & RMSE & $\downarrow$ & $\sqrt{L^{-1}\sum_n e[n]^2}$ \\
PPG & PRD$_{\mathrm{PPG}}$ & $\downarrow$ & $100\sqrt{\sum_n e[n]^2 / \sum_n x_0[n]^2}$ \\
\midrule
EEG & RRMSE$_t$ & $\downarrow$ & $\sqrt{\mathrm{mean}(e^2)} / \sqrt{\mathrm{mean}(x_0^2)}$ \\
EEG & RRMSE$_s$ & $\downarrow$ & Relative RMSE between $\mathrm{PSD}(\hat{x}_0)$ and $\mathrm{PSD}(x_0)$ \\
\midrule
EMG & RMSE & $\downarrow$ & $\sqrt{L^{-1}\sum_n e[n]^2}$ \\
EMG & ARV & $\downarrow$ & $\sqrt{K^{-1}\sum_{k}(a_k(x_0)-a_k(\hat{x}_0))^2}$ \\
EMG & MF & $\downarrow$ & $\sqrt{|A|^{-1}\sum_{m\in A}(\mathrm{MF}_m(x_0)-\mathrm{MF}_m(\hat{x}_0))^2}$ \\
\midrule
Classification & AUROC & $\uparrow$ & One-vs-rest AUC; macro: $C^{-1}\sum_c \mathrm{AUROC}_c$ \\
\bottomrule
\end{tabular}
\end{table}

\subsection{Statistical Evaluation Procedure}
\label{app:statistics}

Unless stated otherwise, reported values carry 95\% bootstrap confidence intervals over 1{,}000 resamples \citep{strotdhoff2021}; Table~\ref{tab:main-modalities} instead reports mean $\pm$ standard deviation over three training seeds (Section~\ref{sec:setup}). Paired comparisons pair samples by test-set index, so each comparison uses the difference between \method{} and a baseline on the same signal. For metric $m$ the paired advantage is $\Delta_i = m_i^{\method} - m_i^{\mathrm{baseline}}$ where higher is better and the negation otherwise, so $\Delta_i>0$ always favours \method. We test $\Delta_i>0$ with a one-sided Wilcoxon signed-rank test \citep{woolson_wilcoxon_2008}, chosen because it does not assume normally distributed differences, excluding zero differences by the standard procedure. Multiple \method--baseline comparisons are corrected by Holm--Bonferroni at $\alpha=0.05$. Table~\ref{tab:paired-statistical-comparison} reports mean paired advantage and adjusted $p$-values; all comparisons favor \method{} and all remain significant after correction.

\section{Extended Results}
\label{app:supp-results}

\subsection{Full ECG results across datasets}
\label{app:full-ecg}

Table~\ref{tab:denoise-ecg-combined} gives all reported metrics on PTB-XL, QTDB and MIT-BIH, including the classical baselines omitted from the main text and the SSD/MAD/CosSim metrics used by prior ECG work \citep{li_descod-ecg_2024,li_tfcdiff_2025}. \method{} is best on every metric and dataset except MAD on QTDB, where it ties TFCDiff.

\providecommand{\best}[1]{\begingroup\bfseries\boldmath #1\endgroup}
\providecommand{\secondbest}[1]{\underline{#1}}
\begin{table*}[htbp]
\centering
\caption{Denoising comparison across ECG datasets (PTB-XL, QTDB, MIT-BIH). Values are mean $\pm$ 95\% CI. $\uparrow$ higher is better, $\downarrow$ lower is better. Bold and underlined values mark the best and second-best denoising method within each dataset and metric.}
\label{tab:denoise-ecg-combined}
\resizebox{\textwidth}{!}{%
\begin{tabular}{llccc|ccc}
\toprule
Dataset & Model & $\Delta$SNR $\uparrow$ & PRD [\%] $\downarrow$ & CC $\uparrow$ & SSD $\downarrow$ & MAD $\downarrow$ & CosSim $\uparrow$ \\
\midrule
\multirow{14}{*}{PTB-XL} & Noisy ($\mathrm{SNR}_\mathrm{in}=-2.0$\,dB) & $0.00$ & $152.86 \pm 3.67$ & $0.61 \pm 0.01$ & $210.66 \pm 17.44$ & $0.60 \pm 0.02$ & $0.60 \pm 0.01$ \\
\cmidrule(lr){2-8}
 & FIR filter & $7.44 \pm 0.14$ & $64.13 \pm 1.44$ & $0.83 \pm 0.01$ & $35.52 \pm 2.61$ & $0.36 \pm 0.01$ & $0.83 \pm 0.01$ \\
 & SWT denoising & $6.22 \pm 0.16$ & $68.16 \pm 1.19$ & $0.76 \pm 0.01$ & $35.02 \pm 2.16$ & $0.35 \pm 0.01$ & $0.76 \pm 0.01$ \\
 & TCDAE & $12.29 \pm 0.17$ & $34.03 \pm 0.65$ & $0.95 \pm 0.00$ & $8.71 \pm 0.76$ & $0.25 \pm 0.01$ & $0.95 \pm 0.00$ \\
 & DeScoD-1 & $11.82 \pm 0.19$ & $36.11 \pm 0.70$ & $0.92 \pm 0.00$ & $10.14 \pm 0.97$ & $0.26 \pm 0.01$ & $0.92 \pm 0.00$ \\
 & DeScoD-3 & $13.18 \pm 0.19$ & $31.04 \pm 0.61$ & $0.94 \pm 0.00$ & $7.73 \pm 0.89$ & $0.22 \pm 0.01$ & $0.94 \pm 0.00$ \\
 & DeScoD-5 & $13.49 \pm 0.20$ & $29.99 \pm 0.59$ & $0.94 \pm 0.00$ & $7.34 \pm 1.01$ & $0.21 \pm 0.01$ & $0.94 \pm 0.00$ \\
 & DeScoD-10 & $13.78 \pm 0.20$ & $29.05 \pm 0.57$ & $0.95 \pm 0.00$ & $6.91 \pm 0.93$ & $0.20 \pm 0.01$ & $0.95 \pm 0.00$ \\
 & TFCDiff-1 & $11.53 \pm 0.18$ & $36.95 \pm 0.71$ & $0.91 \pm 0.00$ & $11.38 \pm 1.57$ & $0.24 \pm 0.01$ & $0.91 \pm 0.00$ \\
 & TFCDiff-3 & $12.86 \pm 0.18$ & $31.98 \pm 0.65$ & $0.93 \pm 0.00$ & $9.16 \pm 1.38$ & $0.22 \pm 0.01$ & $0.93 \pm 0.00$ \\
 & TFCDiff-5 & $13.19 \pm 0.19$ & $30.86 \pm 0.64$ & $0.94 \pm 0.00$ & $8.61 \pm 1.31$ & $0.21 \pm 0.01$ & $0.94 \pm 0.00$ \\
 & TFCDiff-10 & $13.47 \pm 0.19$ & $29.91 \pm 0.61$ & $0.94 \pm 0.00$ & $8.30 \pm 1.38$ & $0.21 \pm 0.01$ & $0.94 \pm 0.00$ \\
\cmidrule(lr){2-8}
 & \method{}-MC-10 & \secondbest{$15.79 \pm 0.21$} & \secondbest{$23.17 \pm 0.51$} & \secondbest{$0.97 \pm 0.00$} & \secondbest{$4.26 \pm 0.51$} & \secondbest{$0.15 \pm 0.01$} & \secondbest{$0.97 \pm 0.00$} \\
 & \method{}-AV-10 & \best{$16.09 \pm 0.21$} & \best{$22.43 \pm 0.50$} & \best{$0.97 \pm 0.00$} & \best{$4.00 \pm 0.50$} & \best{$0.15 \pm 0.01$} & \best{$0.97 \pm 0.00$} \\
\midrule
\multirow{8}{*}{QTDB} & Noisy ($\mathrm{SNR}_\mathrm{in}=-0.8$\,dB) & $0.00$ & $133.58 \pm 0.89$ & $0.65 \pm 0.00$ & $413.54 \pm 10.84$ & $0.80 \pm 0.01$ & $0.64 \pm 0.00$ \\
\cmidrule(lr){2-8}
 & FIR filter & $6.48 \pm 0.04$ & $60.04 \pm 0.36$ & $0.85 \pm 0.00$ & $79.42 \pm 2.06$ & $0.52 \pm 0.01$ & $0.85 \pm 0.00$ \\
 & SWT denoising & $6.25 \pm 0.04$ & $59.43 \pm 0.31$ & $0.82 \pm 0.00$ & $76.15 \pm 1.90$ & $0.47 \pm 0.00$ & $0.81 \pm 0.00$ \\
 & TCDAE & $10.33 \pm 0.05$ & $36.22 \pm 0.16$ & $0.95 \pm 0.00$ & $29.45 \pm 1.09$ & $0.40 \pm 0.00$ & $0.94 \pm 0.00$ \\
 & DeScoD-10 & $11.07 \pm 0.06$ & $33.91 \pm 0.17$ & $0.94 \pm 0.00$ & $42.21 \pm 2.73$ & $0.38 \pm 0.00$ & $0.93 \pm 0.00$ \\
 & TFCDiff-10 & $11.71 \pm 0.06$ & $31.85 \pm 0.17$ & $0.94 \pm 0.00$ & $46.48 \pm 3.25$ & \best{$0.33 \pm 0.00$} & $0.94 \pm 0.00$ \\
\cmidrule(lr){2-8}
 & \method{}-MC-10 & \secondbest{$13.05 \pm 0.06$} & \secondbest{$27.01 \pm 0.14$} & \secondbest{$0.96 \pm 0.00$} & \secondbest{$15.38 \pm 0.50$} & $0.33 \pm 0.00$ & \secondbest{$0.96 \pm 0.00$} \\
 & \method{}-AV-10 & \best{$13.23 \pm 0.06$} & \best{$26.51 \pm 0.14$} & \best{$0.96 \pm 0.00$} & \best{$14.84 \pm 0.49$} & \secondbest{$0.33 \pm 0.00$} & \best{$0.96 \pm 0.00$} \\
\midrule
\multirow{8}{*}{MIT-BIH} & Noisy ($\mathrm{SNR}_\mathrm{in}=-1.7$\,dB) & $0.00$ & $148.07 \pm 2.67$ & $0.62 \pm 0.01$ & $921.71 \pm 48.68$ & $1.27 \pm 0.03$ & $0.61 \pm 0.01$ \\
\cmidrule(lr){2-8}
 & FIR filter & $6.82 \pm 0.11$ & $64.72 \pm 1.05$ & $0.83 \pm 0.00$ & $172.43 \pm 9.05$ & $0.81 \pm 0.02$ & $0.83 \pm 0.00$ \\
 & SWT denoising & $6.48 \pm 0.12$ & $64.42 \pm 0.92$ & $0.78 \pm 0.00$ & $170.17 \pm 8.78$ & $0.75 \pm 0.02$ & $0.78 \pm 0.00$ \\
 & TCDAE & $11.21 \pm 0.13$ & $37.10 \pm 0.52$ & $0.93 \pm 0.00$ & $58.07 \pm 3.42$ & $0.65 \pm 0.02$ & $0.93 \pm 0.00$ \\
 & DeScoD-10 & $11.25 \pm 0.15$ & $37.75 \pm 0.55$ & $0.91 \pm 0.00$ & $73.22 \pm 4.55$ & $0.64 \pm 0.02$ & $0.91 \pm 0.00$ \\
 & TFCDiff-10 & $11.21 \pm 0.15$ & $37.85 \pm 0.58$ & $0.91 \pm 0.00$ & $79.97 \pm 5.19$ & $0.60 \pm 0.02$ & $0.91 \pm 0.00$ \\
\cmidrule(lr){2-8}
 & \method{}-MC-10 & \secondbest{$13.68 \pm 0.16$} & \secondbest{$28.53 \pm 0.47$} & \secondbest{$0.95 \pm 0.00$} & \secondbest{$37.22 \pm 3.08$} & \secondbest{$0.55 \pm 0.02$} & \secondbest{$0.95 \pm 0.00$} \\
 & \method{}-AV-10 & \best{$13.88 \pm 0.16$} & \best{$27.93 \pm 0.47$} & \best{$0.95 \pm 0.00$} & \best{$35.86 \pm 3.00$} & \best{$0.55 \pm 0.02$} & \best{$0.95 \pm 0.00$} \\
\bottomrule
\end{tabular}%
}
\end{table*}

\subsection{Results stratified by corruption level}
\label{app:noise-bins}

\begin{figure}[htbp]
\centering
\includegraphics[width=\linewidth]{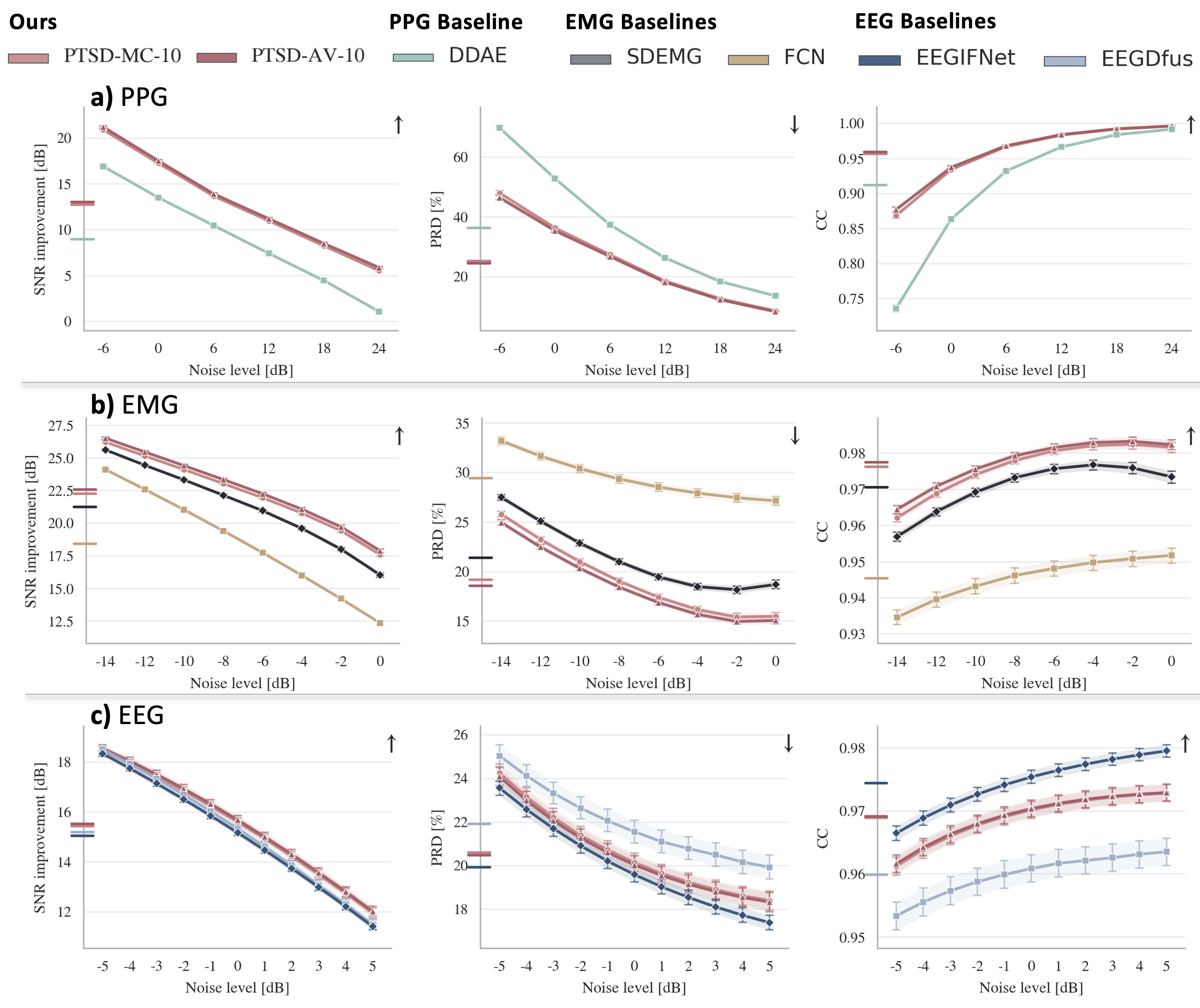}
\caption{Restoration across corruption levels for PPG-DaLiA (a), NinaPro (b) and EEGdenoiseNet (c), grouped by each modality's corruption grid. The $y$-intercept marks the average per model and metric; arrows indicate the better direction.}
\label{fig:cross-modality-bins}
\end{figure}

\begin{figure}[htbp]
\centering
\includegraphics[width=\linewidth]{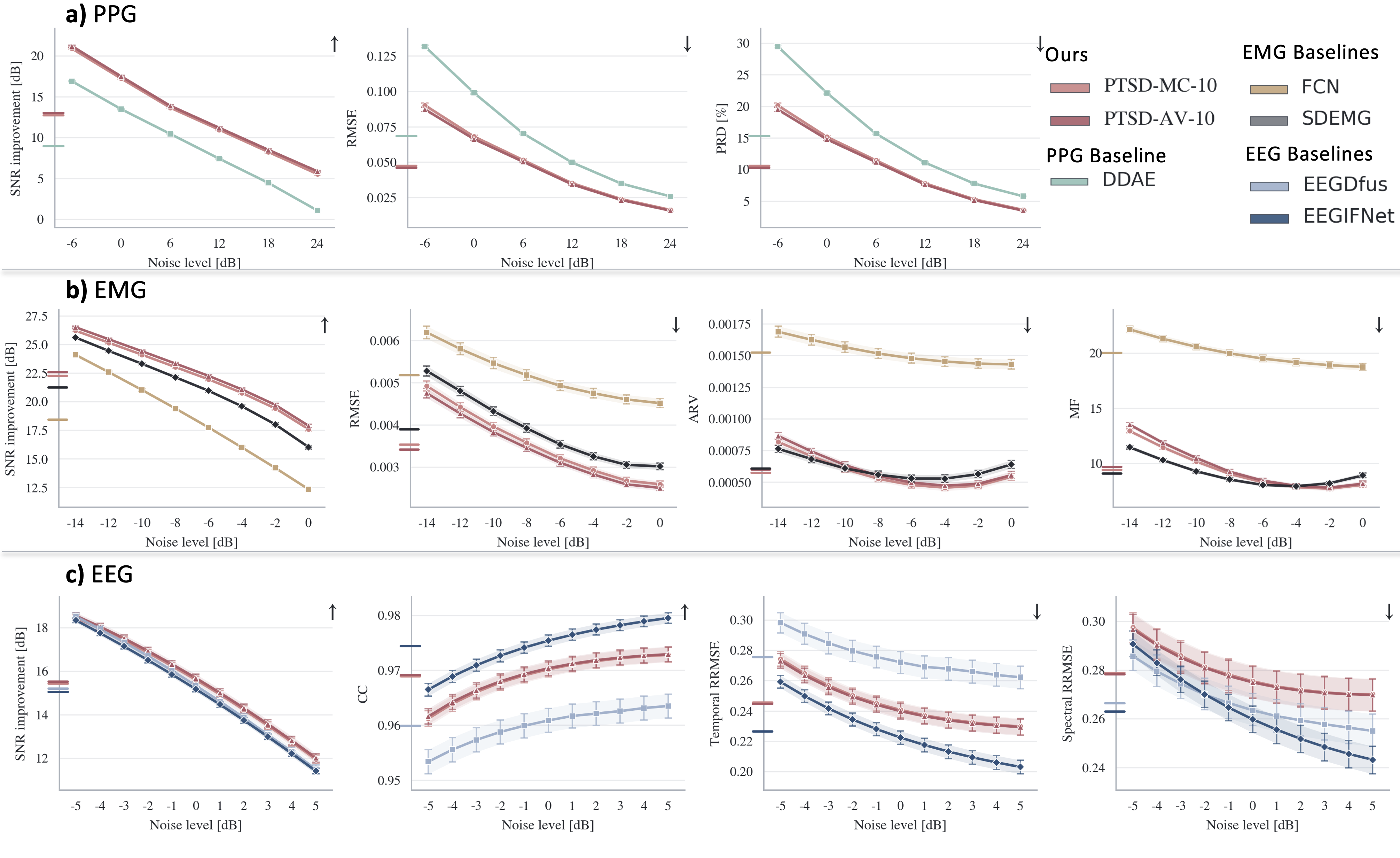}
\caption{Restoration across corruption levels using each benchmark's own metrics. \textbf{(a)} PPG-DaLiA ($\Delta$SNR, RMSE, PRD), following \citet{lai_enhanced_2025}. \textbf{(b)} NinaPro ($\Delta$SNR, RMSE, ARV, MF), following \citet{liu_sdemg_2024}. \textbf{(c)} EEGdenoiseNet ($\Delta$SNR, CC, temporal and spectral RRMSE), following \citet{huang_eegdfus_2025}. On PPG and EEG the ordering follows the main metrics; on EMG, ARV and mean-frequency error are more mixed under heavy corruption, consistent with trajectory averaging mildly smoothing amplitude and spectral summaries (Appendix~\ref{app:proof-av}).}
\label{fig:cross-modality-original}
\end{figure}

Section~\ref{sec:results} summarizes the finding that \method{}'s advantage increases under heavier corruption. Tables~\ref{tab:denoise-ptbxl-bins-combined}--\ref{tab:denoise-mitbih-bins-combined} stratify ECG performance by the corruption scaling factor $\lambda$ and report both our main metrics and the metrics used by prior work. \method{} is best on every metric in every bin on every dataset, except MAD in the $0.2$--$0.6$ and $0.6$--$1.0$ bins. Figure~\ref{fig:cross-modality-bins} gives the corresponding stratification for the other three modalities. Figure~\ref{fig:cross-modality-original} repeats this stratification with each benchmark's own metrics (Table~\ref{tab:metrics}): the ordering follows the main metrics on PPG and EEG, while ARV and mean-frequency error on EMG are mixed under heavy corruption.

% The advantage widens with corruption, which is the pattern predicted by Section~\ref{sec:results-mechanism}: a learned phase estimator should separate from a classical detector precisely where the detector fails.

\providecommand{\best}[1]{\begingroup\bfseries\boldmath #1\endgroup}
\providecommand{\secondbest}[1]{\underline{#1}}
\begin{table*}[htbp]
\centering
\caption{Denoising comparison by noise level on PTB-XL. Values are mean $\pm$ 95\% CI. $\uparrow$ higher is better, $\downarrow$ lower is better. Bold marks the best and underline the second best within each bin and metric.}
\label{tab:denoise-ptbxl-bins-combined}
\resizebox{\textwidth}{!}{%
\begin{tabular}{llccc|ccc}
\toprule
Noise level & Model & $\Delta$SNR $\uparrow$ & PRD [\%] $\downarrow$ & CC $\uparrow$ & SSD $\downarrow$ & MAD $\downarrow$ & CosSim $\uparrow$ \\
\midrule
0.2-0.6 & Noisy & $0.00$ & $55.45 \pm 2.05$ & $0.87 \pm 0.01$ & $22.48 \pm 2.70$ & $0.21 \pm 0.01$ & $0.87 \pm 0.01$ \\
\midrule
 & FIR filter & $6.79 \pm 0.28$ & $25.28 \pm 0.92$ & $0.96 \pm 0.00$ & $4.68 \pm 0.63$ & $0.13 \pm 0.01$ & $0.96 \pm 0.00$ \\
 & SWT denoising & $3.04 \pm 0.38$ & $38.20 \pm 1.12$ & $0.92 \pm 0.01$ & $9.88 \pm 1.57$ & $0.17 \pm 0.01$ & $0.92 \pm 0.01$ \\
 & TCDAE & $8.14 \pm 0.34$ & $21.25 \pm 0.67$ & $0.98 \pm 0.00$ & $2.66 \pm 0.32$ & $0.13 \pm 0.01$ & $0.98 \pm 0.00$ \\
 & DeScoD-1 & $7.49 \pm 0.35$ & $23.23 \pm 0.88$ & $0.97 \pm 0.00$ & $3.73 \pm 0.60$ & $0.12 \pm 0.01$ & $0.97 \pm 0.00$ \\
 & DeScoD-3 & $8.83 \pm 0.37$ & $20.08 \pm 0.80$ & $0.98 \pm 0.00$ & $3.00 \pm 0.59$ & $0.11 \pm 0.01$ & $0.98 \pm 0.00$ \\
 & DeScoD-5 & $9.13 \pm 0.38$ & $19.45 \pm 0.80$ & $0.98 \pm 0.00$ & $2.75 \pm 0.53$ & $0.10 \pm 0.01$ & $0.98 \pm 0.00$ \\
 & DeScoD-10 & $9.39 \pm 0.38$ & $18.94 \pm 0.79$ & $0.98 \pm 0.00$ & $2.63 \pm 0.51$ & $0.10 \pm 0.01$ & $0.98 \pm 0.00$ \\
 & TFCDiff-1 & $7.25 \pm 0.32$ & $23.76 \pm 0.89$ & $0.97 \pm 0.00$ & $4.20 \pm 0.74$ & $0.13 \pm 0.01$ & $0.97 \pm 0.00$ \\
 & TFCDiff-3 & $8.62 \pm 0.34$ & $20.55 \pm 0.88$ & $0.97 \pm 0.00$ & $3.56 \pm 0.79$ & $0.12 \pm 0.01$ & $0.97 \pm 0.00$ \\
 & TFCDiff-5 & $8.99 \pm 0.34$ & $19.75 \pm 0.85$ & $0.98 \pm 0.00$ & $3.33 \pm 0.71$ & $0.12 \pm 0.01$ & $0.98 \pm 0.00$ \\
 & TFCDiff-10 & $9.30 \pm 0.34$ & $19.09 \pm 0.83$ & $0.98 \pm 0.00$ & $3.03 \pm 0.64$ & $0.11 \pm 0.01$ & $0.98 \pm 0.00$ \\
\midrule
 & \method{}-MC-10 & \secondbest{$10.99 \pm 0.37$} & \secondbest{$15.67 \pm 0.63$} & \secondbest{$0.99 \pm 0.00$} & \secondbest{$1.55 \pm 0.19$} & \secondbest{$0.08 \pm 0.00$} & \secondbest{$0.99 \pm 0.00$} \\
 & \method{}-AV-10 & \best{$11.29 \pm 0.37$} & \best{$15.17 \pm 0.60$} & \best{$0.99 \pm 0.00$} & \best{$1.44 \pm 0.18$} & \best{$0.07 \pm 0.00$} & \best{$0.99 \pm 0.00$} \\
\midrule
0.6-1.0 & Noisy & $0.00$ & $113.84 \pm 3.56$ & $0.68 \pm 0.01$ & $100.30 \pm 14.05$ & $0.44 \pm 0.02$ & $0.67 \pm 0.01$ \\
\midrule
 & FIR filter & $7.64 \pm 0.30$ & $47.50 \pm 1.48$ & $0.90 \pm 0.01$ & $17.41 \pm 2.44$ & $0.26 \pm 0.02$ & $0.90 \pm 0.01$ \\
 & SWT denoising & $6.14 \pm 0.30$ & $55.08 \pm 1.31$ & $0.83 \pm 0.01$ & $20.16 \pm 2.15$ & $0.27 \pm 0.01$ & $0.83 \pm 0.01$ \\
 & TCDAE & $12.07 \pm 0.31$ & $28.29 \pm 0.90$ & $0.97 \pm 0.00$ & $5.23 \pm 0.91$ & $0.19 \pm 0.01$ & $0.97 \pm 0.00$ \\
 & DeScoD-1 & $11.50 \pm 0.31$ & $30.46 \pm 1.01$ & $0.95 \pm 0.00$ & $6.56 \pm 1.08$ & $0.19 \pm 0.02$ & $0.95 \pm 0.00$ \\
 & DeScoD-3 & $12.89 \pm 0.33$ & $26.11 \pm 0.90$ & $0.96 \pm 0.00$ & $5.00 \pm 0.90$ & $0.17 \pm 0.01$ & $0.96 \pm 0.00$ \\
 & DeScoD-5 & $13.15 \pm 0.33$ & $25.36 \pm 0.90$ & $0.96 \pm 0.00$ & $4.63 \pm 0.79$ & $0.16 \pm 0.01$ & $0.96 \pm 0.00$ \\
 & DeScoD-10 & $13.47 \pm 0.34$ & $24.48 \pm 0.86$ & $0.97 \pm 0.00$ & $4.37 \pm 0.77$ & $0.16 \pm 0.01$ & $0.97 \pm 0.00$ \\
 & TFCDiff-1 & $10.86 \pm 0.29$ & $32.63 \pm 1.06$ & $0.94 \pm 0.00$ & $7.70 \pm 1.24$ & $0.20 \pm 0.01$ & $0.94 \pm 0.00$ \\
 & TFCDiff-3 & $12.17 \pm 0.29$ & $28.22 \pm 0.97$ & $0.95 \pm 0.00$ & $6.11 \pm 1.12$ & $0.19 \pm 0.01$ & $0.95 \pm 0.00$ \\
 & TFCDiff-5 & $12.51 \pm 0.31$ & $27.18 \pm 0.96$ & $0.96 \pm 0.00$ & $5.81 \pm 1.14$ & $0.18 \pm 0.01$ & $0.96 \pm 0.00$ \\
 & TFCDiff-10 & $12.78 \pm 0.31$ & $26.38 \pm 0.94$ & $0.96 \pm 0.00$ & $5.39 \pm 0.99$ & $0.18 \pm 0.01$ & $0.96 \pm 0.00$ \\
\midrule
 & \method{}-MC-10 & \secondbest{$15.15 \pm 0.37$} & \secondbest{$20.30 \pm 0.76$} & \secondbest{$0.98 \pm 0.00$} & \secondbest{$2.86 \pm 0.53$} & \secondbest{$0.12 \pm 0.01$} & \secondbest{$0.98 \pm 0.00$} \\
 & \method{}-AV-10 & \best{$15.40 \pm 0.37$} & \best{$19.79 \pm 0.75$} & \best{$0.98 \pm 0.00$} & \best{$2.75 \pm 0.52$} & \best{$0.11 \pm 0.01$} & \best{$0.98 \pm 0.00$} \\
\midrule
1.0-1.5 & Noisy & $0.00$ & $172.88 \pm 4.28$ & $0.53 \pm 0.01$ & $224.67 \pm 25.78$ & $0.68 \pm 0.03$ & $0.52 \pm 0.01$ \\
\midrule
 & FIR filter & $7.63 \pm 0.25$ & $71.84 \pm 1.76$ & $0.81 \pm 0.01$ & $39.18 \pm 4.34$ & $0.40 \pm 0.02$ & $0.81 \pm 0.01$ \\
 & SWT denoising & $7.21 \pm 0.23$ & $74.04 \pm 1.46$ & $0.73 \pm 0.01$ & $38.98 \pm 3.84$ & $0.38 \pm 0.01$ & $0.73 \pm 0.01$ \\
 & TCDAE & $13.78 \pm 0.25$ & $35.68 \pm 0.94$ & $0.95 \pm 0.00$ & $8.65 \pm 1.03$ & $0.26 \pm 0.01$ & $0.95 \pm 0.00$ \\
 & DeScoD-1 & $13.27 \pm 0.28$ & $38.29 \pm 1.12$ & $0.91 \pm 0.01$ & $10.77 \pm 1.39$ & $0.28 \pm 0.02$ & $0.91 \pm 0.01$ \\
 & DeScoD-3 & $14.64 \pm 0.29$ & $32.82 \pm 0.98$ & $0.94 \pm 0.00$ & $7.99 \pm 1.00$ & $0.24 \pm 0.01$ & $0.94 \pm 0.00$ \\
 & DeScoD-5 & $14.94 \pm 0.29$ & $31.73 \pm 0.96$ & $0.94 \pm 0.00$ & $7.54 \pm 0.98$ & $0.23 \pm 0.01$ & $0.94 \pm 0.00$ \\
 & DeScoD-10 & $15.27 \pm 0.29$ & $30.63 \pm 0.95$ & $0.94 \pm 0.00$ & $7.11 \pm 0.97$ & $0.22 \pm 0.01$ & $0.94 \pm 0.00$ \\
 & TFCDiff-1 & $12.94 \pm 0.26$ & $39.12 \pm 1.07$ & $0.91 \pm 0.01$ & $10.84 \pm 1.31$ & $0.25 \pm 0.01$ & $0.91 \pm 0.01$ \\
 & TFCDiff-3 & $14.27 \pm 0.27$ & $33.80 \pm 0.97$ & $0.93 \pm 0.01$ & $8.83 \pm 1.29$ & $0.23 \pm 0.01$ & $0.93 \pm 0.01$ \\
 & TFCDiff-5 & $14.62 \pm 0.26$ & $32.48 \pm 0.93$ & $0.94 \pm 0.00$ & $8.03 \pm 1.15$ & $0.22 \pm 0.01$ & $0.94 \pm 0.00$ \\
 & TFCDiff-10 & $14.86 \pm 0.27$ & $31.64 \pm 0.93$ & $0.94 \pm 0.00$ & $7.78 \pm 1.12$ & $0.22 \pm 0.01$ & $0.94 \pm 0.00$ \\
\midrule
 & \method{}-MC-10 & \secondbest{$17.46 \pm 0.30$} & \secondbest{$24.12 \pm 0.86$} & \secondbest{$0.96 \pm 0.00$} & \secondbest{$4.51 \pm 0.73$} & \secondbest{$0.16 \pm 0.01$} & \secondbest{$0.96 \pm 0.00$} \\
 & \method{}-AV-10 & \best{$17.79 \pm 0.31$} & \best{$23.29 \pm 0.83$} & \best{$0.97 \pm 0.00$} & \best{$4.16 \pm 0.63$} & \best{$0.16 \pm 0.01$} & \best{$0.97 \pm 0.00$} \\
\midrule
1.5-2.0 & Noisy & $0.00$ & $240.37 \pm 6.07$ & $0.42 \pm 0.01$ & $434.06 \pm 50.37$ & $0.96 \pm 0.04$ & $0.41 \pm 0.01$ \\
\midrule
 & FIR filter & $7.56 \pm 0.28$ & $100.72 \pm 2.61$ & $0.70 \pm 0.01$ & $71.79 \pm 6.82$ & $0.57 \pm 0.02$ & $0.70 \pm 0.01$ \\
 & SWT denoising & $7.80 \pm 0.26$ & $96.59 \pm 2.13$ & $0.60 \pm 0.01$ & $63.55 \pm 5.65$ & $0.52 \pm 0.02$ & $0.60 \pm 0.01$ \\
 & TCDAE & $14.20 \pm 0.23$ & $47.22 \pm 1.32$ & $0.92 \pm 0.00$ & $16.61 \pm 2.44$ & $0.38 \pm 0.02$ & $0.92 \pm 0.00$ \\
 & DeScoD-1 & $13.99 \pm 0.29$ & $48.96 \pm 1.40$ & $0.86 \pm 0.01$ & $17.82 \pm 2.83$ & $0.40 \pm 0.02$ & $0.86 \pm 0.01$ \\
 & DeScoD-3 & $15.30 \pm 0.30$ & $42.24 \pm 1.25$ & $0.89 \pm 0.01$ & $13.73 \pm 2.79$ & $0.34 \pm 0.02$ & $0.89 \pm 0.01$ \\
 & DeScoD-5 & $15.65 \pm 0.32$ & $40.63 \pm 1.26$ & $0.90 \pm 0.01$ & $13.21 \pm 3.18$ & $0.33 \pm 0.02$ & $0.90 \pm 0.01$ \\
 & DeScoD-10 & $15.93 \pm 0.32$ & $39.43 \pm 1.23$ & $0.91 \pm 0.01$ & $12.42 \pm 2.95$ & $0.32 \pm 0.02$ & $0.91 \pm 0.01$ \\
 & TFCDiff-1 & $13.98 \pm 0.28$ & $48.91 \pm 1.54$ & $0.85 \pm 0.01$ & $20.84 \pm 5.08$ & $0.36 \pm 0.02$ & $0.85 \pm 0.01$ \\
 & TFCDiff-3 & $15.28 \pm 0.29$ & $42.49 \pm 1.38$ & $0.89 \pm 0.01$ & $16.69 \pm 4.57$ & $0.32 \pm 0.02$ & $0.89 \pm 0.01$ \\
 & TFCDiff-5 & $15.55 \pm 0.29$ & $41.18 \pm 1.34$ & $0.90 \pm 0.01$ & $15.88 \pm 4.38$ & $0.31 \pm 0.02$ & $0.90 \pm 0.01$ \\
 & TFCDiff-10 & $15.85 \pm 0.30$ & $39.81 \pm 1.30$ & $0.90 \pm 0.01$ & $15.54 \pm 4.73$ & $0.30 \pm 0.02$ & $0.90 \pm 0.01$ \\
\midrule
 & \method{}-MC-10 & \secondbest{$18.31 \pm 0.32$} & \secondbest{$30.67 \pm 1.23$} & \secondbest{$0.94 \pm 0.00$} & \secondbest{$7.48 \pm 1.62$} & \secondbest{$0.23 \pm 0.02$} & \secondbest{$0.94 \pm 0.00$} \\
 & \method{}-AV-10 & \best{$18.63 \pm 0.33$} & \best{$29.65 \pm 1.21$} & \best{$0.95 \pm 0.00$} & \best{$7.07 \pm 1.62$} & \best{$0.23 \pm 0.02$} & \best{$0.95 \pm 0.00$} \\
\bottomrule
\end{tabular}%
}
\end{table*}
\providecommand{\best}[1]{\begingroup\bfseries\boldmath #1\endgroup}
\providecommand{\secondbest}[1]{\underline{#1}}
\begin{table*}[htbp]
\centering
\caption{Denoising comparison by noise level on QTDB. Values are mean $\pm$ 95\% CI. $\uparrow$ higher is better, $\downarrow$ lower is better. Bold marks the best and underline the second best within each bin and metric.}
\label{tab:denoise-qtdb-bins-combined}
\resizebox{\textwidth}{!}{%
\begin{tabular}{llccc|ccc}
\toprule
Noise level & Model & $\Delta$SNR $\uparrow$ & PRD [\%] $\downarrow$ & CC $\uparrow$ & SSD $\downarrow$ & MAD $\downarrow$ & CosSim $\uparrow$ \\
\midrule
0.2-0.6 & Noisy & $0.00$ & $48.26 \pm 0.49$ & $0.90 \pm 0.00$ & $48.54 \pm 2.29$ & $0.29 \pm 0.00$ & $0.90 \pm 0.00$ \\
\midrule
 & FIR filter & $4.52 \pm 0.09$ & $27.97 \pm 0.26$ & $0.96 \pm 0.00$ & $15.92 \pm 0.79$ & $0.27 \pm 0.01$ & $0.96 \pm 0.00$ \\
 & SWT denoising & $3.05 \pm 0.10$ & $32.84 \pm 0.27$ & $0.95 \pm 0.00$ & $23.19 \pm 1.52$ & \secondbest{$0.24 \pm 0.00$} & $0.94 \pm 0.00$ \\
 & TCDAE & $5.02 \pm 0.10$ & $26.09 \pm 0.21$ & $0.98 \pm 0.00$ & $16.57 \pm 1.51$ & $0.28 \pm 0.01$ & $0.97 \pm 0.00$ \\
 & DeScoD-10 & $6.08 \pm 0.11$ & $23.66 \pm 0.24$ & $0.97 \pm 0.00$ & $24.91 \pm 4.52$ & $0.24 \pm 0.01$ & $0.97 \pm 0.00$ \\
 & TFCDiff-10 & $6.81 \pm 0.11$ & $22.32 \pm 0.27$ & $0.98 \pm 0.00$ & $32.81 \pm 6.16$ & \best{$0.21 \pm 0.01$} & $0.97 \pm 0.00$ \\
\midrule
 & \method{}-MC-10 & \secondbest{$7.47 \pm 0.11$} & \secondbest{$20.05 \pm 0.19$} & \secondbest{$0.98 \pm 0.00$} & \secondbest{$8.04 \pm 0.41$} & $0.27 \pm 0.01$ & \secondbest{$0.98 \pm 0.00$} \\
 & \method{}-AV-10 & \best{$7.62 \pm 0.11$} & \best{$19.76 \pm 0.19$} & \best{$0.98 \pm 0.00$} & \best{$7.84 \pm 0.40$} & $0.27 \pm 0.01$ & \best{$0.98 \pm 0.00$} \\
\midrule
0.6-1.0 & Noisy & $0.00$ & $97.02 \pm 0.77$ & $0.73 \pm 0.00$ & $189.20 \pm 8.74$ & $0.59 \pm 0.01$ & $0.73 \pm 0.00$ \\
\midrule
 & FIR filter & $6.51 \pm 0.08$ & $45.43 \pm 0.32$ & $0.91 \pm 0.00$ & $41.12 \pm 1.88$ & $0.40 \pm 0.01$ & $0.90 \pm 0.00$ \\
 & SWT denoising & $6.14 \pm 0.08$ & $47.07 \pm 0.31$ & $0.89 \pm 0.00$ & $45.60 \pm 2.35$ & $0.37 \pm 0.01$ & $0.88 \pm 0.00$ \\
 & TCDAE & $9.90 \pm 0.08$ & $30.94 \pm 0.24$ & $0.96 \pm 0.00$ & $22.39 \pm 1.95$ & $0.33 \pm 0.01$ & $0.96 \pm 0.00$ \\
 & DeScoD-10 & $10.54 \pm 0.09$ & $29.28 \pm 0.27$ & $0.96 \pm 0.00$ & $37.99 \pm 6.11$ & $0.31 \pm 0.01$ & $0.95 \pm 0.00$ \\
 & TFCDiff-10 & $11.05 \pm 0.10$ & $28.02 \pm 0.31$ & $0.96 \pm 0.00$ & $46.69 \pm 8.23$ & \best{$0.29 \pm 0.01$} & $0.95 \pm 0.00$ \\
\midrule
 & \method{}-MC-10 & \secondbest{$12.26 \pm 0.09$} & \secondbest{$24.06 \pm 0.23$} & \secondbest{$0.97 \pm 0.00$} & \secondbest{$11.82 \pm 0.63$} & $0.30 \pm 0.01$ & \secondbest{$0.97 \pm 0.00$} \\
 & \method{}-AV-10 & \best{$12.43 \pm 0.10$} & \best{$23.63 \pm 0.23$} & \best{$0.97 \pm 0.00$} & \best{$11.45 \pm 0.61$} & \secondbest{$0.30 \pm 0.01$} & \best{$0.97 \pm 0.00$} \\
\midrule
1.0-1.5 & Noisy & $0.00$ & $152.18 \pm 1.06$ & $0.58 \pm 0.00$ & $449.84 \pm 17.38$ & $0.91 \pm 0.01$ & $0.57 \pm 0.00$ \\
\midrule
 & FIR filter & $7.14 \pm 0.07$ & $66.49 \pm 0.41$ & $0.83 \pm 0.00$ & $84.66 \pm 3.11$ & $0.57 \pm 0.01$ & $0.82 \pm 0.00$ \\
 & SWT denoising & $7.30 \pm 0.06$ & $64.76 \pm 0.38$ & $0.79 \pm 0.00$ & $81.28 \pm 3.05$ & $0.52 \pm 0.01$ & $0.79 \pm 0.00$ \\
 & TCDAE & $12.14 \pm 0.06$ & $37.57 \pm 0.25$ & $0.94 \pm 0.00$ & $29.46 \pm 1.94$ & $0.41 \pm 0.01$ & $0.94 \pm 0.00$ \\
 & DeScoD-10 & $12.79 \pm 0.08$ & $35.51 \pm 0.28$ & $0.93 \pm 0.00$ & $43.71 \pm 5.26$ & $0.39 \pm 0.01$ & $0.93 \pm 0.00$ \\
 & TFCDiff-10 & $13.32 \pm 0.08$ & $33.64 \pm 0.29$ & $0.94 \pm 0.00$ & $46.71 \pm 6.06$ & $0.35 \pm 0.01$ & $0.94 \pm 0.00$ \\
\midrule
 & \method{}-MC-10 & \secondbest{$14.88 \pm 0.08$} & \secondbest{$28.13 \pm 0.24$} & \secondbest{$0.96 \pm 0.00$} & \secondbest{$15.45 \pm 0.71$} & \secondbest{$0.34 \pm 0.01$} & \secondbest{$0.95 \pm 0.00$} \\
 & \method{}-AV-10 & \best{$15.05 \pm 0.08$} & \best{$27.62 \pm 0.24$} & \best{$0.96 \pm 0.00$} & \best{$14.93 \pm 0.70$} & \best{$0.34 \pm 0.01$} & \best{$0.96 \pm 0.00$} \\
\midrule
1.5-2.0 & Noisy & $0.00$ & $213.44 \pm 1.36$ & $0.45 \pm 0.00$ & $856.33 \pm 31.74$ & $1.27 \pm 0.02$ & $0.45 \pm 0.00$ \\
\midrule
 & FIR filter & $7.37 \pm 0.06$ & $91.35 \pm 0.58$ & $0.74 \pm 0.00$ & $157.12 \pm 6.07$ & $0.77 \pm 0.01$ & $0.73 \pm 0.00$ \\
 & SWT denoising & $7.87 \pm 0.06$ & $85.62 \pm 0.51$ & $0.69 \pm 0.00$ & $138.88 \pm 5.34$ & $0.69 \pm 0.01$ & $0.68 \pm 0.00$ \\
 & TCDAE & $13.15 \pm 0.06$ & $47.31 \pm 0.33$ & $0.91 \pm 0.00$ & $45.53 \pm 2.73$ & $0.55 \pm 0.01$ & $0.90 \pm 0.00$ \\
 & DeScoD-10 & $13.82 \pm 0.08$ & $44.33 \pm 0.34$ & $0.89 \pm 0.00$ & $57.87 \pm 5.62$ & $0.53 \pm 0.01$ & $0.88 \pm 0.00$ \\
 & TFCDiff-10 & $14.61 \pm 0.08$ & $40.85 \pm 0.36$ & $0.91 \pm 0.00$ & $56.65 \pm 5.84$ & $0.44 \pm 0.01$ & $0.90 \pm 0.00$ \\
\midrule
 & \method{}-MC-10 & \secondbest{$16.38 \pm 0.08$} & \secondbest{$33.89 \pm 0.33$} & \secondbest{$0.94 \pm 0.00$} & \secondbest{$24.19 \pm 1.55$} & \secondbest{$0.40 \pm 0.01$} & \secondbest{$0.93 \pm 0.00$} \\
 & \method{}-AV-10 & \best{$16.58 \pm 0.08$} & \best{$33.16 \pm 0.32$} & \best{$0.94 \pm 0.00$} & \best{$23.22 \pm 1.51$} & \best{$0.40 \pm 0.01$} & \best{$0.93 \pm 0.00$} \\
\bottomrule
\end{tabular}%
}
\end{table*}
\providecommand{\best}[1]{\begingroup\bfseries\boldmath #1\endgroup}
\providecommand{\secondbest}[1]{\underline{#1}}
\begin{table*}[htbp]
\centering
\caption{Denoising comparison by noise level on MIT-BIH. Values are mean $\pm$ 95\% CI. $\uparrow$ higher is better, $\downarrow$ lower is better. Bold marks the best and underline the second best within each bin and metric.\looseness-1}
\label{tab:denoise-mitbih-bins-combined}
\resizebox{\textwidth}{!}{%
\begin{tabular}{llccc|ccc}
\toprule
Noise level & Model & $\Delta$SNR $\uparrow$ & PRD [\%] $\downarrow$ & CC $\uparrow$ & SSD $\downarrow$ & MAD $\downarrow$ & CosSim $\uparrow$ \\
\midrule
0.2-0.6 & Noisy & $0.00$ & $52.72 \pm 1.51$ & $0.88 \pm 0.00$ & $98.31 \pm 9.28$ & $0.44 \pm 0.02$ & $0.88 \pm 0.01$ \\
\midrule
 & FIR filter & $5.15 \pm 0.26$ & $28.56 \pm 0.79$ & $0.96 \pm 0.00$ & $27.96 \pm 2.82$ & \secondbest{$0.37 \pm 0.02$} & $0.96 \pm 0.00$ \\
 & SWT denoising & $3.28 \pm 0.29$ & $34.95 \pm 0.86$ & $0.93 \pm 0.00$ & $49.65 \pm 6.14$ & \best{$0.36 \pm 0.02$} & $0.93 \pm 0.00$ \\
 & TCDAE & $6.63 \pm 0.29$ & $23.87 \pm 0.60$ & $0.97 \pm 0.00$ & $22.28 \pm 3.97$ & $0.41 \pm 0.02$ & $0.97 \pm 0.00$ \\
 & DeScoD-10 & $6.58 \pm 0.31$ & $24.96 \pm 0.77$ & $0.96 \pm 0.00$ & $26.46 \pm 3.73$ & $0.38 \pm 0.02$ & $0.96 \pm 0.00$ \\
 & TFCDiff-10 & $6.27 \pm 0.33$ & $26.59 \pm 0.91$ & $0.96 \pm 0.00$ & $40.82 \pm 6.68$ & $0.39 \pm 0.03$ & $0.96 \pm 0.00$ \\
\midrule
 & \method{}-MC-10 & \secondbest{$8.30 \pm 0.30$} & \secondbest{$20.22 \pm 0.61$} & \secondbest{$0.98 \pm 0.00$} & \secondbest{$14.20 \pm 1.97$} & $0.42 \pm 0.03$ & \secondbest{$0.98 \pm 0.00$} \\
 & \method{}-AV-10 & \best{$8.50 \pm 0.30$} & \best{$19.82 \pm 0.61$} & \best{$0.98 \pm 0.00$} & \best{$13.71 \pm 1.92$} & $0.42 \pm 0.03$ & \best{$0.98 \pm 0.00$} \\
\midrule
0.6-1.0 & Noisy & $0.00$ & $108.02 \pm 2.33$ & $0.70 \pm 0.01$ & $400.45 \pm 33.69$ & $0.93 \pm 0.03$ & $0.69 \pm 0.01$ \\
\midrule
 & FIR filter & $6.90 \pm 0.21$ & $48.62 \pm 0.95$ & $0.89 \pm 0.00$ & $83.29 \pm 7.07$ & $0.63 \pm 0.03$ & $0.89 \pm 0.00$ \\
 & SWT denoising & $6.44 \pm 0.21$ & $50.63 \pm 0.93$ & $0.86 \pm 0.00$ & $92.60 \pm 7.87$ & $0.58 \pm 0.02$ & $0.86 \pm 0.00$ \\
 & TCDAE & $11.26 \pm 0.22$ & $29.96 \pm 0.80$ & $0.95 \pm 0.00$ & $37.98 \pm 6.81$ & $0.52 \pm 0.03$ & $0.95 \pm 0.00$ \\
 & DeScoD-10 & $10.81 \pm 0.27$ & $32.62 \pm 1.02$ & $0.94 \pm 0.00$ & $53.45 \pm 8.82$ & $0.53 \pm 0.03$ & $0.94 \pm 0.00$ \\
 & TFCDiff-10 & $10.61 \pm 0.28$ & $33.34 \pm 1.06$ & $0.94 \pm 0.00$ & $60.49 \pm 10.16$ & $0.51 \pm 0.03$ & $0.94 \pm 0.00$ \\
\midrule
 & \method{}-MC-10 & \secondbest{$13.17 \pm 0.26$} & \secondbest{$24.78 \pm 0.79$} & \secondbest{$0.96 \pm 0.00$} & \secondbest{$26.31 \pm 5.83$} & \secondbest{$0.47 \pm 0.03$} & \secondbest{$0.96 \pm 0.00$} \\
 & \method{}-AV-10 & \best{$13.38 \pm 0.27$} & \best{$24.25 \pm 0.78$} & \best{$0.96 \pm 0.00$} & \best{$25.45 \pm 5.80$} & \best{$0.47 \pm 0.03$} & \best{$0.96 \pm 0.00$} \\
\midrule
1.0-1.5 & Noisy & $0.00$ & $170.77 \pm 3.19$ & $0.54 \pm 0.01$ & $1029.43 \pm 68.65$ & $1.48 \pm 0.04$ & $0.53 \pm 0.01$ \\
\midrule
 & FIR filter & $7.38 \pm 0.19$ & $72.60 \pm 1.42$ & $0.80 \pm 0.01$ & $189.45 \pm 13.19$ & $0.91 \pm 0.03$ & $0.80 \pm 0.01$ \\
 & SWT denoising & $7.54 \pm 0.18$ & $70.71 \pm 1.22$ & $0.75 \pm 0.01$ & $187.36 \pm 13.55$ & $0.84 \pm 0.03$ & $0.75 \pm 0.01$ \\
 & TCDAE & $12.65 \pm 0.17$ & $40.04 \pm 0.85$ & $0.92 \pm 0.00$ & $62.99 \pm 5.26$ & $0.70 \pm 0.03$ & $0.92 \pm 0.00$ \\
 & DeScoD-10 & $12.71 \pm 0.22$ & $40.83 \pm 0.95$ & $0.90 \pm 0.01$ & $83.84 \pm 8.08$ & $0.71 \pm 0.03$ & $0.90 \pm 0.01$ \\
 & TFCDiff-10 & $12.73 \pm 0.22$ & $40.63 \pm 0.97$ & $0.91 \pm 0.01$ & $86.36 \pm 9.15$ & $0.66 \pm 0.03$ & $0.91 \pm 0.01$ \\
\midrule
 & \method{}-MC-10 & \secondbest{$15.38 \pm 0.22$} & \secondbest{$30.29 \pm 0.82$} & \secondbest{$0.94 \pm 0.00$} & \secondbest{$38.33 \pm 3.98$} & \secondbest{$0.58 \pm 0.03$} & \secondbest{$0.94 \pm 0.00$} \\
 & \method{}-AV-10 & \best{$15.58 \pm 0.22$} & \best{$29.67 \pm 0.81$} & \best{$0.95 \pm 0.00$} & \best{$36.92 \pm 3.94$} & \best{$0.57 \pm 0.03$} & \best{$0.94 \pm 0.00$} \\
\midrule
1.5-2.0 & Noisy & $0.00$ & $235.06 \pm 4.32$ & $0.42 \pm 0.01$ & $1898.28 \pm 137.32$ & $1.99 \pm 0.06$ & $0.41 \pm 0.01$ \\
\midrule
 & FIR filter & $7.51 \pm 0.21$ & $99.41 \pm 1.88$ & $0.71 \pm 0.01$ & $345.47 \pm 24.24$ & $1.20 \pm 0.04$ & $0.71 \pm 0.01$ \\
 & SWT denoising & $8.00 \pm 0.19$ & $93.34 \pm 1.66$ & $0.64 \pm 0.01$ & $312.92 \pm 23.36$ & $1.10 \pm 0.04$ & $0.64 \pm 0.01$ \\
 & TCDAE & $13.38 \pm 0.16$ & $50.73 \pm 1.05$ & $0.89 \pm 0.01$ & $98.09 \pm 9.47$ & $0.89 \pm 0.03$ & $0.88 \pm 0.01$ \\
 & DeScoD-10 & $13.88 \pm 0.25$ & $49.16 \pm 1.12$ & $0.85 \pm 0.01$ & $115.16 \pm 11.91$ & $0.87 \pm 0.04$ & $0.85 \pm 0.01$ \\
 & TFCDiff-10 & $14.11 \pm 0.23$ & $47.84 \pm 1.16$ & $0.86 \pm 0.01$ & $119.92 \pm 13.99$ & $0.78 \pm 0.04$ & $0.86 \pm 0.01$ \\
\midrule
 & \method{}-MC-10 & \secondbest{$16.67 \pm 0.24$} & \secondbest{$36.67 \pm 1.11$} & \secondbest{$0.92 \pm 0.01$} & \secondbest{$63.68 \pm 9.14$} & \secondbest{$0.70 \pm 0.03$} & \secondbest{$0.92 \pm 0.01$} \\
 & \method{}-AV-10 & \best{$16.87 \pm 0.24$} & \best{$35.88 \pm 1.10$} & \best{$0.92 \pm 0.01$} & \best{$61.27 \pm 8.72$} & \best{$0.69 \pm 0.03$} & \best{$0.92 \pm 0.01$} \\
\bottomrule
\end{tabular}%
}
\end{table*}

\subsection{Cross-dataset generalization}
\label{app:crossdataset}

\begin{figure}[htbp]
\centering
\includegraphics[width=\linewidth]{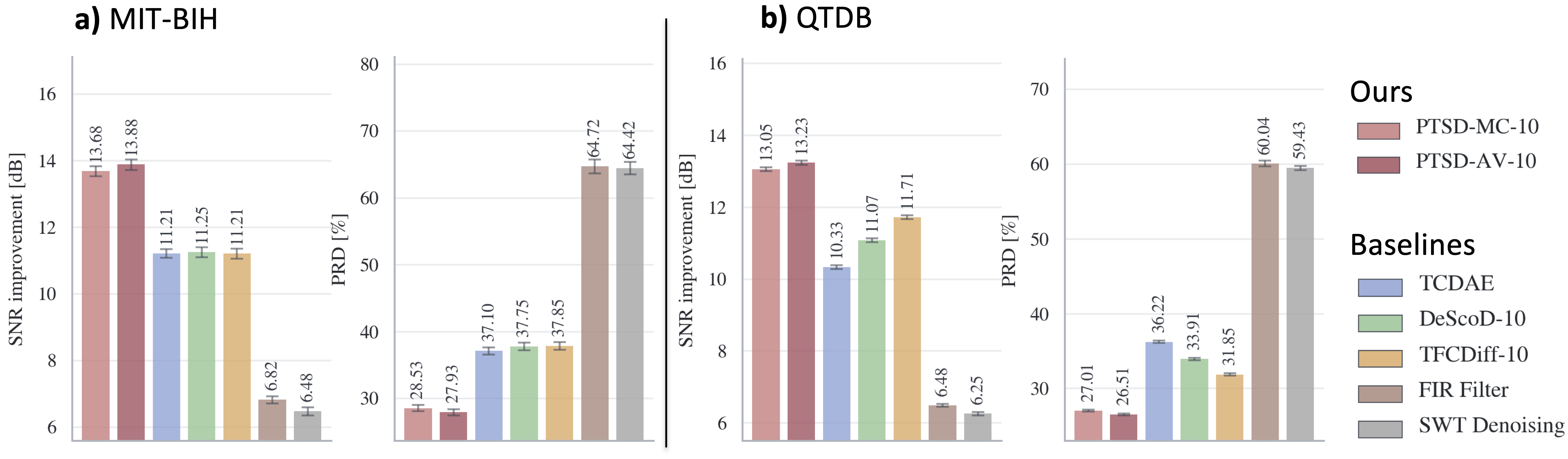}
\caption{ECG Cross-dataset generalization to MIT-BIH (a) and QTDB (b) without retraining. Error bars are 95\% CI.}
\label{fig:cross-dataset}
\end{figure}

Figure~\ref{fig:cross-dataset} evaluates ECG models on MIT-BIH and QTDB with no dataset-specific retraining. \method-AV-10 achieves the highest $\Delta$SNR and lowest PRD on both. On MIT-BIH, the diffusion baselines lose their advantage over the deterministic TCDAE while \method{} does not; we read this as evidence for beat-local rather than global rhythm conditioning, since arrhythmic beats violate a global period but leave a local phase field well-defined. Table~\ref{tab:crossdataset-modalities} extends the comparison to the other modalities. On PPG (PPG-DaLiA to BIDMC) and EMG (NinaPro DB2 to DB3), \method{}-AV-10 improves on the strongest baseline in all three metrics, by $2.08$ and $1.49$\,dB $\Delta$SNR over DDAE and SDEMG, respectively. On EEG (EEGdenoiseNet to EEGMMIDB), EEGIFNet stays ahead on all three metrics, by $0.37$\,dB $\Delta$SNR, consistent with EEG being the modality with the smallest in-domain margin (Table~\ref{tab:main-modalities}). \method{} does however outperform the EEGDfus baseline on all three metrics, by $2.02$\,dB $\Delta$SNR, $13.54$ PRD points and $0.105$ CC. On this dataset, all restorers nonetheless perform markedly worse than in-domain, because they remove much of the clean EEG below 4\,Hz, the band occupied by the EOG artefacts seen in training.

% \begin{table}[htbp]
% \centering
% \caption{\ph{\texttt{placeholder\_crossdataset\_modalities}: cross-dataset transfer for the non-ECG modalities, matching the ECG protocol (train on one dataset, evaluate on a held-out dataset with no retraining). Candidate held-out sets: BIDMC or WESAD for PPG, GRABMyo or NinaPro DB3 for EMG, a semi-simulated EEG artifact corpus for EEG. ECG rows are measured.}}
% \label{tab:placeholder_crossdataset}
% \ph{%
% \begin{tabular}{lllcc}
% \toprule
% Modality & Train & Held-out & $\Delta$SNR $\uparrow$ & $\Delta$ vs.\ best baseline \\
% \midrule
% PPG & PPG-DaLiA     & BIDMC & $7.48$ & $+2.12$  \\
% ECG & PTB-XL        & MIT-BIH & $13.88$ & $+2.63$ \\
% ECG & PTB-XL        & QTDB    & $13.23$ & $+1.52$ \\
% EMG & NinaPro DB2   & \ph{placeholder\_dataset} & $\square\square.\square\square$ & $+\square.\square\square$ \\
% EEG & EEGdenoiseNet & \ph{placeholder\_dataset} & $\square\square.\square\square$ & $+\square.\square\square$ \\
% \bottomrule
% \end{tabular}}
% \end{table}

\begin{table*}[htbp]
\centering
\caption{Dataset shift for all modalities: train on one dataset, evaluate on a held-out dataset with no retraining, against the strongest restoration baseline on the held-out set (named per row). $\Delta$ columns give the margin of \method{}-AV-10 over that baseline (positive = \method{} better) on each of the three main metrics.}
\label{tab:crossdataset-modalities}
\resizebox{\linewidth}{!}{%
\begin{tabular}{lllccccccc}
\toprule
Modality & Train & Held-out & Best baseline & $\Delta$SNR $\uparrow$ & $\Delta$ & PRD [\%] $\downarrow$ & $\Delta$ & CC $\uparrow$ & $\Delta$ \\
\midrule
PPG & PPG-DaLiA     & BIDMC       & DDAE       & $16.41$ & $+2.08$ & $32.06$ & $+8.09$ & $0.945$ & $+0.048$ \\
ECG & PTB-XL        & MIT-BIH     & DeScoD-10  & $13.88$ & $+2.63$ & $27.93$ & $+9.83$ & $0.949$ & $+0.039$ \\
ECG & PTB-XL        & QTDB        & TFCDiff-10 & $13.23$ & $+1.52$ & $26.51$ & $+5.35$ & $0.963$ & $+0.019$ \\
EMG & NinaPro DB2   & NinaPro DB3 & SDEMG      & $23.45$ & $+1.49$ & $16.16$ & $+3.20$ & $0.985$ & $+0.007$ \\
EEG & EEGdenoiseNet & EEGMMIDB    & EEGIFNet   & $5.35$  & $-0.37$ & $56.18$ & $-2.75$ & $0.808$ & $-0.029$ \\

\bottomrule
\end{tabular}%
}
\end{table*}

\subsection{Paired statistical comparison}
\label{app:paired}
Table~\ref{tab:paired-statistical-comparison} reports the paired tests of Appendix~\ref{app:statistics} behind the ECG significance claims of Section~\ref{sec:results}. \method{}-AV-10 is significantly better than every learned and classical baseline on all three main metrics and all three ECG datasets after Holm--Bonferroni correction.

\begin{table}[htbp]
\centering
\caption{Paired statistical comparison of \method{}-AV-10 against denoising baselines across the main metrics. Each cell reports the mean paired difference (\method{}-AV-10 $-$ baseline) in the metric's native units, with the Holm--Bonferroni-adjusted $p$-value below. $\Delta$SNR (dB) and CC are higher-is-better, PRD (percentage points) is lower-is-better; stars mark where \method{}-AV-10 is significantly better.}
\label{tab:paired-statistical-comparison}
\small
\begin{tabular}{llccc}
\toprule
\textbf{Baseline} & \textbf{Metric} & \textbf{MIT-BIH} & \textbf{PTB-XL} & \textbf{QTDB} \\
\midrule
\multirow{3}{*}{TFCDiff-10} & $\Delta$SNR & \begin{tabular}{@{}c@{}}$+2.68^{***}$\\[-0.15em]{\scriptsize $p_{adj}=<1e-300$}\end{tabular} & \begin{tabular}{@{}c@{}}$+2.62^{***}$\\[-0.15em]{\scriptsize $p_{adj}=<1e-300$}\end{tabular} & \begin{tabular}{@{}c@{}}$+1.52^{***}$\\[-0.15em]{\scriptsize $p_{adj}=<1e-300$}\end{tabular} \\
 & PRD [\%] & \begin{tabular}{@{}c@{}}$-9.92^{***}$\\[-0.15em]{\scriptsize $p_{adj}=<1e-300$}\end{tabular} & \begin{tabular}{@{}c@{}}$-7.48^{***}$\\[-0.15em]{\scriptsize $p_{adj}=<1e-300$}\end{tabular} & \begin{tabular}{@{}c@{}}$-5.35^{***}$\\[-0.15em]{\scriptsize $p_{adj}=<1e-300$}\end{tabular} \\
 & CC & \begin{tabular}{@{}c@{}}$+0.035^{***}$\\[-0.15em]{\scriptsize $p_{adj}=<1e-300$}\end{tabular} & \begin{tabular}{@{}c@{}}$+0.024^{***}$\\[-0.15em]{\scriptsize $p_{adj}=3.4e-290$}\end{tabular} & \begin{tabular}{@{}c@{}}$+0.019^{***}$\\[-0.15em]{\scriptsize $p_{adj}=<1e-300$}\end{tabular} \\
\midrule
\multirow{3}{*}{TCDAE} & $\Delta$SNR & \begin{tabular}{@{}c@{}}$+2.67^{***}$\\[-0.15em]{\scriptsize $p_{adj}=<1e-300$}\end{tabular} & \begin{tabular}{@{}c@{}}$+3.80^{***}$\\[-0.15em]{\scriptsize $p_{adj}=<1e-300$}\end{tabular} & \begin{tabular}{@{}c@{}}$+2.90^{***}$\\[-0.15em]{\scriptsize $p_{adj}=<1e-300$}\end{tabular} \\
 & PRD [\%] & \begin{tabular}{@{}c@{}}$-9.17^{***}$\\[-0.15em]{\scriptsize $p_{adj}=<1e-300$}\end{tabular} & \begin{tabular}{@{}c@{}}$-11.60^{***}$\\[-0.15em]{\scriptsize $p_{adj}=<1e-300$}\end{tabular} & \begin{tabular}{@{}c@{}}$-9.71^{***}$\\[-0.15em]{\scriptsize $p_{adj}=<1e-300$}\end{tabular} \\
 & CC & \begin{tabular}{@{}c@{}}$+0.019^{***}$\\[-0.15em]{\scriptsize $p_{adj}=<1e-300$}\end{tabular} & \begin{tabular}{@{}c@{}}$+0.013^{***}$\\[-0.15em]{\scriptsize $p_{adj}=1.3e-263$}\end{tabular} & \begin{tabular}{@{}c@{}}$+0.018^{***}$\\[-0.15em]{\scriptsize $p_{adj}=<1e-300$}\end{tabular} \\
\midrule
\multirow{3}{*}{DeScoD-10} & $\Delta$SNR & \begin{tabular}{@{}c@{}}$+2.63^{***}$\\[-0.15em]{\scriptsize $p_{adj}=<1e-300$}\end{tabular} & \begin{tabular}{@{}c@{}}$+2.30^{***}$\\[-0.15em]{\scriptsize $p_{adj}=<1e-300$}\end{tabular} & \begin{tabular}{@{}c@{}}$+2.16^{***}$\\[-0.15em]{\scriptsize $p_{adj}=<1e-300$}\end{tabular} \\
 & PRD [\%] & \begin{tabular}{@{}c@{}}$-9.83^{***}$\\[-0.15em]{\scriptsize $p_{adj}=<1e-300$}\end{tabular} & \begin{tabular}{@{}c@{}}$-6.62^{***}$\\[-0.15em]{\scriptsize $p_{adj}=<1e-300$}\end{tabular} & \begin{tabular}{@{}c@{}}$-7.40^{***}$\\[-0.15em]{\scriptsize $p_{adj}=<1e-300$}\end{tabular} \\
 & CC & \begin{tabular}{@{}c@{}}$+0.039^{***}$\\[-0.15em]{\scriptsize $p_{adj}=<1e-300$}\end{tabular} & \begin{tabular}{@{}c@{}}$+0.021^{***}$\\[-0.15em]{\scriptsize $p_{adj}=<1e-300$}\end{tabular} & \begin{tabular}{@{}c@{}}$+0.028^{***}$\\[-0.15em]{\scriptsize $p_{adj}=<1e-300$}\end{tabular} \\
\midrule
\multirow{3}{*}{FIR Filter} & $\Delta$SNR & \begin{tabular}{@{}c@{}}$+7.06^{***}$\\[-0.15em]{\scriptsize $p_{adj}=<1e-300$}\end{tabular} & \begin{tabular}{@{}c@{}}$+8.65^{***}$\\[-0.15em]{\scriptsize $p_{adj}=<1e-300$}\end{tabular} & \begin{tabular}{@{}c@{}}$+6.75^{***}$\\[-0.15em]{\scriptsize $p_{adj}=<1e-300$}\end{tabular} \\
 & PRD [\%] & \begin{tabular}{@{}c@{}}$-36.80^{***}$\\[-0.15em]{\scriptsize $p_{adj}=<1e-300$}\end{tabular} & \begin{tabular}{@{}c@{}}$-41.70^{***}$\\[-0.15em]{\scriptsize $p_{adj}=<1e-300$}\end{tabular} & \begin{tabular}{@{}c@{}}$-33.53^{***}$\\[-0.15em]{\scriptsize $p_{adj}=<1e-300$}\end{tabular} \\
 & CC & \begin{tabular}{@{}c@{}}$+0.12^{***}$\\[-0.15em]{\scriptsize $p_{adj}=<1e-300$}\end{tabular} & \begin{tabular}{@{}c@{}}$+0.13^{***}$\\[-0.15em]{\scriptsize $p_{adj}=<1e-300$}\end{tabular} & \begin{tabular}{@{}c@{}}$+0.11^{***}$\\[-0.15em]{\scriptsize $p_{adj}=<1e-300$}\end{tabular} \\
\midrule
\multirow{3}{*}{SWT Denoising} & $\Delta$SNR & \begin{tabular}{@{}c@{}}$+7.40^{***}$\\[-0.15em]{\scriptsize $p_{adj}=<1e-300$}\end{tabular} & \begin{tabular}{@{}c@{}}$+9.87^{***}$\\[-0.15em]{\scriptsize $p_{adj}=<1e-300$}\end{tabular} & \begin{tabular}{@{}c@{}}$+6.98^{***}$\\[-0.15em]{\scriptsize $p_{adj}=<1e-300$}\end{tabular} \\
 & PRD [\%] & \begin{tabular}{@{}c@{}}$-36.50^{***}$\\[-0.15em]{\scriptsize $p_{adj}=<1e-300$}\end{tabular} & \begin{tabular}{@{}c@{}}$-45.73^{***}$\\[-0.15em]{\scriptsize $p_{adj}=<1e-300$}\end{tabular} & \begin{tabular}{@{}c@{}}$-32.92^{***}$\\[-0.15em]{\scriptsize $p_{adj}=<1e-300$}\end{tabular} \\
 & CC & \begin{tabular}{@{}c@{}}$+0.16^{***}$\\[-0.15em]{\scriptsize $p_{adj}=<1e-300$}\end{tabular} & \begin{tabular}{@{}c@{}}$+0.21^{***}$\\[-0.15em]{\scriptsize $p_{adj}=<1e-300$}\end{tabular} & \begin{tabular}{@{}c@{}}$+0.14^{***}$\\[-0.15em]{\scriptsize $p_{adj}=<1e-300$}\end{tabular} \\
\bottomrule
\end{tabular}
\vspace{0.5em}

\footnotesize{$^{***}p_{\mathrm{adj}} < 0.001$, $^{**}p_{\mathrm{adj}} < 0.01$, $^{*}p_{\mathrm{adj}} < 0.05$ after Holm--Bonferroni correction.}
\end{table}

\subsection{Extended downstream evaluation}
\label{app:downstream}

Tables~\ref{tab:ptbxl-macro-classifier} and \ref{tab:mitbih_binary_classifier} give the per-class AUROC and the binary beat-classification results summarized in Section~\ref{sec:results}; Table~\ref{tab:ptbxl-macro-classifier-appendix} adds sensitivity, specificity and F1. The threshold-dependent metrics show that the AUROC gains are not a pure ranking effect, but also that \method{} is not uniformly best on sensitivity and specificity, so the downstream benefit depends on the operating point. The PTB-XL classifier is an Inception1d model \citep{strotdhoff2021} trained on folds 1--8; MIT-BIH uses a CNN adapted from \citet{tahsin_leightweight} on $\pm144$-sample windows around annotated R-peaks, collapsed to normal versus abnormal. Because the classifier is single-lead, its absolute performance is below 12-lead references \citep{strotdhoff2021}; the comparison across restoration methods is unaffected. The extended Table~\ref{tab:placeholder_downstream}, introduced in Table \ref{tab:downstream} in Section \ref{sec:results}, widens the comparison to PPG, EMG and EEG, with clean and corrupted references. As for ECG, the task models are trained on clean signals and applied frozen to clean, corrupted and restored test signals, with bootstrap intervals clustered by clean segment since each appears once per corruption. For PPG, an Inception1d regressor \citep{strotdhoff2021} predicts the PPG-DaLiA reference heart rate (from the chest ECG, aligned to each 8\,s segment) on the restoration subject split; we report MAE with Pearson $r$ and $R^2$. For EMG, a small MLP on fixed amplitude features detects muscle activation, labelled from NinaPro's movement cue, in 1{,}024-sample windows of the DB2 test recordings (trained on subjects 1--8, tested on subject 10). Every input is divided by the root-mean-square of the clean training windows, and we report balanced accuracy since 71\% of windows are active. For EEG, we compute $\alpha$ (8--13\,Hz) and $\beta$ (13--30\,Hz) band power per epoch with Welch's method \citep{welch_use_1967} and report the relative error to the clean epoch, averaged over both bands.

\providecommand{\best}[1]{\begingroup\bfseries\boldmath #1\endgroup}
\providecommand{\secondbest}[1]{\underline{#1}}
\begin{table*}[t]
\centering
\caption{Per-superdiagnostic-class and macro-averaged AUROC on PTB-XL. Values are point estimates with 95\% bootstrap confidence intervals. Bold/underlined: best/second-best (excluding clean/noisy).}
\label{tab:ptbxl-macro-classifier}
\resizebox{\textwidth}{!}{%
\begin{tabular}{lcccccc}
\toprule
 & \multicolumn{5}{c}{Per-class AUROC} & Macro \\
\cmidrule(lr){2-6}
Model & CD & HYP & MI & NORM & STTC & AUROC \\
\midrule
clean & .871 $\pm$ .019 & .782 $\pm$ .031 & .845 $\pm$ .020 & .916 $\pm$ .012 & .878 $\pm$ .017 & .859 $\pm$ .011 \\
noisy & .830 $\pm$ .022 & .722 $\pm$ .034 & .753 $\pm$ .024 & .866 $\pm$ .015 & .822 $\pm$ .020 & .799 $\pm$ .012 \\
\midrule
FIR Filter & .828 $\pm$ .023 & .729 $\pm$ .034 & .758 $\pm$ .025 & .855 $\pm$ .015 & .818 $\pm$ .020 & .798 $\pm$ .012 \\
SWT Denoising & .788 $\pm$ .024 & .716 $\pm$ .032 & .658 $\pm$ .028 & .798 $\pm$ .019 & .772 $\pm$ .024 & .746 $\pm$ .013 \\
TCDAE & .855 $\pm$ .020 & .760 $\pm$ .033 & .829 $\pm$ .022 & .902 $\pm$ .013 & .853 $\pm$ .018 & .840 $\pm$ .012 \\
DeScoD & .854 $\pm$ .020 & .769 $\pm$ .034 & .826 $\pm$ .021 & .901 $\pm$ .013 & .853 $\pm$ .018 & .840 $\pm$ .012 \\
TFCDiff & .851 $\pm$ .021 & .760 $\pm$ .034 & .825 $\pm$ .022 & .895 $\pm$ .013 & .850 $\pm$ .019 & .836 $\pm$ .012 \\
\midrule
\method{}-MC-10 & \best{.864 $\pm$ .020} & \secondbest{.774 $\pm$ .032} & \secondbest{.836 $\pm$ .021} & \best{.906 $\pm$ .013} & \best{.860 $\pm$ .018} & \best{.848 $\pm$ .011} \\
\method{}-AV-10 & \secondbest{.863 $\pm$ .020} & \best{.775 $\pm$ .032} & \best{.836 $\pm$ .020} & \secondbest{.905 $\pm$ .012} & \secondbest{.858 $\pm$ .018} & \secondbest{.848 $\pm$ .011} \\
\bottomrule
\end{tabular}%
}
\end{table*}
\providecommand{\best}[1]{\begingroup\bfseries\boldmath #1\endgroup}
\providecommand{\secondbest}[1]{\underline{#1}}
\begin{table*}[t]
\centering
\caption{Downstream classification on MIT-BIH binary beat classification. Values are point estimates with 95\% bootstrap confidence intervals. Bold/underlined: best/second-best (excluding clean/noisy).}
\label{tab:mitbih_binary_classifier}

\begin{tabular}{lcccc}
\toprule
Model & AUROC & Specificity & Sensitivity & F1 \\
\midrule
clean & .926 $\pm$ .004 & .991 $\pm$ .001 & .580 $\pm$ .014 & .703 $\pm$ .012 \\
noisy & .718 $\pm$ .008 & .989 $\pm$ .001 & .151 $\pm$ .010 & .243 $\pm$ .014 \\
\midrule
FIR Filter & .734 $\pm$ .008 & \best{.990 $\pm$ .001} & .157 $\pm$ .010 & .254 $\pm$ .014 \\
SWT Denoising & .722 $\pm$ .008 & .971 $\pm$ .002 & .206 $\pm$ .011 & .286 $\pm$ .014 \\
TCDAE & .839 $\pm$ .007 & \secondbest{.979 $\pm$ .001} & .436 $\pm$ .013 & .543 $\pm$ .013 \\
DeScoD & .822 $\pm$ .007 & .919 $\pm$ .003 & .541 $\pm$ .014 & .490 $\pm$ .011 \\
TFCDiff & .803 $\pm$ .007 & .935 $\pm$ .002 & .460 $\pm$ .013 & .463 $\pm$ .012 \\
\midrule
\method{}-MC-10 & \secondbest{.880 $\pm$ .006} & .965 $\pm$ .002 & \secondbest{.579 $\pm$ .014} & \secondbest{.622 $\pm$ .012} \\
\method{}-AV-10 & \best{.883 $\pm$ .006} & .962 $\pm$ .002 & \best{.602 $\pm$ .013} & \best{.630 $\pm$ .011} \\
\bottomrule
\end{tabular}%
\end{table*}

\providecommand{\best}[1]{\begingroup\bfseries\boldmath #1\endgroup}
\providecommand{\secondbest}[1]{\underline{#1}}
\begin{table*}[htbp]
\centering
\caption{Per-superdiagnostic-class and macro-averaged sensitivity, specificity, and F1 on PTB-XL. Values are point estimates with 95\% bootstrap confidence intervals. Bold/underlined: best/second-best per row (excluding clean/noisy).}
\label{tab:ptbxl-macro-classifier-appendix}
\resizebox{\textwidth}{!}{%
\begin{tabular}{lccccccccc}
\toprule
Class & clean & noisy & FIR Filter & SWT Denoising & TCDAE & DeScoD & TFCDiff & \method{}-MC-10 & \method{}-AV-10 \\
\midrule
\multicolumn{10}{l}{\textit{Sensitivity}} \\
CD & .695 $\pm$ .041 & .592 $\pm$ .043 & \secondbest{.727 $\pm$ .040} & \best{.775 $\pm$ .035} & .641 $\pm$ .043 & .647 $\pm$ .041 & .683 $\pm$ .041 & .691 $\pm$ .039 & .679 $\pm$ .041 \\
HYP & .520 $\pm$ .066 & .143 $\pm$ .046 & .161 $\pm$ .050 & .184 $\pm$ .053 & .534 $\pm$ .066 & \best{.592 $\pm$ .067} & \secondbest{.583 $\pm$ .064} & .538 $\pm$ .065 & .543 $\pm$ .065 \\
MI & .662 $\pm$ .044 & .827 $\pm$ .036 & \secondbest{.880 $\pm$ .031} & \best{.954 $\pm$ .021} & .600 $\pm$ .048 & .585 $\pm$ .045 & .624 $\pm$ .042 & .621 $\pm$ .045 & .624 $\pm$ .045 \\
NORM & .892 $\pm$ .019 & .446 $\pm$ .033 & .425 $\pm$ .031 & .166 $\pm$ .024 & \secondbest{.893 $\pm$ .019} & .880 $\pm$ .020 & .860 $\pm$ .022 & \best{.905 $\pm$ .019} & \best{.905 $\pm$ .018} \\
STTC & .669 $\pm$ .044 & .862 $\pm$ .031 & \best{.675 $\pm$ .043} & \secondbest{.669 $\pm$ .041} & .614 $\pm$ .047 & .650 $\pm$ .045 & .638 $\pm$ .044 & .610 $\pm$ .046 & .606 $\pm$ .045 \\
Macro & .688 $\pm$ .022 & .574 $\pm$ .018 & .574 $\pm$ .018 & .550 $\pm$ .016 & .656 $\pm$ .024 & .671 $\pm$ .023 & \best{.677 $\pm$ .023} & \secondbest{.673 $\pm$ .022} & .671 $\pm$ .022 \\
\midrule
\multicolumn{10}{l}{\textit{Specificity}} \\
CD & .886 $\pm$ .015 & .905 $\pm$ .014 & .786 $\pm$ .019 & .618 $\pm$ .024 & \best{.899 $\pm$ .015} & \secondbest{.887 $\pm$ .015} & .863 $\pm$ .017 & .882 $\pm$ .015 & .881 $\pm$ .015 \\
HYP & .838 $\pm$ .017 & .959 $\pm$ .009 & \best{.962 $\pm$ .009} & \secondbest{.945 $\pm$ .011} & .824 $\pm$ .018 & .791 $\pm$ .019 & .790 $\pm$ .019 & .810 $\pm$ .018 & .800 $\pm$ .019 \\
MI & .827 $\pm$ .018 & .499 $\pm$ .023 & .438 $\pm$ .022 & .189 $\pm$ .020 & \best{.858 $\pm$ .017} & \secondbest{.855 $\pm$ .017} & .838 $\pm$ .018 & .852 $\pm$ .017 & .852 $\pm$ .017 \\
NORM & .789 $\pm$ .024 & .939 $\pm$ .014 & \secondbest{.935 $\pm$ .014} & \best{.990 $\pm$ .006} & .755 $\pm$ .026 & .783 $\pm$ .025 & .781 $\pm$ .025 & .765 $\pm$ .025 & .760 $\pm$ .025 \\
STTC & .894 $\pm$ .015 & .571 $\pm$ .024 & .787 $\pm$ .020 & .731 $\pm$ .022 & .880 $\pm$ .016 & .867 $\pm$ .016 & .862 $\pm$ .017 & \secondbest{.895 $\pm$ .015} & \best{.900 $\pm$ .015} \\
Macro & .847 $\pm$ .007 & .775 $\pm$ .008 & .781 $\pm$ .008 & .695 $\pm$ .008 & \best{.843 $\pm$ .007} & .837 $\pm$ .008 & .827 $\pm$ .008 & \secondbest{.841 $\pm$ .008} & .838 $\pm$ .008 \\
\midrule
\multicolumn{10}{l}{\textit{F1}} \\
CD & .677 $\pm$ .035 & .627 $\pm$ .036 & .607 $\pm$ .035 & .522 $\pm$ .031 & .654 $\pm$ .034 & .647 $\pm$ .035 & .647 $\pm$ .034 & \best{.671 $\pm$ .034} & \secondbest{.661 $\pm$ .033} \\
HYP & .365 $\pm$ .048 & .193 $\pm$ .058 & .220 $\pm$ .061 & .225 $\pm$ .058 & \secondbest{.358 $\pm$ .048} & \best{.358 $\pm$ .046} & .352 $\pm$ .044 & .347 $\pm$ .047 & .340 $\pm$ .045 \\
MI & .566 $\pm$ .037 & .436 $\pm$ .030 & .430 $\pm$ .030 & .371 $\pm$ .026 & .556 $\pm$ .038 & .543 $\pm$ .037 & .551 $\pm$ .037 & \secondbest{.564 $\pm$ .038} & \best{.565 $\pm$ .039} \\
NORM & .836 $\pm$ .018 & .588 $\pm$ .031 & .567 $\pm$ .030 & .283 $\pm$ .035 & .821 $\pm$ .018 & .826 $\pm$ .018 & .814 $\pm$ .019 & \best{.832 $\pm$ .018} & \secondbest{.830 $\pm$ .017} \\
STTC & .672 $\pm$ .034 & .544 $\pm$ .029 & .581 $\pm$ .033 & .538 $\pm$ .034 & .621 $\pm$ .038 & \secondbest{.633 $\pm$ .035} & .620 $\pm$ .037 & .633 $\pm$ .036 & \best{.635 $\pm$ .036} \\
Macro & .623 $\pm$ .016 & .478 $\pm$ .018 & .481 $\pm$ .019 & .388 $\pm$ .018 & .602 $\pm$ .018 & .601 $\pm$ .018 & .597 $\pm$ .016 & \best{.609 $\pm$ .018} & \secondbest{.606 $\pm$ .017} \\
\bottomrule
\end{tabular}%
}
\end{table*}

\begin{table*}[t]
\centering
\caption{Downstream utility across all four modalities.}
\label{tab:placeholder_downstream}
\resizebox{\textwidth}{!}{%
\begin{tabular}{llcccc}
\toprule
Modality & Task (metric) & Clean & Corrupted & Best baseline & \method-AV-10 \\
\midrule
PPG & HR estimation (MAE, bpm) $\downarrow$        & $4.772 $ & $8.483$ & $5.947$ & $\mathbf{5.155}$ \\
ECG & PTB-XL superdiagnostic (macro AUROC) $\uparrow$   & $.859$ & $.799$ & $.840$ & $\mathbf{.848}$ \\
ECG & MIT-BIH beat classification (F1) $\uparrow$       & $.703$ & $.243$ & $.543$ & $\mathbf{.630}$ \\
EMG & NinaPro DB2 muscle activation (balanced acc.) $\uparrow$ & $.755$& $.501$& $.749$ & $\mathbf{.750}$ \\
EEG & Band-power preservation ($\alpha,\beta$ rel.\ error, \%) $\downarrow$ &  $0.00$& $4.75$ & $4.56$ & $\mathbf{3.31}$ \\
\bottomrule
\end{tabular}%
}
\end{table*}
% \subsection{Synthetic Cyclostationarity Sweep}
% \label{app:jitter}

% \ph{\texttt{placeholder\_fig\_jitter\_sweep}. The four-modality evidence of Section~\ref{sec:results-scaling} establishes an ordering across four points where $\Pi$ co-varies with sampling rate, corruption type, morphology and dataset size. A controlled study isolates the variable. Protocol: synthesize signals as a fixed template convolved with an event train whose inter-event intervals are drawn with controlled jitter $\sigma_{\mathrm{jit}}$, plus a band-limited aperiodic component with controllable variance share; sweep both to trace $\Pi$ across $[0,1]$ with all else fixed; train the full model and the no-phase ablation at each setting; plot realized gain against measured $\Pi$. The prediction is a monotone curve through the origin, upper-bounded by Proposition~\ref{prop:mmse}. The four real modalities then provide external validity for the synthetic result.}
\newpage

\subsection{Component ablation across modalities}
\label{app:ablation-modalities}

Table~\ref{tab:placeholder_ablation_modalities} replicates the ECG ablation of Table~\ref{tab:ablation} in every modality, retraining without one component at a time (Section~\ref{sec:results}). Removing phase costs most on PPG ($-1.82$\,dB) and ECG ($-0.64$\,dB) and little on EEG and EMG, whereas removing the frame costs $0.5$--$0.7$\,dB on ECG, EMG and EEG but slightly helps PPG.

\begin{table}[htbp]
\centering
\caption{Component ablation replicated across modalities. Cells report the change in $\Delta$SNR relative to the full model on the same dataset as mean $\pm$ 95\% bootstrap CI. The analytical phase row is reported only for ECG, where the detector comparison is part of the main mechanism analysis. Values use one antithetic-variates pair.}
\label{tab:placeholder_ablation_modalities}
\resizebox{\textwidth}{!}{%
\begin{tabular}{lcccc}
\toprule
Ablation ($\Delta$ from full model, dB) & PPG & ECG & EMG & EEG \\
\midrule
$-$ Phase conditioning & $-1.82 \pm 0.03$ & $-0.64 \pm 0.05$ & $+0.05 \pm 0.01$ & $-0.24 \pm 0.01$ \\
$-$ Temporal context & $+0.28 \pm 0.02$ & $-0.72 \pm 0.07$ & $-0.35 \pm 0.01$ & $-0.27 \pm 0.01$ \\
$-$ Shift-covariant frame & $+0.37 \pm 0.03$ & $-0.51 \pm 0.05$ & $-0.51 \pm 0.01$ & $-0.74 \pm 0.02$ \\
$-$ Auxiliary losses & $-0.59 \pm 0.02$ & $-0.48 \pm 0.05$ & $+0.02 \pm 0.01$ & $-2.80 \pm 0.04$ \\
Analytical phase (vs.\ learned) & -- & $-0.66 \pm 0.05$ & -- & -- \\
\bottomrule
\end{tabular}
}
\end{table}

\subsection{Cyclostationarity index across datasets}
\label{app:cyclicity}

Section~\ref{sec:results} relates the gain from phase conditioning to how strongly each signal is organized by its cycle. Table~\ref{tab:cyclicity-index} measures this on the clean test sets, using the autocorrelation proxy $\Pi_{\mathrm{AC}}$ for all nine datasets and the detector-based $\hat{\Pi}$ for ECG (Appendix~\ref{app:cyclo-estimator}). Averaged per modality, $\Pi_{\mathrm{AC}}$ orders the modalities PPG ($0.74$) $>$ ECG ($0.71$) $>$ EEG ($0.37$) $>$ EMG ($0.10$), the same order as the cost of removing phase conditioning in Table~\ref{tab:placeholder_ablation_modalities}.

\begin{table}[t]
\centering
\small
\setlength{\tabcolsep}{5pt}
\begin{tabular}{llccc}
\toprule
Modality & Dataset & AC window [Hz] & $\Pi_{\mathrm{AC}}$ [95\% CI] & $\hat{\Pi}$ [95\% CI] \\
\midrule
PPG & PPG-DaLiA               & 0.5--3 & $0.62$ [$0.60$, $0.63$] & -- \\
    & BIDMC$^{\dagger}$       & 0.5--3 & $0.85$ [$0.81$, $0.89$] & -- \\
    & \textit{Mean}           &        & $0.74$ & -- \\
\midrule
ECG & PTB-XL                  & 0.5--3 & $0.71$ [$0.70$, $0.72$] & $0.87$ [$0.86$, $0.87$] \\
    & QTDB$^{\dagger}$        & 0.5--3 & $0.81$ [$0.78$, $0.83$] & $0.94$ [$0.92$, $0.95$] \\
    & MIT-BIH$^{\dagger}$     & 0.5--3 & $0.61$ [$0.56$, $0.66$] & $0.80$ [$0.70$, $0.87$] \\
    & \textit{Mean}           &        & $0.71$ & $0.87$ \\
\midrule
EEG & EEGdenoiseNet           & 1--30  & $0.34$ [$0.33$, $0.35$] & -- \\
    & EEGMMIDB$^{\dagger}$    & 1--30  & $0.40$ [$0.38$, $0.42$] & -- \\
    & \textit{Mean}           &        & $0.37$ & -- \\
\midrule
EMG & NinaPro DB2             & 5--50  & $0.11$ [$0.10$, $0.12$] & -- \\
    & NinaPro DB3$^{\dagger}$ & 5--50  & $0.09$ [$0.09$, $0.10$] & -- \\
    & \textit{Mean}           &        & $0.10$ & -- \\
\bottomrule
\end{tabular}
\caption{Cyclostationarity of the clean test sets, ordered by the mean $\Pi_{\mathrm{AC}}$, which is the
autocorrelation proxy of Appendix~\ref{app:cyclo-estimator}, computed identically for
every dataset in a per-modality lag window fixed before analysis. $\hat{\Pi}$
(Definition~\ref{def:pi}) requires a reliable cycle detector and is reported for ECG
only (Table~\ref{tab:analytical-encoder}). $^{\dagger}$Held-out datasets, evaluated
without retraining (Table~\ref{tab:crossdataset-modalities}); \textit{Mean} rows average
the datasets of each modality. Because the proxy takes its peak at a single lag within
each window, beat-to-beat rate variability lowers it, and it reads below $\hat{\Pi}$ on ECG.
Brackets are 95\% cluster-bootstrap CIs.}
\label{tab:cyclicity-index}
\end{table}

\subsection{Synthetic Cyclostationarity Sweep}
\label{app:jitter}

We isolate cyclostationarity from modality-specific confounders using synthetic signals composed of a fixed cyclic template and band-limited aperiodic background:
\begin{equation}
x_0=\mathcal{N}\!\left(\sqrt{c}\,x_{\mathrm{cyc}}
+\sqrt{1-c}\,x_{\mathrm{ap}}\right),
\end{equation}
where $c\in\{0,0.4,0.6,0.8,0.95\}$ is the nominal cyclic variance share. Cycle intervals have mean $P$ and coefficient of variation $\sigma_{\mathrm{jit}}\in\{0,0.03,0.07,0.15,0.30\}$. We train the full model and no-periodicity ablation at each of the 21 distinct conditions using three seeds. The measured gain is
$\Delta\mathrm{SNR}_{\mathrm{full}}-\Delta\mathrm{SNR}_{\mathrm{no\mbox{-}period}}$.
Figure~\ref{fig:jitter_sweep} shows that stronger measured cyclostationarity predicts a larger benefit from periodicity conditioning ($r=0.85$ overall). The relationship also holds within fixed-share groups ($r=0.82$--$0.88$), indicating that it is not driven solely by changing $c$. These results support the central prediction that periodicity conditioning becomes more useful as phase explains a larger fraction of signal structure. Curves are descriptive guides only and do not imply a particular functional relationship.

%\begin{figure}
%    \centering
    %\includegraphics[width=1.0\linewidth]{figures/jitter_sweep.png}
    %\caption{Synthetic cyclicity sweep relating the cyclostationarity index $\Pi$ to the periodicity-encoder gain in $\Delta$SNR, measured relative to the no-periodicity ablation. Marker shape denotes the cyclic share $c$, while color denotes temporal jitter $\sigma_{\mathrm{jit}}$ from light to dark purple. Error bars show 95\% paired bootstrap confidence intervals across training seeds and test samples. Dark curves show within-share trends, with the corresponding Pearson correlations reported in the legend; the light gray line shows the overall linear trend.}
    %\label{fig:jitter_sweep}
%\end{figure}

\subsection{Frame and objective selection}
\label{app:swt-loss-ablation}

Table~\ref{tab:swt-ablation} sweeps wavelet family and decomposition level; Table~\ref{tab:loss-ablation} sweeps the auxiliary loss weights. The frame choice matters more than the loss weighting: $\Delta$SNR ranges from $14.58$ (coif3, $J=8$) to $15.98$ (sym4, $J=4$) across the frame sweep, against a narrower band across loss weights. Performance degrades consistently for $J>4$ at every wavelet family, as deeper decompositions push the approximation band below the signal's fundamental and the coarsest channel stops carrying cycle-scale morphology. These sweeps support the frame of Section~\ref{sec:method-frame} ($J=4$, \texttt{sym4}) and the loss weights of Equation~\ref{eq:loss-aux}.

\providecommand{\best}[1]{\begingroup\bfseries\boldmath #1\endgroup}
\providecommand{\secondbest}[1]{\underline{#1}}
\begin{table*}[htbp]
\centering
\caption{\method{} SWT ablation on PTB-XL: $\Delta$SNR (dB, higher better) and PRD (\%, lower better) per (wavelet, decomposition level J), single antithetic-variate (AV) pair. Cells are mean $\pm$ 95\% CI; best per metric in \textbf{bold}.}
\label{tab:swt-ablation}
\resizebox{\textwidth}{!}{%
\begin{tabular}{cclccccccccc}
\toprule
 &  & \multirow{2}{*}{Metric} & \multicolumn{3}{c}{Daubechies} & \multicolumn{3}{c}{Symlets} & \multicolumn{3}{c}{Coiflets} \\
\cmidrule(lr){4-6} \cmidrule(lr){7-9} \cmidrule(lr){10-12}
 &  &  & db$_{4}$ & db$_{6}$ & db$_{8}$ & sym$_{4}$ & sym$_{6}$ & sym$_{8}$ & coif$_{3}$ & coif$_{4}$ & coif$_{5}$ \\
\midrule
\multirow{8}{*}{\rotatebox[origin=c]{90}{$J$}} & \multirow{2}{*}{2} & $\Delta$SNR\,$\uparrow$ & $15.57 \pm 0.21$ & $15.46 \pm 0.20$ & $15.16 \pm 0.21$ & $15.13 \pm 0.21$ & $15.57 \pm 0.21$ & $15.47 \pm 0.20$ & $15.21 \pm 0.20$ & $15.39 \pm 0.21$ & $15.38 \pm 0.21$ \\
 &  & PRD\,$\downarrow$ & $24.21 \pm 0.62$ & $24.49 \pm 0.58$ & $25.80 \pm 0.69$ & $25.13 \pm 0.57$ & $24.01 \pm 0.58$ & $24.43 \pm 0.60$ & $24.79 \pm 0.56$ & $24.91 \pm 0.64$ & $24.54 \pm 0.58$ \\
\addlinespace
 & \multirow{2}{*}{4} & $\Delta$SNR\,$\uparrow$ & $15.67 \pm 0.21$ & $15.50 \pm 0.21$ & $15.35 \pm 0.21$ & \best{$15.98 \pm 0.21$} & $15.76 \pm 0.21$ & $15.60 \pm 0.21$ & $15.49 \pm 0.21$ & $15.43 \pm 0.21$ & $15.41 \pm 0.21$ \\
 &  & PRD\,$\downarrow$ & $23.62 \pm 0.56$ & $24.09 \pm 0.54$ & $24.54 \pm 0.56$ & \best{$22.73 \pm 0.50$} & $23.33 \pm 0.52$ & $24.14 \pm 0.62$ & $24.28 \pm 0.58$ & $24.33 \pm 0.57$ & $24.43 \pm 0.56$ \\
\addlinespace
 & \multirow{2}{*}{6} & $\Delta$SNR\,$\uparrow$ & $15.07 \pm 0.20$ & $14.86 \pm 0.21$ & $14.91 \pm 0.21$ & $15.24 \pm 0.21$ & $14.93 \pm 0.21$ & $15.07 \pm 0.20$ & $14.86 \pm 0.20$ & $14.89 \pm 0.20$ & $14.95 \pm 0.20$ \\
 &  & PRD\,$\downarrow$ & $25.20 \pm 0.55$ & $25.72 \pm 0.55$ & $25.52 \pm 0.54$ & $24.66 \pm 0.54$ & $25.70 \pm 0.57$ & $25.14 \pm 0.54$ & $25.77 \pm 0.55$ & $25.60 \pm 0.55$ & $25.36 \pm 0.54$ \\
\addlinespace
 & \multirow{2}{*}{8} & $\Delta$SNR\,$\uparrow$ & $14.88 \pm 0.20$ & $14.71 \pm 0.21$ & $14.65 \pm 0.21$ & $14.85 \pm 0.20$ & $14.87 \pm 0.20$ & $14.80 \pm 0.20$ & $14.58 \pm 0.21$ & $14.67 \pm 0.20$ & $14.78 \pm 0.20$ \\
 &  & PRD\,$\downarrow$ & $25.59 \pm 0.52$ & $26.10 \pm 0.54$ & $26.32 \pm 0.56$ & $25.65 \pm 0.52$ & $25.56 \pm 0.53$ & $25.78 \pm 0.53$ & $26.49 \pm 0.54$ & $26.19 \pm 0.55$ & $25.73 \pm 0.52$ \\
\bottomrule
\end{tabular}%
}
\end{table*}\providecommand{\best}[1]{\begingroup\bfseries\boldmath #1\endgroup}
\providecommand{\secondbest}[1]{\underline{#1}}
\begin{table*}[htbp]
\centering
\caption{\method{} auxiliary-loss weighting ablation on PTB-XL: $\Delta$SNR (dB, higher better) and PRD (\%, lower better) per (morphology, reconstruction) loss weight, single antithetic-variate (AV) pair. Cells are mean $\pm$ 95\% CI; best per metric in \textbf{bold}. Ablations use the full model with a db4 wavelet and 4 decomposition levels.}
\label{tab:loss-ablation}
\resizebox{\textwidth}{!}{%
\begin{tabular}{cclcccccc}
\toprule
 &  & \multirow{2}{*}{Metric} & \multicolumn{6}{c}{$\lambda_{\mathrm{reconstruction}}$} \\
\cmidrule(lr){4-9}
 &  &  & 0.05 & 0.1 & 0.2 & 0.3 & 0.5 & 1 \\
\midrule
\multirow{12}{*}{\rotatebox[origin=c]{90}{$\lambda_{\mathrm{morphology}}$}} & \multirow{2}{*}{0.05} & $\Delta$SNR\,$\uparrow$ & $15.61 \pm 0.20$ & $15.65 \pm 0.21$ & $15.57 \pm 0.21$ & $15.66 \pm 0.20$ & $15.74 \pm 0.21$ & $15.57 \pm 0.21$ \\
 &  & PRD\,$\downarrow$ & $23.74 \pm 0.53$ & $23.70 \pm 0.56$ & $24.16 \pm 0.62$ & $23.78 \pm 0.57$ & $23.48 \pm 0.54$ & $24.03 \pm 0.58$ \\
\addlinespace
 & \multirow{2}{*}{0.1} & $\Delta$SNR\,$\uparrow$ & $15.54 \pm 0.21$ & $15.57 \pm 0.21$ & $15.30 \pm 0.21$ & \best{$15.79 \pm 0.19$} & $15.59 \pm 0.21$ & $15.00 \pm 0.21$ \\
 &  & PRD\,$\downarrow$ & $23.98 \pm 0.55$ & $23.86 \pm 0.54$ & $24.54 \pm 0.54$ & \best{$23.29 \pm 0.54$} & $23.84 \pm 0.54$ & $25.25 \pm 0.54$ \\
\addlinespace
 & \multirow{2}{*}{0.2} & $\Delta$SNR\,$\uparrow$ & $15.47 \pm 0.21$ & $15.61 \pm 0.22$ & $15.60 \pm 0.21$ & $15.65 \pm 0.21$ & $15.66 \pm 0.21$ & $15.65 \pm 0.21$ \\
 &  & PRD\,$\downarrow$ & $24.14 \pm 0.52$ & $23.79 \pm 0.54$ & $23.89 \pm 0.56$ & $23.87 \pm 0.59$ & $23.92 \pm 0.60$ & $23.68 \pm 0.55$ \\
\addlinespace
 & \multirow{2}{*}{0.3} & $\Delta$SNR\,$\uparrow$ & $15.55 \pm 0.21$ & $15.66 \pm 0.20$ & $15.61 \pm 0.21$ & $15.13 \pm 0.21$ & $15.52 \pm 0.20$ & $15.70 \pm 0.20$ \\
 &  & PRD\,$\downarrow$ & $23.96 \pm 0.53$ & $23.66 \pm 0.55$ & $23.87 \pm 0.56$ & $24.92 \pm 0.54$ & $24.11 \pm 0.57$ & $23.68 \pm 0.57$ \\
\addlinespace
 & \multirow{2}{*}{0.5} & $\Delta$SNR\,$\uparrow$ & $15.55 \pm 0.21$ & $15.79 \pm 0.20$ & $15.64 \pm 0.21$ & $15.55 \pm 0.20$ & $15.65 \pm 0.21$ & $15.53 \pm 0.20$ \\
 &  & PRD\,$\downarrow$ & $23.95 \pm 0.55$ & $23.30 \pm 0.55$ & $23.80 \pm 0.56$ & $24.05 \pm 0.57$ & $23.95 \pm 0.60$ & $24.18 \pm 0.58$ \\
\addlinespace
 & \multirow{2}{*}{1} & $\Delta$SNR\,$\uparrow$ & $15.57 \pm 0.21$ & $15.72 \pm 0.21$ & $15.67 \pm 0.21$ & $15.61 \pm 0.21$ & $15.65 \pm 0.21$ & $15.62 \pm 0.21$ \\
 &  & PRD\,$\downarrow$ & $23.91 \pm 0.54$ & $23.35 \pm 0.52$ & $23.65 \pm 0.54$ & $23.82 \pm 0.56$ & $23.77 \pm 0.57$ & $23.94 \pm 0.58$ \\
\bottomrule
\end{tabular}%
}
\end{table*}
\subsection{Phase-shuffle experiment}
\label{app:shuffle}

\begin{figure}[htbp]
\centering
\includegraphics[width=0.7\linewidth]{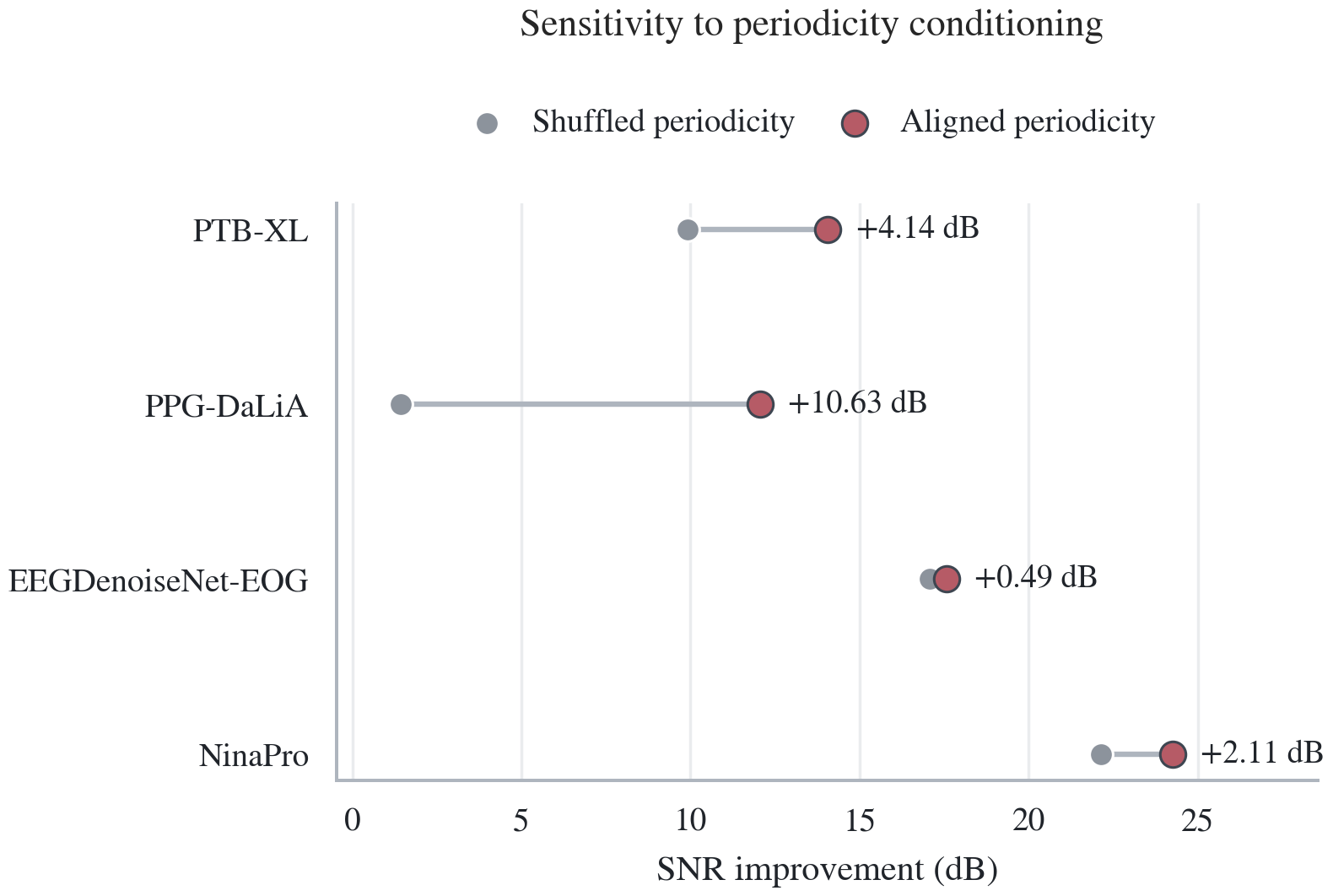}
\caption{Change in restoration performance when phase-encoder features are shuffled across samples within a batch. Values are the $\Delta$SNR difference between correctly matched and shuffled phase fields.}
\label{fig:shuffle}
\end{figure}

Figure~\ref{fig:shuffle} gives the raw phase-shuffle measurements underlying Figure~\ref{fig:scaling}a. At evaluation time the phase-encoder outputs are permuted across samples within a batch while corrupted inputs are left intact, so the model receives a structurally valid but mismatched phase field. All four modalities show a positive drop, confirming that the conditioning channel is used rather than ignored.

The shuffled condition is not equivalent to removing phase conditioning. A mismatched field is actively misleading whereas an absent one is merely uninformative, so $\Delta_{\mathrm{shuf}}$ upper-bounds the ablation gap: $+4.14$\,dB shuffled against $0.64$\,dB ablated on ECG (Table~\ref{tab:ablation}). It therefore measures reliance on the phase channel, the quantity plotted in Figure~\ref{fig:scaling}, and is available for all four modalities; Appendix~\ref{app:ablation-modalities} instead shows the retrained ablation.

\subsection{Antithetic sampling}
\label{app:av-empirical}

\begin{figure}[htbp]
    \centering
    \includegraphics[width=1.0\linewidth]{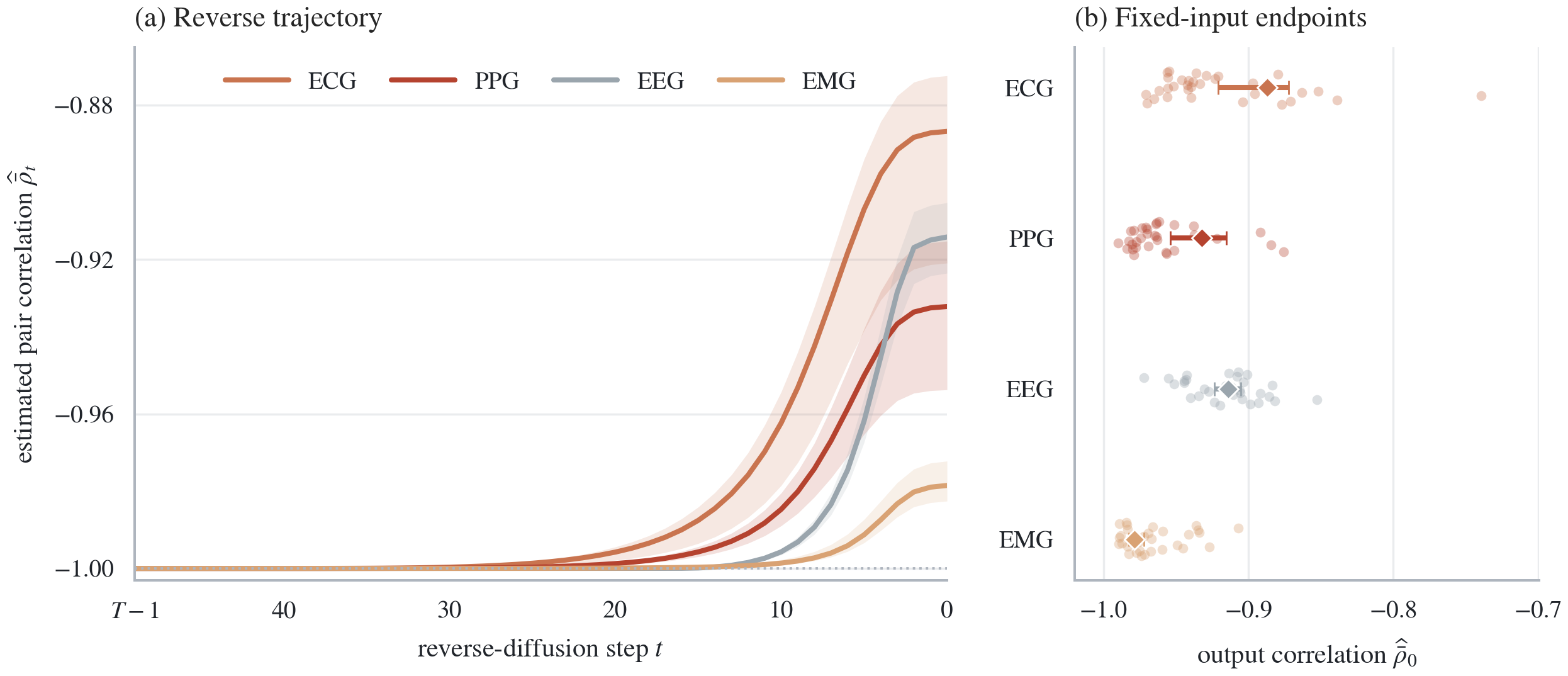}
    \caption{Empirical antithetic-pair correlation. \textbf{(a)} Variance-weighted correlation $\widehat{\bar{\rho}}_t$ over the 50 post-update states $t=T-1,\ldots,0$, with 95\% confidence intervals. \textbf{(b)} Output correlation $\widehat{\bar{\rho}}_0$: transparent points are single fixed observations, diamonds are modality aggregates with 95\% confidence intervals.}
    \label{fig:rho_measurement}
\end{figure}

\begin{figure}[htbp]
\centering
\includegraphics[width=\linewidth]{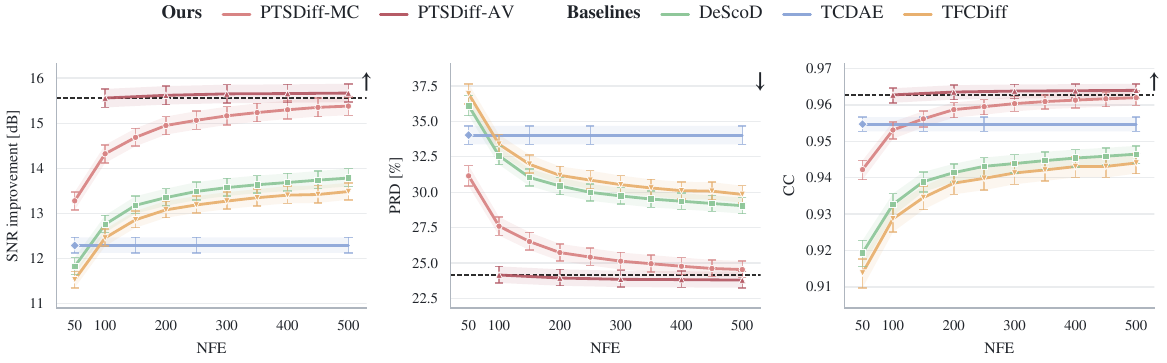}
\caption{Quality--compute frontier on ECG. PRD, CC and $\Delta$SNR against NFE, for independent (MC) and antithetically coupled (AV) sampling. The dotted line marks a single antithetic pair, which already exceeds ten independent trajectories. TCDAE is deterministic and therefore flat. Arrows indicate the better direction.}
\label{fig:shots}
\end{figure}

Proposition~\ref{prop:av} (proof in Appendix~\ref{app:proof-av}) guarantees a variance reduction at matched NFE only when the pair correlation $\bar{\rho}$ is negative, which a learned nonlinear reverse process does not ensure, so we measure it directly. For each modality, we hold 32 corrupted observations fixed and draw 128 antithetic pairs per observation. At every reverse state, the covariance and marginal variance of the pair members are estimated across pairs separately for each observation, then aggregated over signal coordinates and observations with the variance weighting of Equation~\ref{eq:mse-decomp}, giving $\widehat{\bar{\rho}}_t$. Confidence intervals come from a hierarchical bootstrap over observations and, within each, independent blocks of 16 pairs. The correlation remains negative throughout the reverse process and at $t=0$ for every modality and fixed observation evaluated (Figure~\ref{fig:rho_measurement}), so the condition of Proposition~\ref{prop:av} holds. Figure~\ref{fig:shots} shows the resulting quality--compute frontier on ECG: a single antithetic pair already exceeds ten independent trajectories, and further pairs add little once the model-bias term of Equation~\ref{eq:mse-decomp} dominates. Table~\ref{tab:placeholder_av_transfer} applies the same sampler swap to DeScoD-ECG and TFCDiff without retraining: one antithetic pair exceeds ten independent trajectories for both at a fifth of the wall-clock time, so the gain comes from the sampler rather than from \method{}.

\begin{table}[htbp]
\centering
\caption{Antithetic coupling applied to competing diffusion restorers without retraining (sampler swap only), with matched-NFE compute accounting. Establishes whether Proposition~\ref{prop:av} is a property of the sampler or of \method.}
\label{tab:placeholder_av_transfer}
\begin{tabular}{lccccc}
\toprule
Model / sampler & NFE & Params [M] & Wall-clock [ms] & $\Delta$SNR $\uparrow$ & $\Delta$ vs.\ MC \\
\midrule
DeScoD-ECG, MC-10   & $500$ & $1.93$ & $361$ & $13.78$ & --- \\
DeScoD-ECG, AV-2    & $100$ & $1.93$ & $72$ & $13.86$ & $+0.08$ \\
TFCDiff, MC-10      & $500$ & $4.12$ & $803$ & $13.47$ & --- \\
TFCDiff, AV-2       & $100$ & $4.12$ & $161$ & $13.67$ & $+0.20$ \\
\method, MC-10      & $500$     & $3.44$ & $267$ & $15.79$ & --- \\
\method, AV-2       & $100$     & $3.44$ & $53$ & $16.00$ & $+0.21$ \\
\bottomrule
\end{tabular}
\end{table}

\subsection{Per-sample distribution}
\label{app:persample}

\begin{figure}[htbp]
\centering
\includegraphics[width=\linewidth]{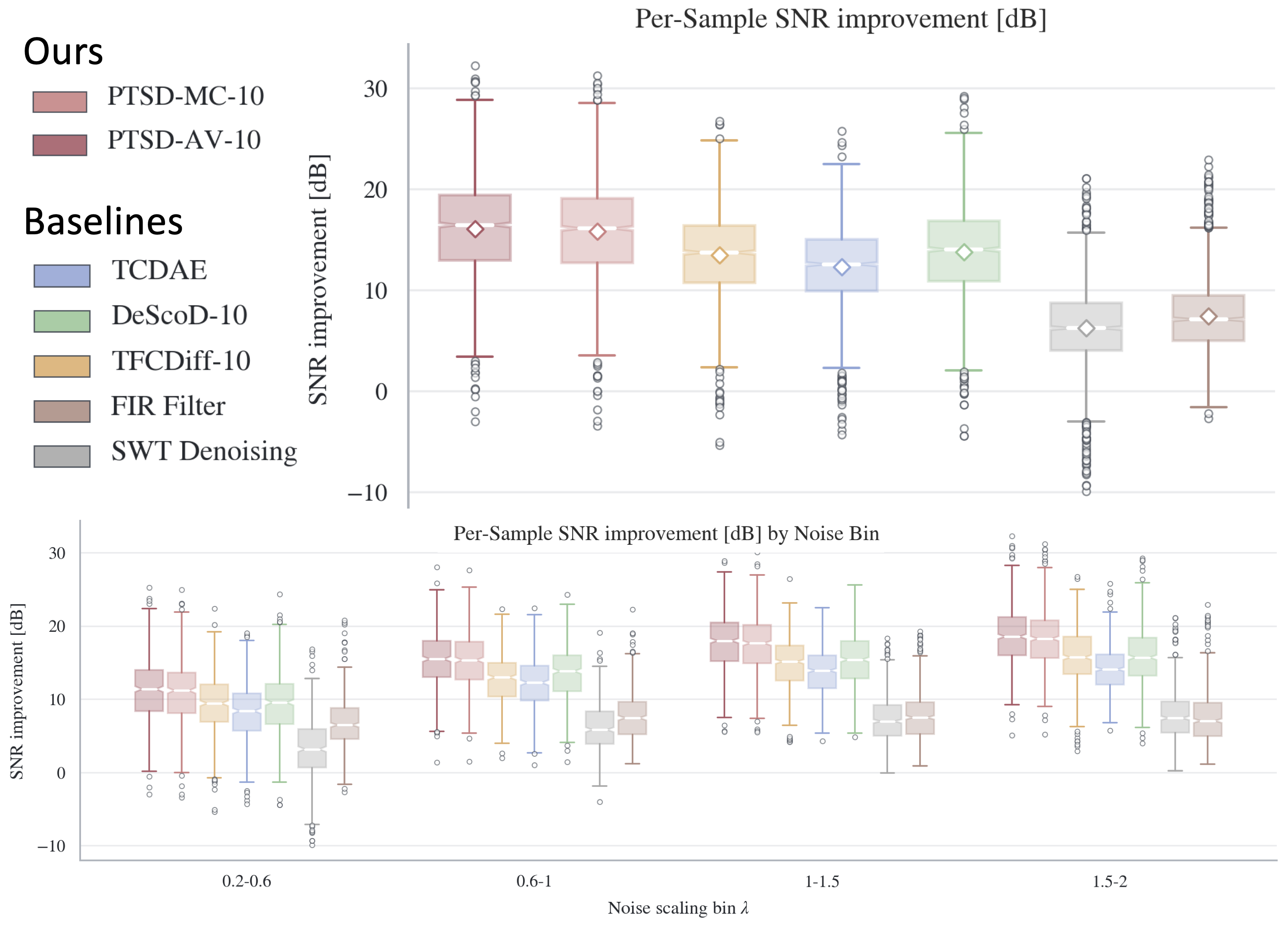}
\caption{Per-sample $\Delta$SNR on the PTB-XL test set. Notched box plots: box is the IQR, white line the median, notch its 95\% CI, whiskers span $1.5\times$IQR, open circles are outliers. \textbf{Top:} overall (white diamond = mean). \textbf{Bottom:} stratified by corruption scaling factor $\lambda$.}
\label{fig:persample}
\end{figure}

Figure~\ref{fig:persample} complements the PTB-XL means of Table~\ref{tab:denoise-ecg-combined} with the full per-sample $\Delta$SNR distribution, overall and stratified by corruption level, confirming that the mean improvements reflect a shifted distribution rather than a heavy tail.

\subsection{The frame decomposition}
\label{app:swt-example}

\begin{figure}[htbp]
\centering
\includegraphics[width=\textwidth]{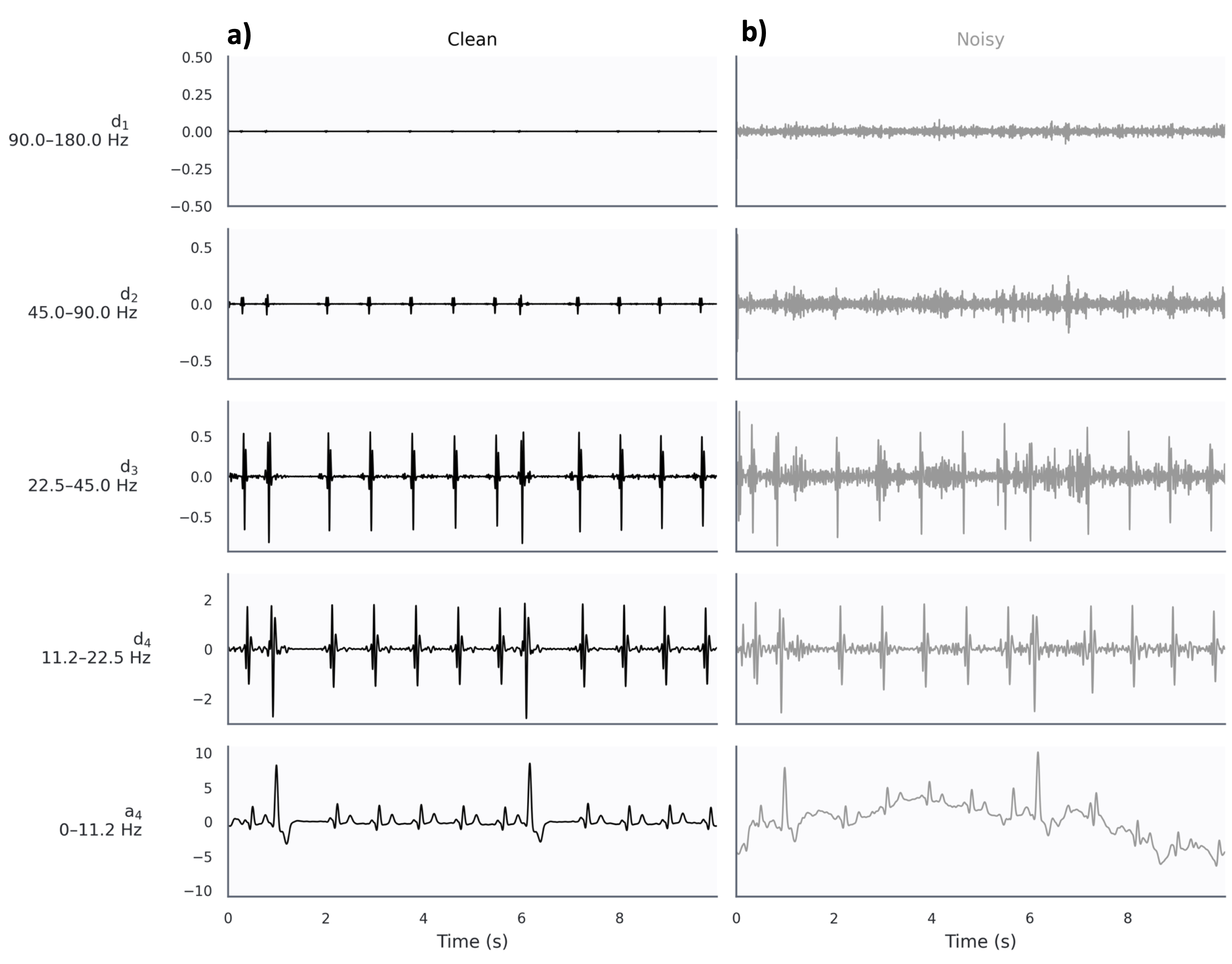}
\caption{Four-level stationary wavelet decomposition with the \texttt{sym4} wavelet: one approximation and four detail channels, all at the input length. \textbf{(a)} Clean signal. \textbf{(b)} Corrupted signal.}
\label{fig:swt-example}
\end{figure}

Figure~\ref{fig:swt-example} illustrates the frame $\mathcal{W}$ of Section~\ref{sec:method-frame} on a clean and a corrupted ECG at 360\,Hz. Baseline wander concentrates in the approximation channel $a_4$, while electrode motion and muscle artifacts spread across the detail channels \citep{chatterjee_review_2020}, which is why $d_1$ and $d_2$ carry predominantly artifact energy.

\subsection{Qualitative results}
\label{app:qualitative}

\begin{figure}[htbp]
\centering
\includegraphics[width=\linewidth]{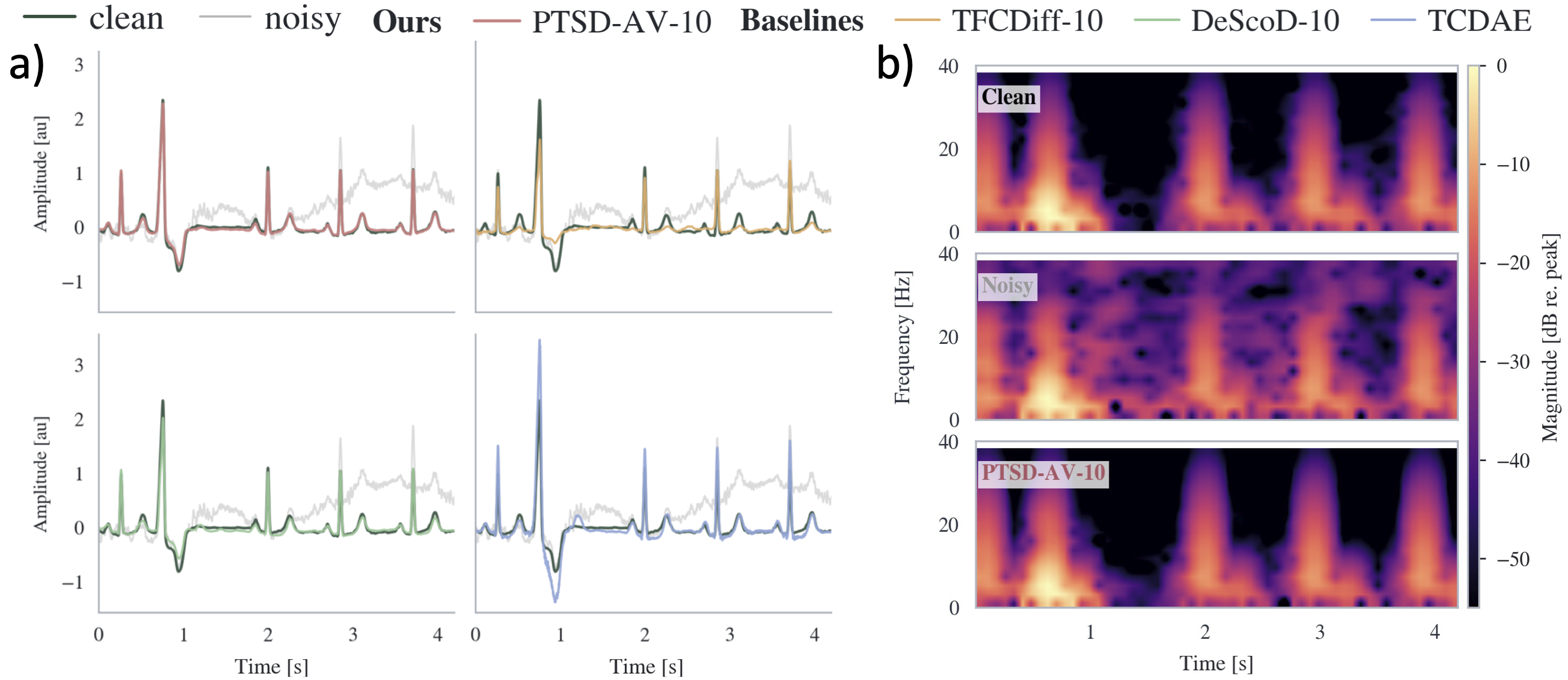}
\caption{ECG under heavy corruption. \textbf{Left:} clean, corrupted, and reconstructions from \method-AV-10 and the strongest baselines on a representative PTB-XL test sample. \textbf{Right:} time--frequency representations of the clean (top), corrupted (middle) and \method-AV-10 reconstructed (bottom) signal.}
\label{fig:ecg-qualitative-high}
\end{figure}

\begin{figure}[htbp]
\centering
\includegraphics[width=\linewidth]{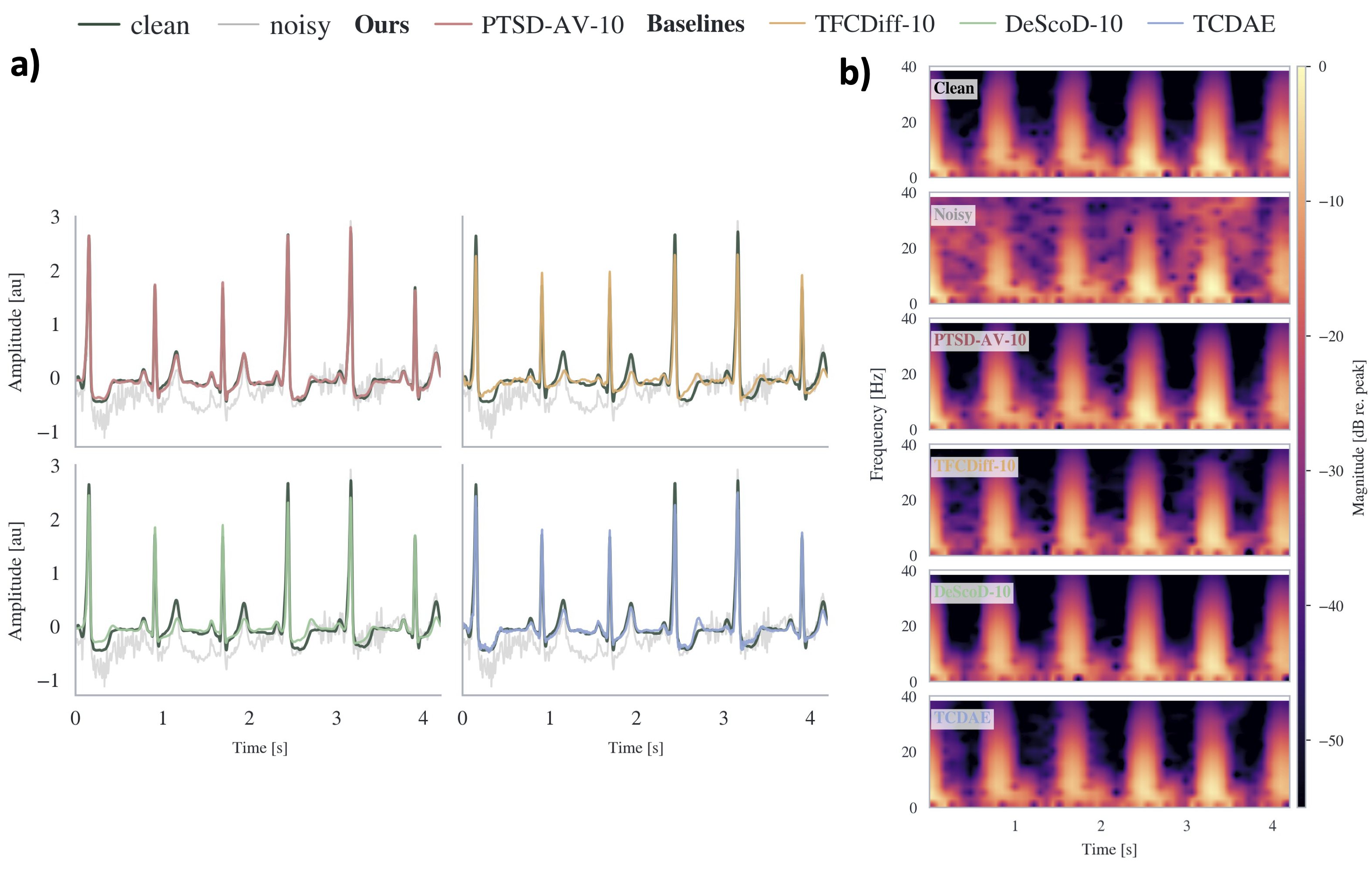}
\caption{ECG under light corruption ($\lambda \in [0.2,0.6]$). \textbf{(a)} Clean, corrupted, and reconstructions. \textbf{(b)} Corresponding spectrograms.}
\label{fig:ecg-qualitative-low}
\end{figure}

\begin{figure}[htbp]
\centering
\includegraphics[width=\linewidth]{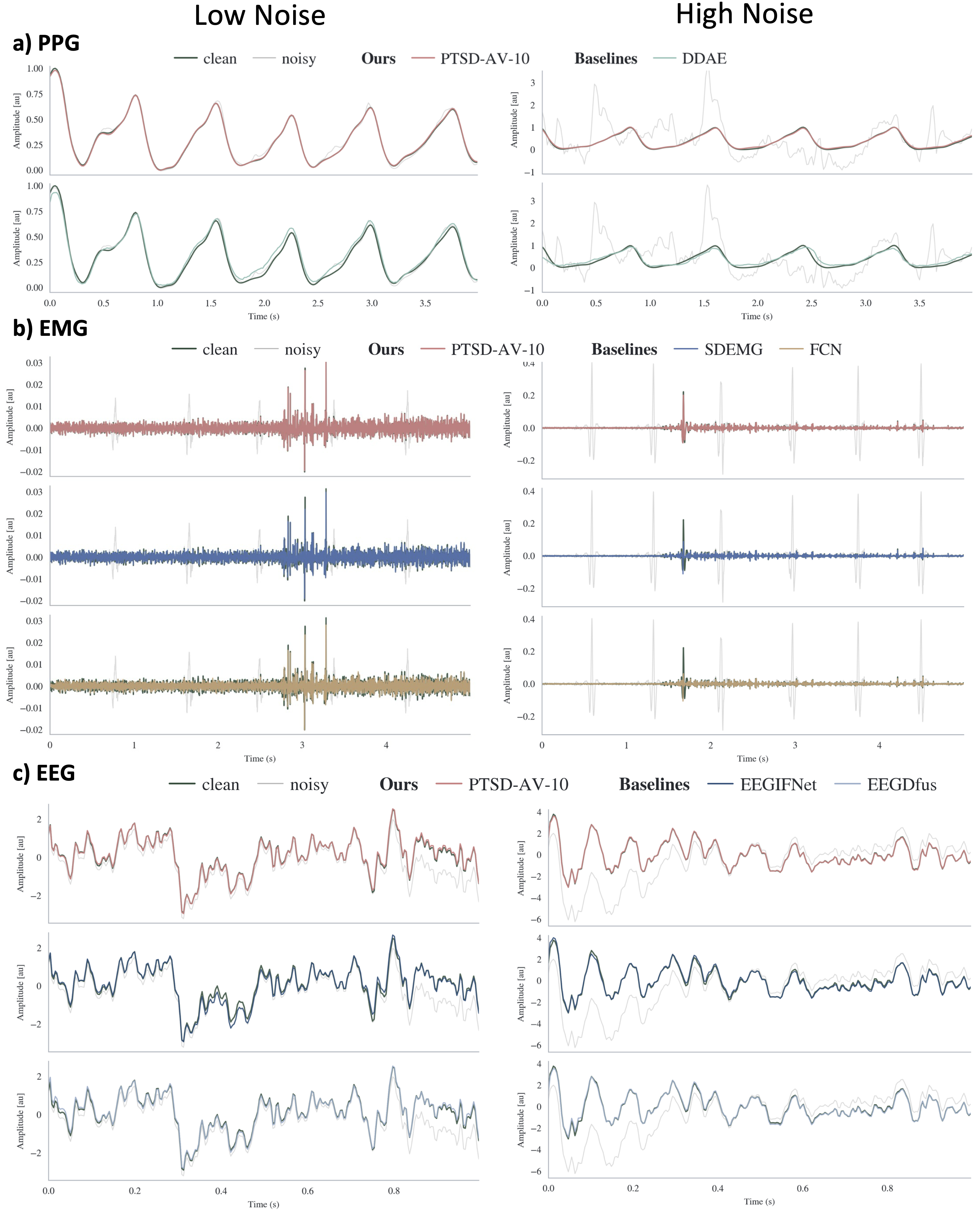}
\caption{Reconstructions for the non-ECG modalities at a light and a heavy corruption level: \textbf{(a)} PPG ($24$ and $-6$\,dB), \textbf{(b)} EMG ($0$ and $-14$\,dB), \textbf{(c)} EEG ($5$ and $-5$\,dB).}
\label{fig:all-modalities-qualitative}
\end{figure}

Figures~\ref{fig:ecg-qualitative-high}--\ref{fig:all-modalities-qualitative} show representative reconstructions. These are single samples and should be read as illustration rather than evidence.

% \begin{table}[htbp]
% \centering
% \caption{\ph{Pending experiments, ordered by value-to-cost. The top four change what the paper can claim; the remainder close gaps a reviewer will otherwise raise.}}
% \label{tab:placeholder_summary}
% \ph{%
% \small
% \begin{tabular}{@{}llp{5.9cm}@{}}
% \toprule
% Tag & Cost & What it establishes \\
% \midrule
% \texttt{placeholder\_av\_transfer} & Low & Antithetic coupling is a property of the sampler, not of \method{}. Checkpoint reuse, no retraining. \\
% \texttt{placeholder\_cyclicity\_index} & Low & Removes the circularity in Fig.~\ref{fig:scaling}b by replacing a model-derived axis with a clean-data one. \\
% \texttt{placeholder\_fig\_jitter\_sweep} & Medium & Turns an $n{=}4$ ordering into a continuum. The load-bearing evidence for the central claim. \\
% \texttt{placeholder\_ablation\_modalities} & Medium & Tests the differential prediction: frame ablation uniform, phase ablation ordered by $\Pi$. \\
% \midrule
% \texttt{placeholder\_downstream} & Medium & Task utility beyond distortion for PPG/EMG/EEG. EEG band-power and PPG HR need no training. \\
% \texttt{placeholder\_rho\_measurement} & Low & Verifies the $\rho<0$ condition Prop.~\ref{prop:av} depends on. \\
% \texttt{placeholder\_compute} & Low & Params, NFE, wall-clock, throughput. Reviewers ask for this. \\
% \texttt{placeholder\_seeds} & Medium & Seed variance on the headline table; several ablation gaps are CI-sized. \\
% \texttt{placeholder\_crossdataset\_mod.} & High & Distribution shift for the non-ECG modalities. \\
% \bottomrule
% \end{tabular}}
% \end{table}

\end{document}